%% file: paper.tex
\newif\ifconf\conffalse

\ifconf
  \documentclass[runningheads]{llncs}
\else
  \documentclass[10pt,twocolumn,letterpaper]{article}
\fi

\newif\ifappendix
\newif\ifacks

\ifconf
  \usepackage{eccv}
  \appendixfalse
  \ackstrue
\else
  \usepackage[pagenumbers]{cvpr}
  \appendixtrue
  \ackstrue
\fi

\ifconf
  \usepackage{hyperref}
  \usepackage[accsupp]{axessibility}
  \usepackage{orcidlink}
\else
  \usepackage[pagebackref,breaklinks,colorlinks,allcolors=cvprblue]{hyperref}
\fi
\usepackage[utf8]{inputenc}
\usepackage[T1]{fontenc}
\usepackage{url}
\usepackage{booktabs}
\usepackage{amsfonts}
\usepackage{nicefrac}
\usepackage{silence}
\usepackage{microtype}
\usepackage{color}
\usepackage{graphicx}
\usepackage{tikz}
\usetikzlibrary{backgrounds,calc,arrows.meta}
\usepackage{amsmath,amssymb}
\usepackage{mathtools}
\usepackage[normalem]{ulem}
\usepackage{tabu}
\usepackage{multirow}
\usepackage{pgfplots,pgfplotstable}
\usepgfplotslibrary{fillbetween}
\usepackage{calc}
\usepackage{pbox}
\usepackage{algorithm}
\usepackage[noend]{algpseudocode}
\usepackage{caption}
\usepackage{afterpage}
\usepackage{subcaption}
\usepackage{listings}
\usepackage{tcolorbox}
\usepackage{arrayjobx}
\usepackage{lipsum}
\usepackage[outline]{contour}
\usepackage{placeins}

\definecolor{cvprblue}{rgb}{0.21,0.49,0.74}

\pgfplotsset{compat=1.14}
\pgfplotsset{compat/show suggested version=false}

\ifconf
\makeatletter\renewcommand{\paragraph}{%
\@startsection{paragraph}{4}%
  {\z@}{1.25ex \@plus .25ex \@minus .75ex}{-1.25em}%
  {\normalfont\normalsize\itshape}%
}\makeatother
\else
\makeatletter\renewcommand{\paragraph}{%
  \@startsection{paragraph}{4}%
  {\z@}{1.25ex \@plus .25ex \@minus .75ex}{-1.25em}%
  {\normalfont\normalsize\bfseries}%
}\makeatother
\fi

\graphicspath{{./figures/}}
\input{macros}

\input{data/data-metrics}
\ifappendix
  \input{data/data-comparison}

\fi
\ifconf
  \input{figures-conf}
\else
  \input{figures}

\fi

\ifconf
  \title{Finite Difference Flow Optimization for RL Post-Training of Text-to-Image Models}
\else
  
  \title{Finite Difference Flow Optimization for RL Post-Training of \\ Text-to-Image Models}
\fi

\ifconf
  \authorrunning{McAllister et al.}
  \titlerunning{FDFO for RL Post-Training of Text-to-Image Models}
  \author{%
  David McAllister\inst{1}\thanks{Work done during an internship at NVIDIA.} \and
  Miika Aittala\inst{2} \and
  Tero Karras\inst{2} \and
  Janne Hellsten\inst{2} \and \\
  Angjoo Kanazawa\inst{1} \and
  Timo Aila\inst{2} \and
  Samuli Laine\inst{2}%
  }%
  \institute{$^1$ UC Berkeley \quad $^2$ NVIDIA}
\else
  \newcommand{\makeauthor}[2]{{#1}\\{#2}}
  \newcommand{\padauthor}{}
  \author{%
  \padauthor{}%
  \and\makeauthor{David McAllister\thanks{Work done during an internship at NVIDIA.}}{UC Berkeley}%
  \and\makeauthor{Miika Aittala}{NVIDIA}%
  \and\makeauthor{Tero Karras}{NVIDIA}%
  \and\makeauthor{Janne Hellsten}{NVIDIA}%
  \and\padauthor{}%
  \and\makeauthor{Angjoo Kanazawa}{UC Berkeley}%
  \and\makeauthor{Timo Aila}{NVIDIA}%
  \and\makeauthor{Samuli Laine}{NVIDIA}%
  }
\fi

\begin{document}
\maketitle

\ifappendix
\newcommand{\refAppResults}{\ref{app:results}}
\newcommand{\refAppImplementation}{\ref{app:implementation}}
\newcommand{\refAppStochasticity}{\ref{app:stochasticity}}
\newcommand{\refAppSchedules}{\ref{app:schedules}}
\newcommand{\refAppTheory}{\ref{app:theory}}
\else
\newcommand{\refAppResults}{A}
\newcommand{\refAppImplementation}{B}
\newcommand{\refAppStochasticity}{B.1}
\newcommand{\refAppSchedules}{B.2}
\newcommand{\refAppTheory}{C}
\fi

\begin{abstract}

Reinforcement learning (RL) has become a standard technique for post-training diffusion-based image synthesis models,
  as it enables learning from reward signals to explicitly improve desirable aspects such as image quality and prompt alignment.
In this paper, we propose an online RL variant that reduces the variance in the model updates by
  sampling paired trajectories and pulling the flow velocity in the direction of the more favorable image.
Unlike existing methods that treat each sampling step as a separate policy action,
  we consider the entire sampling process as a single action.
We experiment with both high-quality vision language models and off-the-shelf quality metrics for rewards,
  and evaluate the outputs using a broad set of metrics.
Our method converges faster and yields higher output quality and prompt alignment than previous approaches.

\end{abstract}

\section{Introduction}
\label{sec:intro}

Generation of synthetic images based on text prompts has become a ubiquitous task for deep learning models.
The dominant paradigm today is using diffusion models \cite{SohlDickstein2015,Ho2020,Song2020ddim,Song2019gradients,Song2021sde}
  that generate images by reversing a stochastic corruption process.
These models learn a prompt-conditional probability flow that 
  maps a random sample of noise into an image on the data manifold.
Flow matching \cite{Liu2022flow,Lipman2023FlowMatching,Heitz2023,Albergo2023NormalizingFlows,Delbracio2023inversion}
  is a popular parameterization for this flow \cite{albergo2025interpolants}.

As with large language models, the training of image synthesis models is commonly divided into
  pre-training and post-training stages that have different goals \cite{Ouyang2022_InstructGPT}.
Pre-training is performed on large-scale, weakly curated data.
The goal is to inject as much real-world knowledge into the model as possible
  with relatively little concern for image or prompt quality.
In the typically much shorter post-training phase, the model's sample distribution is then concentrated on regions
representing
desirable outputs based on an external reward metric or a carefully curated fine-tuning set.

The post-training of image generators typically has a wide variety of vague goals, such as making the images more beautiful, have nicer composition, more expressive lighting, etc.
This ``true reward'' cannot be expressed as a loss function, and thus cannot be directly optimized.
What can be done instead is to define a set of quantifiable \emph{proxy rewards} that hopefully address many of the desirable aspects.
Reinforcement learning (RL) is a general method that can, in principle, optimize any such proxy reward.
Common proxy rewards include learned models trained to mimic human preferences \cite{kirstain2023,wu2023_hpsv2} along with more targeted goals, such as the clarity of text rendering.

The pre-training objective of diffusion models is minimized by a unique flow field determined by the training dataset.
As such, the training constantly supervises that the generated distribution stays in alignment.
Unfortunately, this is no longer true during post-training,
  and aspects that are irrelevant to the specified reward are left to drift freely;
  for example, a reward targeting accurate text rendering can disregard or even harm the overall image quality.
While this ``reward hacking'' \cite{Amodei2016hacking} is in the nature of RL\,---\,%
  the policy is only designed to optimize the provided reward\,---\,%
  the RL algorithm should not exacerbate the problem by, e.g.,
  actively drifting in random directions along the underspecified dimensions.
Such bad behavior can be partially mitigated by, e.g., specifying multiple rewards,
  by limiting the amount of allowed change via KL regularization,
  or by mixing in some pre-training objective.

Existing RL post-training methods for diffusion models typically recast stochastic
  sampling as a Markov decision process (MDP), i.e., a sequence of policy-guided
  actions with Gaussian transitions~\cite{Black2023_DDPO,10.5555/3666122.3669619}.
DDPO \cite{Black2023_DDPO} adopts the PPO \cite{Schulman2017PPO} algorithm
  for reward optimization under the MDP formulation.
Flow-GRPO \cite{Liu2025_FlowGRPO} and DanceGRPO \cite{Xue2025_DanceGRPO}
  apply the same idea in the context of flow matching.
In this approach, the proposed updates to the flow are random perturbations that are independent
  between the sampling steps.
These perturbations are reinforced if their overall effect on the sampling trajectory lands on a
  higher-than-average reward.

The noise in these updates is a serious weakness from the viewpoint just laid out.
While the aggregate update will improve the reward,
  only a small fraction of its magnitude contributes to this,
  while the rest is reward-neutral noise that pushes the flow around in random directions.
This poses several problems.
The speed of progress per update is significantly limited,
  as a large part of any individual update does not contribute to the goal.
The noise also causes the unrelated dimensions to drift freely, e.g.,
  cycling through random image styles if they are not constrained by any reward.
Finally, the drift also slowly accumulates into detrimental side effects,
  essentially setting a cap on how much fine-tuning can be done in total.
As an example of this, we show that Flow-GRPO starts to introduce artifacts into the
  generated images upon extended training.

The goal of our approach is to improve the signal-to-noise ratio of the flow updates.
Our method uses similar random perturbations to discover high-reward images,
  but decouples the resulting update from this random walk.
Specifically, our method generates two nearby images and uses their difference as an
  approximate gradient.
This image difference is weighted by the respective reward difference,
  so that it is guaranteed to point from the lower-reward image
  to the higher-reward one.
We then \emph{uniformly} update the flow directions along the generating trajectory towards this direction,
  relying on the de-facto ``non-rotational'' behavior specific to diffusion flows~\cite{khrulkov2022understandingddpmlatentcodes}.
Each sampling step thus receives an update that directly benefits the reward.
This is in contrast to the MDP formulation where, roughly speaking, close to half of the
  individual flow updates may be detrimental to the reward.

Compared to Flow-GRPO, our method converges significantly faster, to higher rewards,
  and with fewer reward hacking artifacts.
It can be used as a drop-in replacement for SOTA RL algorithms in post-training of diffusion models.
The simplicity and data efficiency of our approach also enables training on \emph{on-policy} human preference feedback.
These results position our formulation as a promising alternative for RL post-training.

Our implementation and trained models are available at {\small{\url{https://github.com/NVlabs/finite-difference-flow-optimization}}}

\section{Previous Work}
\label{sec:related}

Many methods have been proposed for post-training neural networks,
  and we will briefly review the most closely related ones here.
For a broader context,
  Liu et al.~\cite{Liu2024_AlignmentDiffusionSurvey} present a
  survey on image generator post-training,
  constructing a taxonomy for a wide range of literature on related
  algorithms and evaluation.
Also, Uehara et al.~\cite{Uehara2024_Understanding} present a more RL-focused
  survey that examines the
  connections between different post-training methods in detail.

\paragraph{Supervised fine-tuning}
A common approach is to simply continue training using a smaller fine-tuning dataset.
Direct preference optimization (DPO) \cite{Rafailov2023_DPO},
  on the other hand, 
  relies on annotated human preference pairs.
It is simple to implement, building on supervised training without needing an online reward model.
However, it is limited to cases where annotated preference data is available.
Diffusion DPO \cite{Wallace2024_DiffusionDPO} applies the
  method to image generators. 

\paragraph{Differentiable rewards}
If the reward function is differentiable with respect to the generated image,
  it is possible to backpropagate the reward gradients to the generator. 
Commonly used approaches in this category include
  ReFL \cite{Xu2023_ImageReward_ReFL},
  DRaFT \cite{Clark2024_DRaFT},
  and Deep Reward Supervision (DRS) \cite{Wu2024_DRS}.
These methods differ mainly in where
  exactly the supervision happens inside the generator\,---\,%
  ReFL supervises on a random step, DRaFT on a sequence of last steps, and
  DRS on a sparse subset of steps.
Dual-process image generation \cite{luo2025dualprocess} propagates gradients through
  a VLM reward to adjust image generator weights at inference time. 

Domingo-Enrich et al.~\cite{DomingoEnrich2025_AdjointMatching} cast
  reward fine-tuning as a stochastic optimal control (SOC) problem, which is strongly connected to RL~\cite{levine2018reinforcementlearningcontrolprobabilistic}. They formulate a noise schedule that is free of value function bias that typical post-training objectives suffer from.
They also present adjoint matching, a general method for solving
  SOC problems that removes high-variance importance weights from the regression objective.

\paragraph{Non-differentiable rewards}
In many practically relevant scenarios, the reward function is not differentiable, most notably when humans provide direct reward signals.
Although this is an obvious application of RL, a number of
  RL-adjacent supervised learning approaches have been proposed as well. 
  Off-policy sample-evaluate-update loops are used in language modeling~\cite{nakano2021webgptbrowserassistedquestionansweringhuman,zelikman2022starbootstrappingreasoningreasoning} and control settings~\cite{10.1023/A:1010091220143, 10.1145/1273496.1273590}. 
  In image generation, several authors apply these techniques through reward-weighted regression 
  \cite{Lee2023_AligningTextToImage,Black2023_DDPO,Fan2025_Online} as well as rejection sampling \cite{Dong2023_RAFT}.

True RL methods, applicable to any reward, include the aforementioned MDP-based methods
  DDPO \cite{Black2023_DDPO}, Flow-GRPO \cite{Liu2025_FlowGRPO} and DanceGRPO \cite{Xue2025_DanceGRPO},
  with the last two demonstrating current state-of-the-art results.
Other works perform on-policy RL with the standard denoising objective~\cite{mcallister2025flowmatchingpolicygradients,xue2025advantageweightedmatchingaligning}, though they rely on strong regularization for stability in image generation settings. Some works also apply variants of value-based RL~\cite{Watkins1992} to diffusion models \cite{Jia2025_LaSRO, fql_park2025, psenka2024qsm}.

GFlowNets~\cite{Bengio2021_GFlowNets,Zhang2024_ImprovingGFlowNets,Liu2025_NablaGFlowNet} also rely on the MDP formulation
  but propose a novel training objective
  that is closely related to path consistency learning \cite{Nachum2017_PCL, Uehara2024_Understanding}.
The aim is to tilt the distribution proportionally to the reward, the same goal as KL-regularized RL~\cite{levine2018reinforcementlearningcontrolprobabilistic}. However, performance vs.~mainstream SOTA (Flow-GRPO, DanceGRPO)
  is unknown.

\ifconf\else
\input{data/trajectories_def}\begin{figure*}[t]
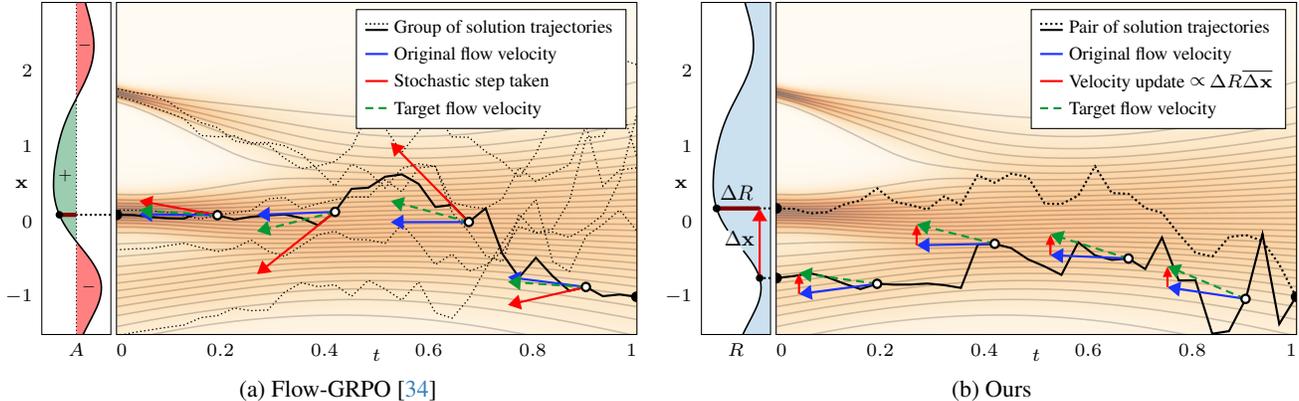
%
\centering%
\hspace*{-1mm}\figTrajectoriesRewardPlotBegin{A}%
\begin{scope}[on background layer]
\figTrajectoriesPolygonRewPos{\fill[ForestGreen!40!white]}%
\figTrajectoriesPolygonRewNegA{\fill[red!50!white]}%
\figTrajectoriesPolygonRewNegB{\fill[red!50!white]}%
\end{scope}
\node at (0.62,  2.33) {\resizebox{1.8mm}{!}{$-$}};
\node at (0.34,  0.60) {\resizebox{1.8mm}{!}{$+$}};
\node at (0.68, -0.87) {\resizebox{1.8mm}{!}{$-$}};
\addplot[black, forget plot] table [x index=0, y index=1] {data/trajectories_reward.txt};
\draw[black, densely dotted] (0.5, \figTrajectoriesPlotxmin) -- (0.5, \figTrajectoriesPlotxmax);
\draw[black, thick, densely dotted] (1.0, \figTrajectoriesGrpoBaseEndX) -- (.5, \figTrajectoriesGrpoBaseEndX);
\draw[red!50!black, ultra thick] (.5, \figTrajectoriesGrpoBaseEndX) -- (\figTrajectoriesGrpoBaseR, \figTrajectoriesGrpoBaseEndX);
\fill[TC0] (\figTrajectoriesGrpoBaseR, \figTrajectoriesGrpoBaseEndX) circle[radius=.5mm];
\figTrajectoriesRewardPlotEnd{}%
\figTrajectoriesCommonBegin{}%
\foreach \grpidx in {\figTrajectoriesGrpoGroupFirst,...,\figTrajectoriesGrpoGroupLast}
  \addplot[black, line width=0.5pt, densely dotted, forget plot] table [x index=0, y index=\grpidx] {data/trajectories.txt};
\addplot[TC0, thick, forget plot] table [x index=0, y index=\figTrajectoriesGrpoBase] {data/trajectories.txt};
\addlegendimage{legend image with pair}
\addlegendentry[right]{Group of solution trajectories}
\figTrajectoriesGrpoUpdateStepdirs{\draw[TC2, thick, -Triangle]}{--}{}
\addplot[TC1, thick](-1,0);\addlegendentry[right]{Original flow velocity}
\addplot[TC2, thick](-1,0);\addlegendentry[right]{Stochastic step taken}
\figTrajectoriesGrpoUpdateVelocities{\draw[TC1, thick, -Triangle]}{--}{}
\figTrajectoriesGrpoUpdateVelocitiesUpdated{\draw[TC3, thick, densely dashed, -Triangle]}{--}{}
\addplot[TC3, densely dashed, thick](-1,0);\addlegendentry[right]{Target flow velocity}
\figTrajectoriesGrpoUpdatePts{\fill[TC0]}{circle[radius=.7mm]}
\figTrajectoriesGrpoUpdatePts{\fill[white]}{circle[radius=.4mm]}
\fill[TC0] \figTrajectoriesGrpoBaseStart circle[radius=.7mm];
\fill[TC0] \figTrajectoriesGrpoBaseEnd circle[radius=.7mm];
\figTrajectoriesCommonEnd%
\figTrajectoriesRewardPlotBegin{R}%
\begin{scope}[on background layer]
\figTrajectoriesPolygonRewFull{\fill[Gray!20!CornflowerBlue!30!white]}%
\end{scope}
\addplot[black, forget plot] table [x index=0, y index=1, restrict y to domain=\figTrajectoriesPlotxmin:-0.35] {data/trajectories_reward.txt};
\addplot[black, forget plot] table [x index=0, y index=1, restrict y to domain=-0.15:\figTrajectoriesPlotxmax] {data/trajectories_reward.txt};
\draw[black, thick, densely dotted] (1.0, \figTrajectoriesMatchingBaseEndX) -- (\figTrajectoriesMatchingBaseR, \figTrajectoriesMatchingBaseEndX);
\draw[black, thick, densely dotted] (1.0, \figTrajectoriesMatchingChurnedEndX) -- (\figTrajectoriesMatchingBaseR, \figTrajectoriesMatchingChurnedEndX);
\draw[red!50!black, ultra thick] (\figTrajectoriesMatchingBaseR, \figTrajectoriesMatchingChurnedEndX) -- (\figTrajectoriesMatchingChurnedR, \figTrajectoriesMatchingChurnedEndX);
\draw[TC2, thick, -Triangle] (\figTrajectoriesMatchingBaseR, \figTrajectoriesMatchingBaseEndX) -- (\figTrajectoriesMatchingBaseR, \figTrajectoriesMatchingChurnedEndX);
\fill[TC0] (\figTrajectoriesMatchingBaseR, \figTrajectoriesMatchingBaseEndX) circle[radius=.5mm];
\fill[TC0] (\figTrajectoriesMatchingChurnedR, \figTrajectoriesMatchingChurnedEndX) circle[radius=.5mm];
\fill[TC0] (\figTrajectoriesMatchingChurnedR, \figTrajectoriesMatchingChurnedEndX) circle[radius=.5mm];
\node at (0.51,  0.38) {\resizebox{!}{1.2ex}{$\DeltaR$}};
\node at (0.55, -0.22) {\resizebox{!}{1.2ex}{$\Deltax$}};
\figTrajectoriesRewardPlotEnd{}%
\figTrajectoriesCommonBegin{}%
\addplot[color=TC0, thick, forget plot] table [x index=0, y index=\figTrajectoriesMatchingBase] {data/trajectories.txt};
\addlegendimage{legend image with pairb}
\addlegendentry[right]{Pair of solution trajectories}
\addplot[color=TC0, thick, densely dotted, forget plot] table [x index=0, y index=\figTrajectoriesMatchingChurned] {data/trajectories.txt};
\addplot[TC1, thick](-1,0);\addlegendentry[right]{Original flow velocity}
\addplot[TC2, thick](-1,0);\addlegendentry[right]{Velocity update \raisebox{.2ex}{$\propto$}$\,\DeltaR\smash{\overline{\Delta\boldx}}$}
\figTrajectoriesMatchingUpdateVelocities{\draw[TC1, thick, -Triangle]}{--}{}
\figTrajectoriesMatchingUpdateVelocitiesUpdated{\draw[TC3, thick, densely dashed, -Triangle]}{--}{}
\addplot[TC3, densely dashed, thick](-1,0);\addlegendentry[right]{Target flow velocity}
\figTrajectoriesMatchingUpdateAdded{\draw[TC2, thick, -{Triangle[angle=45:2pt 3]}]}{--}{}
\figTrajectoriesMatchingUpdatePoints{\fill[TC0]}{circle[radius=.7mm]}
\figTrajectoriesMatchingUpdatePoints{\fill[white]}{circle[radius=.4mm]}
\fill[TC0] \figTrajectoriesMatchingBaseStart circle[radius=.7mm];
\fill[TC0] \figTrajectoriesMatchingBaseEnd circle[radius=.7mm];
\fill[TC0] \figTrajectoriesMatchingChurnedEnd circle[radius=.7mm];
\figTrajectoriesCommonEnd%
\\
\makebox[1mm][c]{\ }%
\makebox[87mm][c]{\small (a) Flow-GRPO \cite{Liu2025_FlowGRPO}}%
\makebox[86.8mm][c]{\small (b) Ours}%
\vspace*{-.5ex}%
\caption{\label{figTrajectories}%
Illustration of differences between denoising MDPs, including Flow-GRPO \cite{Liu2025_FlowGRPO}, and our method.
\textbf{(a)} For each prompt, Flow-GRPO samples a group of trajectories using stochastic Euler--Maruyama sampling.
The relative advantage $A$ of each sample is the reward normalized with the group statistics.
For each time step, the flow velocity is optimized towards the stochastic perturbation the trajectory took,
  or away from it if the sample's advantage is negative.
\textbf{(b)} In our method, we sample a pair of trajectories and compute the difference $\Delta\boldx$ between the output images.
The flow velocity is optimized towards normalized $\smash{\overline{\Delta\boldx}}$
  scaled by the reward delta $\DeltaR$.
Note that both Flow-GRPO and our method apply updates to all sampled trajectories,
  whereas only one is highlighted here for clarity.
}%
\end{figure*}

\fi

\paragraph*{Inference-time optimization}
Orthogonal to model post-training,
  inference-time tweaks to the sampling process can provide improvements
  to output quality at the cost of reducing diversity.
These techniques include, e.g., 
  classifier-free guidance (CFG) \cite{Ho2021classifierfree}, 
  gradient-guided diffusion \cite{GGD_2024},
  and loss-guided diffusion \cite{Song2023_LossGuidedDiffusion}.

\section{Background}
\label{sec:background}

Flow matching \cite{Liu2022flow,Lipman2023FlowMatching,Heitz2023,Albergo2023NormalizingFlows,Delbracio2023inversion} is
  a parameterization of a probability flow diffusion process \cite{SohlDickstein2015,Ho2020,Song2020ddim,Song2019gradients,Song2021sde}.
It is popular for its intuitive formulation and ease of implementation,
  as well as good practical performance stemming from favorable built-in design choices. 

Flow matching works by drawing an initial random noise image $\noisedraw \sim \mathcal{N}(\boldzero, \boldI_N)$ of $N$ pixels, and numerically solving an ordinary differential equation (ODE) 
\begin{equation}\vspace{-1mm}
\mathrm{d}\boldx(t) = \vel(\boldx; t, \condemb) \mathrm{d}t
\end{equation}
backward in time from \mbox{$t=1$} to \mbox{$t=0$} with initial condition \mbox{$\boldx(1) = \noisedraw$}.
Here, $\vel(\cdot)$ is the \emph{velocity function},
  implemented as a neural network with weights $\theta$,
  that points towards the average of noise-free images consistent with the noisy image $\boldx(t)$ and time parameter $t$,
  conditioned on prompt embedding $\condemb$ for conditional generation.
Following the ODE pushes the image towards the evolving denoising estimate under the training image dataset,
  gradually reducing the noise level and revealing a clean image $\boldx(0)$ from this distribution.

In practice, the ODE is solved by evaluating $\boldx(t)$ at discrete time steps
  \mbox{$t_0=1,\,t_1,\,\ldots,\,t_{T-1},\,t_T=0$},
  where each step yields an intermediate image {$\boldx_i := \boldx(t_i)$} that is slightly less noisy than the previous one. 
Using the simple Euler solver scheme,
  the step from time $t_i$ to $t_{i+1}$ is given by $\boldx_{i+1} = \boldx_i + (t_{i+1} - t_i) \vel(\boldx_i; t_i,\condemb)$,
  so that after $T$ steps the initial noise image at $t_0$ is transformed into the generated image at $t_T$.
We refer to a sequence of images \mbox{$\boldx_0,\,\boldx_1,\,\ldots,\,\boldx_T$} as a sampling trajectory.

\ifconf
\input{data/trajectories_def}\begin{figure*}[t]%
\centering%
\hspace*{-1mm}\figTrajectoriesRewardPlotBegin{A}%
\begin{scope}[on background layer]
\figTrajectoriesPolygonRewPos{\fill[ForestGreen!40!white]}%
\figTrajectoriesPolygonRewNegA{\fill[red!50!white]}%
\figTrajectoriesPolygonRewNegB{\fill[red!50!white]}%
\end{scope}
\node at (0.62,  2.33) {\resizebox{1.8mm}{!}{$-$}};
\node at (0.34,  0.60) {\resizebox{1.8mm}{!}{$+$}};
\node at (0.68, -0.87) {\resizebox{1.8mm}{!}{$-$}};
\addplot[black, forget plot] table [x index=0, y index=1] {data/trajectories_reward.txt};
\draw[black, densely dotted] (0.5, \figTrajectoriesPlotxmin) -- (0.5, \figTrajectoriesPlotxmax);
\draw[black, thick, densely dotted] (1.0, \figTrajectoriesGrpoBaseEndX) -- (.5, \figTrajectoriesGrpoBaseEndX);
\draw[red!50!black, ultra thick] (.5, \figTrajectoriesGrpoBaseEndX) -- (\figTrajectoriesGrpoBaseR, \figTrajectoriesGrpoBaseEndX);
\fill[TC0] (\figTrajectoriesGrpoBaseR, \figTrajectoriesGrpoBaseEndX) circle[radius=.5mm];
\figTrajectoriesRewardPlotEnd{}%
\figTrajectoriesCommonBegin{}%
\foreach \grpidx in {\figTrajectoriesGrpoGroupFirst,...,\figTrajectoriesGrpoGroupLast}
  \addplot[black, line width=0.5pt, densely dotted, forget plot] table [x index=0, y index=\grpidx] {data/trajectories.txt};
\addplot[TC0, thick, forget plot] table [x index=0, y index=\figTrajectoriesGrpoBase] {data/trajectories.txt};
\addlegendimage{legend image with pair}
\addlegendentry[right]{Group of solution trajectories}
\figTrajectoriesGrpoUpdateStepdirs{\draw[TC2, thick, -Triangle]}{--}{}
\addplot[TC1, thick](-1,0);\addlegendentry[right]{Original flow velocity}
\addplot[TC2, thick](-1,0);\addlegendentry[right]{Stochastic step taken}
\figTrajectoriesGrpoUpdateVelocities{\draw[TC1, thick, -Triangle]}{--}{}
\figTrajectoriesGrpoUpdateVelocitiesUpdated{\draw[TC3, thick, densely dashed, -Triangle]}{--}{}
\addplot[TC3, densely dashed, thick](-1,0);\addlegendentry[right]{Target flow velocity}
\figTrajectoriesGrpoUpdatePts{\fill[TC0]}{circle[radius=.7mm]}
\figTrajectoriesGrpoUpdatePts{\fill[white]}{circle[radius=.4mm]}
\fill[TC0] \figTrajectoriesGrpoBaseStart circle[radius=.7mm];
\fill[TC0] \figTrajectoriesGrpoBaseEnd circle[radius=.7mm];
\figTrajectoriesCommonEnd%
\figTrajectoriesRewardPlotBegin{R}%
\begin{scope}[on background layer]
\figTrajectoriesPolygonRewFull{\fill[Gray!20!CornflowerBlue!30!white]}%
\end{scope}
\addplot[black, forget plot] table [x index=0, y index=1, restrict y to domain=\figTrajectoriesPlotxmin:-0.35] {data/trajectories_reward.txt};
\addplot[black, forget plot] table [x index=0, y index=1, restrict y to domain=-0.15:\figTrajectoriesPlotxmax] {data/trajectories_reward.txt};
\draw[black, thick, densely dotted] (1.0, \figTrajectoriesMatchingBaseEndX) -- (\figTrajectoriesMatchingBaseR, \figTrajectoriesMatchingBaseEndX);
\draw[black, thick, densely dotted] (1.0, \figTrajectoriesMatchingChurnedEndX) -- (\figTrajectoriesMatchingBaseR, \figTrajectoriesMatchingChurnedEndX);
\draw[red!50!black, ultra thick] (\figTrajectoriesMatchingBaseR, \figTrajectoriesMatchingChurnedEndX) -- (\figTrajectoriesMatchingChurnedR, \figTrajectoriesMatchingChurnedEndX);
\draw[TC2, thick, -Triangle] (\figTrajectoriesMatchingBaseR, \figTrajectoriesMatchingBaseEndX) -- (\figTrajectoriesMatchingBaseR, \figTrajectoriesMatchingChurnedEndX);
\fill[TC0] (\figTrajectoriesMatchingBaseR, \figTrajectoriesMatchingBaseEndX) circle[radius=.5mm];
\fill[TC0] (\figTrajectoriesMatchingChurnedR, \figTrajectoriesMatchingChurnedEndX) circle[radius=.5mm];
\fill[TC0] (\figTrajectoriesMatchingChurnedR, \figTrajectoriesMatchingChurnedEndX) circle[radius=.5mm];
\node at (0.51,  0.38) {\resizebox{!}{1.2ex}{$\DeltaR$}};
\node at (0.55, -0.22) {\resizebox{!}{1.2ex}{$\Deltax$}};
\figTrajectoriesRewardPlotEnd{}%
\figTrajectoriesCommonBegin{}%
\addplot[color=TC0, thick, forget plot] table [x index=0, y index=\figTrajectoriesMatchingBase] {data/trajectories.txt};
\addlegendimage{legend image with pairb}
\addlegendentry[right]{Pair of solution trajectories}
\addplot[color=TC0, thick, densely dotted, forget plot] table [x index=0, y index=\figTrajectoriesMatchingChurned] {data/trajectories.txt};
\addplot[TC1, thick](-1,0);\addlegendentry[right]{Original flow velocity}
\addplot[TC2, thick](-1,0);\addlegendentry[right]{Velocity update \raisebox{.2ex}{$\propto$}$\,\DeltaR\smash{\overline{\Delta\boldx}}$}
\figTrajectoriesMatchingUpdateVelocities{\draw[TC1, thick, -Triangle]}{--}{}
\figTrajectoriesMatchingUpdateVelocitiesUpdated{\draw[TC3, thick, densely dashed, -Triangle]}{--}{}
\addplot[TC3, densely dashed, thick](-1,0);\addlegendentry[right]{Target flow velocity}
\figTrajectoriesMatchingUpdateAdded{\draw[TC2, thick, -{Triangle[angle=45:2pt 3]}]}{--}{}
\figTrajectoriesMatchingUpdatePoints{\fill[TC0]}{circle[radius=.7mm]}
\figTrajectoriesMatchingUpdatePoints{\fill[white]}{circle[radius=.4mm]}
\fill[TC0] \figTrajectoriesMatchingBaseStart circle[radius=.7mm];
\fill[TC0] \figTrajectoriesMatchingBaseEnd circle[radius=.7mm];
\fill[TC0] \figTrajectoriesMatchingChurnedEnd circle[radius=.7mm];
\figTrajectoriesCommonEnd%
\\
\makebox[1mm][c]{\ }%
\makebox[87mm][c]{\small (a) Flow-GRPO \cite{Liu2025_FlowGRPO}}%
\makebox[86.8mm][c]{\small (b) Ours}%
\vspace*{-.5ex}%
\caption{\label{figTrajectories}%
Illustration of differences between denoising MDPs, including Flow-GRPO \cite{Liu2025_FlowGRPO}, and our method.
\textbf{(a)} For each prompt, Flow-GRPO samples a group of trajectories using stochastic Euler--Maruyama sampling.
The relative advantage $A$ of each sample is the reward normalized with the group statistics.
For each time step, the flow velocity is optimized towards the stochastic perturbation the trajectory took,
  or away from it if the sample's advantage is negative.
\textbf{(b)} In our method, we sample a pair of trajectories and compute the difference $\Delta\boldx$ between the output images.
The flow velocity is optimized towards normalized $\smash{\overline{\Delta\boldx}}$
  scaled by the reward delta $\DeltaR$.
Note that both Flow-GRPO and our method apply updates to all sampled trajectories,
  whereas only one is highlighted here for clarity.
}%
\end{figure*}

\fi

\subsection{Reward Maximization}
\label{sec:rewardmax}

Given a reward function $\Reward(\boldx)$ that maps images to scalar reward values,
  our goal is to fine-tune the pre-trained velocity function $\vel$ to maximize the expected reward over draws from the deterministically sampled generative model:
\begin{align}
  \argmax_\theta \mathbb{E}_{\boldc\sim\trainingprompts,\,\boldx_0 \sim \mathcal{N}(\boldzero, \boldI)} \Reward\big(f_\theta(\boldx_0; \condemb)\big)
  \text{,}
\end{align}
where $\trainingprompts$ is the distribution of fine-tuning prompt embeddings,
  and $f_\theta(\cdot)$ represents the process of drawing a sample starting from an initial noise $\boldx_0$
  using velocity function $\vel$.
Geometrically, the goal is to redirect the flow velocities such that a larger mass of noises map to high-reward regions of the image space. Generally, the reward may be non-differentiable or highly discontinuous.

\subsection{Markov Decision Process Formulation}
As an alternative to ODE-based deterministic sampling, diffusion models also allow for SDE-based \emph{stochastic} sampling under the same flow. Stochastic sampling injects fresh Gaussian noise at each time step to randomize the trajectory.

MDP based approaches~\cite{Black2023_DDPO, 10.5555/3666122.3669619, Liu2025_FlowGRPO, Xue2025_DanceGRPO} take advantage of stochasticity and cast each stochastic step as a Gaussian-distributed action distribution whose mean is defined by the flow velocity. Beneficial actions are reinforced by pulling the velocity toward the random steps of high-reward trajectories.

Fig.~\ref{figTrajectories}a illustrates the group-relative approach used by Flow-GRPO \cite{Liu2025_FlowGRPO} and DanceGRPO \cite{Xue2025_DanceGRPO} algorithms. They adopt the denoising MDP for flow schedules and estimate the advantage of each trajectory by normalizing its corresponding reward within a group.
The advantages determine the weight of the corresponding updates to the flow.
While these updates point towards more favorable images on average, a significant proportion of them point opposite of reward increase,
  contributing to undesirable variance.
  
\section{Our Method}
\label{sec:algo}

We adopt an approximate on-policy approach of alternating between
  (1) generating a set of trajectory rollouts from the current model, and 
  (2) training the network velocity predictions along these trajectories, so as to redirect them towards
      collecting higher reward at their endpoints.
To obtain a training signal without direct access to gradient of the reward,
  we propose to roll out \emph{pairs} of images with variations in detail
  and to reinforce the probability of the more favorable image among the two.

Specifically, we generate a pair of rollouts $\boldx_i$ and $\boldxhat_i$
  from a shared initial noise \mbox{$\boldx_0 = \boldxhat_0 = \noisedraw$},
  but apply a modest amount of stochasticity that perturbs them along the sampling trajectory.
This induces random differences in generated image details,
  whereby one of the output images \mbox{$\{\boldx_T,\boldxhat_T\}$} usually collects a higher reward than the other.
Then, the difference vector \mbox{$\Deltax = \boldxhat_T -\boldx_T$}
  weighted by the reward difference \mbox{$\DeltaR = \Reward(\boldxhat_T) - \Reward(\boldx_T)$}
will point towards the higher-reward image.
We train the velocities at all time steps along both trajectories to bend towards \mbox{$\DeltaR\Deltax$}.
The approach is illustrated in Fig.~\ref{figTrajectories}b.

\subsection{Analysis}
\label{sec:analysis}

The weighted difference $\DeltaR\Deltax$ points towards desirable changes in the generated image at
  \mbox{$t=0$}, whereas our updates divert the flow toward this direction at intermediate steps.
Our method thus relies on the expectation that this induces a similar change in the image.

To justify this informally, we note that the denoising process essentially fades out noise to reveal a hidden
  signal in a coarse-to-fine fashion.
Then, adding ``signal'', i.e., the reinforced image difference, at an intermediate noisy image will approximately
  pass it to the generated image.
This is a core underlying assumption in a broad range of methods
  (e.g., \cite{meng2023sdedit,chen2024tino,choi2021ilvr,kim2022diffusionclip})
that perform coarse edits on a partially
  noised image (say, adding a blob of green pixels),
and rely on the remainder of the flow to flesh out the detail while maintining the broad direction of the edit
  (say, into a tree).

More formally, 
  our update aims to satisfy
  $\hnabla \Reward(\boldx_T)\transp \boldJ_i(\boldx_i) \left[ \DeltaR\Deltax \right] \ge 0$ for each step $i$, where $\boldJ_i(\boldx_i)$ is the Jacobian matrix of the mapping induced by the flow from step $i$ onward.
While it \emph{is} possible to construct counter-examples that violate this condition,
  there are significant arguments for it holding at least to the extent required in practical image generation.
For example, it can be shown that under certain assumptions the condition holds on expectation if $\boldJ_i$ is positive semi-definite (see Appendix~\refAppTheory{}), which would be true for an optimal transport mapping~\cite{santambrogio2015optimal}.
The apparent similarity of diffusion flows and optimal transport mappings,
  while not theoretically exact~\cite{Lavenant2022flowmap},
  has inspired studies where a strong numerical similarity is
  nevertheless found~\cite{khrulkov2022understandingddpmlatentcodes}.

\algStochFM

\subsection{Stochastic Sampling}
Like Flow-GRPO, our approach requires a means to generate nearby random image variants.
As discussed by Karras et al.~\cite{Karras2022elucidating}, an Euler--Maruyama sampler
(used in, e.g., Flow-GRPO)
suffers from two related numerical problems when applied to flows. The velocity network is called with a slightly inconsistent time conditioning and noise amount at each step, as the impact of the noise injection sub-step is not accounted for. This is exacerbated by oversized noise injections that occur at some steps, because the amount of noise added is not proportioned to the existing noise level in the sample.

In Algorithm~\ref{alg:stochastic} we adapt the key idea from the EDM stochastic sampler \cite{Karras2022elucidating}
  to introduce stochasticity into the flow-matching solution trajectories.
When stepping from $t_i$ to $t_{i+1}$, 
  we take the ODE direction but overshoot the target time to a lower noise level,
  and then compensate by adding fresh random noise to land at $t_{i+1}$ (with appropriate scale corrections, see Appendix~\refAppStochasticity{} for details).

The parameters $\gamma_i$ specify the strength of stochastic randomization at each solver step.
Roughly, $\gamma_i$ expresses the fraction of noise in $\boldx_i$ that is re-randomized on that step,
  with value \mbox{$\gamma_i=0$} corresponding to deterministic flow matching.
Intuitively, re-randomizing at an intermediate noise level replaces the to-be-generated finer image detail
  that is still encoded by the noise with a different random realization,
  while keeping the already resolved coarser detail intact.

\subsection{Implementation Details}
\label{sec:implementation}

We present the full method pseudocode in Appendix~\refAppImplementation{}.

\paragraph{Stochasticity strength}
Where not specified otherwise, we use a uniform stochasticity schedule of \mbox{$\gamma_i = 0.0025$} for all timesteps. This induces variations in semantic detail of an image pair (e.g., moving or swapping out parts), but mostly retains their overall layout and content. We experimented with more complex schedules but found no consistent benefit to them. See Appendix~\refAppSchedules{} for details.

\paragraph{Normalization}
Neural network training benefits from approximately uniform gradient magnitude among training samples.
The random stochastic perturbations result in different magnitudes for $\Deltax$ between rollouts,
  and we can also expect $\DeltaR$ to be approximately (to the first order) proportional to $\Deltax$.
The raw training signal $\DeltaR\Deltax$ thus counts the norm $\|\Deltax\|_\mathrm{RMS}$ twice into its magnitude.
We cancel this effect by using normalized $\overline{\Deltax} = \Deltax/(\|\Deltax\|_\mathrm{RMS}^2 + 10^{-6})$ as the
  actual training signal.

\paragraph{Batching}

We train the model for up to 1000 epochs.
For each epoch, we generate 432 pairs of rollouts with randomly drawn prompts and store all \mbox{$T={}$40} steps of their trajectories, resulting in the same memory consumption as Flow-GRPO.
The resulting 432$\times$2$\times$40 samples are then randomly split into~4 training batches, totaling 8640 samples/batch.
The weighted image differences are backpropagated through the velocity function and accumulated,
  until they are updated into model weights $\theta$ after each batch using 
  AdamW~\cite{Loshchilov2019AdamW}.

\paragraph{On-policy optimization}

Making several training steps on a frozen set of rollouts throughout the epoch is sample-efficient, as computing fresh rollouts frequently would be expensive. However, it runs the risk of training on stale data, as the parameters $\theta$ may have changed significantly since the set of rollouts was refreshed. Following the usual RL policy optimization recipe, we apply clipping to downweight updates to velocities that have moved  too far from their original position since the last refresh. We use Simple Policy Optimization (SPO) \cite{Xie2025SPO} clipping that is similar to the widely used Proximal Policy Optimization (PPO-Clip) \cite{Schulman2017PPO}.

\section{Results}
\label{sec:results}

We will now evaluate our proposed RL algorithm experimentally to compare its performance against
  Flow-GRPO \cite{Liu2025_FlowGRPO}, a current state-of-the-art method.
We start by considering the effectiveness of our method in increasing the proxy reward as quickly as possible,
  using rewards of different complexity to highlight differences between the algorithms.
In these tests, we are mainly interested in convergence speed, highest obtainable reward, and potential algorithm-dependent artifacts.

We then employ external control metrics to analyze how the rewards, RL algorithms, and CFG strength affect image quality, prompt alignment, and diversity.
We ablate the major design choices of our algorithm as well as the key differences to Flow-GRPO.
Finally, we demonstrate on-policy human-in-the-loop training with human feedback as the live reward.
Additional results, metrics, and details for all experiments are provided in Appendix~\refAppResults{}.

\subsection{Experiment Setup}
\label{sec:setup}

We use the official implementation of Flow-GRPO \cite{Liu2025_FlowGRPO}.
We explored its hyperparameters to a considerable extent and concluded that the official defaults are close to optimal in all cases.
We thus use them with the following exceptions.
First, we disable exponential moving averaging of model weights as it was strictly harmful. 
Second, unless stated otherwise, we disable KL regularization~\cite{Liu2025_FlowGRPO} and CFG~\cite{Ho2021classifierfree}
  because they can mask the differences between RL algorithms and ideally would not be needed.
Third, we use 40 sampling steps during both training and inference.
The role of CFG is explored in Section~\ref{sec:bigpic}, and reducing the number of sampling steps during training is evaluated in Section~\ref{sec:wallclock}.

We perform all experiments using Stable Diffusion 3.5 Medium~\cite{StableDiffusion35}, extended with low-rank adaptation layers (LoRA) \cite{Hu2022lora} and update only these layers during post-training, in line with Flow-GRPO.
We generate all images at 512$\times$512 resolution.
The number of model and reward evaluations per epoch is the same for Flow-GRPO and our method.

We use two primary reward functions, PickScore\cite{Radford2021} and a novel VLM reward, that focus on predicting human preferences and prompt alignment, respectively.
Both are evaluated using the Pick-a-Pic~\cite{kirstain2023} training prompts.
Our use of a VLM reward is motivated by the observation made by Lin~et~al.~\cite{lin2024vqascore} that VLMs are surprisingly effective in estimating things like prompt alignment compared to explicit human preference models, and their performance can be expected to further improve as new models are released.
To this end, we adopt a scheme similar to VQA\-Score~\cite{lin2024vqascore} using Qwen2.5-VL-7B-Instruct~\cite{QWEN25}.
We feed in the generated image along with a text query
\emph{``Does this image match the caption {\texttt{"}}{$\,\cdots\,\!$}{\texttt{"}}? Answer Yes or No.''},
where $\,\cdots$ corresponds to the prompt that was used to generate the image.
We then run the VLM to predict the logits for the next token and calculate the reward as
  $100 \cdot \mathrm{sigmoid}(\mathrm{logits[Yes]} - \mathrm{logits[No]})$.
We also consider the weighted combination of these two rewards, defined as PickScore $+$ VLM alignment $/~10$.

For evaluation, we use the reward itself as well as a set of external control metrics, including OneIG-Bench~\cite{chang2025oneig} and HPSv2~\cite{wu2023_hpsv2}.
We calculate these after-the-fact based on checkpoints exported during training, so that the evaluation protocol remains independent of the post-training algorithm.

\ifconf
\else
  \begin{figure*}[p]
  \figRLWorksPlot\\[4ex]
  \figRLWorksImagesB\\[4ex]
  \figArtifacts\\[50ex]
  \end{figure*}

  \begin{figure*}[p]
  \figMetricsPlot\\[4ex]
  \figGuidancePlot\\[4ex]
  \figBiggerPictureImages\\[50ex]
  \vspace*{20mm}
  \end{figure*}
\fi

\ifconf\else
\figAblations
\fi

\subsection{Reward Optimization}
\label{sec:rlworks}

Fig.~\ref{figRLWorksPlot} compares our method with Flow-GRPO when optimizing three different rewards:
  (a) PickScore,
  (b) VLM-based prompt alignment, and 
  (c) their weighted combination.
In this test, an epoch is equally expensive for both methods.
PickScore is the easiest to optimize and both methods work well, although our method reaches a slightly higher reward much faster. 
The result achieved by Flow-GRPO matches the original paper closely.

The VLM-based alignment reward is more difficult to optimize; Flow-GRPO starts to struggle and the gap to our method widens significantly. 
The combined reward paints a similar picture with our method converging significantly faster to a higher reward.
More complicated rewards consisting of multiple goals can be beneficial as they can reduce the aspects the RL optimization is blind to.

Fig.~\ref{figRLWorksImagesB} shows representative image progressions for the combined reward.
The images are drawn from training snapshots corresponding to the dots annotated in Fig.~\ref{figRLWorksPlot} and demonstrate that higher rewards correspond to visually better results, as expected.
While subjective comparison is inherently difficult, we feel that our method stacks well against Flow-GRPO
  in reward-matched pairwise comparison.
Furthermore, there is a clear style drift in the Flow-GRPO images\,---\,this was a consistent behavior with all
the prompts we experimented with.
We attribute this style drift to Flow-GRPO's more noisy flow updates that lead to a greater amount of random mutation
of the flow until a given reward is reached.

We also note that Flow-GRPO eventually starts introducing grid-like reward hacking artifacts that fade in and out during training.
Fig.~\ref{figArtifacts} shows handpicked examples where this effect is severe.
The fact that these artifacts again disappear over time implies that the combined reward likely does penalize them, but not very strongly.  
We have never seen these artifacts in our method, even after equally long training.

\ifconf
  \begin{figure*}[p]
  \figRLWorksPlot\\[2.5ex]
  \figRLWorksImagesB\\[2.1ex]
  \figArtifacts\\[10ex]
  \end{figure*}

  \begin{figure*}[p]
  \figMetricsPlot\\[2.5ex]
  \figGuidancePlot\\[3.5ex]
  \figBiggerPictureImages\\[10ex]
  \vspace*{20mm}
  \end{figure*}
\fi

\ifconf
\figAblations
\fi

\subsection{Control Metrics}
\label{sec:bigpic}

Let us now further analyze the post-trained models
to see how the choice of reward affects alignment with prompts, alignment with human preferences, and diversity.
To estimate these, we use OneIG-Bench~\cite{chang2025oneig} (prompt alignment, diversity) and HPSv2~\cite{wu2023_hpsv2} (human preference) as independent control metrics.

Fig.~\ref{figMetricsPlot}a shows that the prompt alignment control metric is poorly optimized by the PickScore reward alone, as could be expected.
Our VLM reward, which specifically targets this aspect, does a much better job, but the combined reward works even more reliably.
Same conclusions apply to both our algorithm and Flow-GRPO, and the benefits of our method that were observed with the raw reward (Fig.~\ref{figRLWorksPlot}b) also hold for the control metric.
Fig.~\ref{figMetricsPlot}b shows that human preferences are well optimized by PickScore that focuses on this aspect.
VLM reward does not work particularly well here,
  but again the combined reward comes out on top using our algorithm.
Fig.~\ref{figMetricsPlot}c shows that all rewards reduce diversity,
  with PickScore causing the fastest decline and VLM reward the slowest.
Considering that our algorithm optimizes the rewards much faster than Flow-GRPO,
  diversity loss is approximately equally fast for the two algorithms on an iso-reward basis.
These observations position the combined reward as a clear choice over the more commonly used PickScore.

Fig.~\ref{figGuidancePlot} uses the same control metrics to assess the effects of CFG.
In these tests, we enable CFG for the base model and redo the entire RL post-training and evaluation
  using the CFG-adjusted velocity predictions in place of the original ones.
We can see that increasing the guidance strength improves alignment with prompts and human preferences,
  at the cost of further diversity loss.
This is not surprising, and the observations apply equally to our method and Flow-GRPO.

Fig.~\ref{figBiggerPictureImages} provides visual examples of this tradeoff.
Before post-training,
  the image quality of the base model is poor without CFG (Fig.~\ref{figBiggerPictureImages}a).
Enabling CFG in that setting (Fig.~\ref{figBiggerPictureImages}b) improves
  prompt alignment and image quality dramatically
  while reducing diversity (see Fig.~\ref{figGuidancePlot}abc, epoch~0).
With post-training, it is debatable whether CFG is useful, especially given its high inference cost.
Subjective inspection of the result images suggests that post-training without CFG improves
  image fidelity and detail significantly
  (Fig.~\ref{figBiggerPictureImages}c), and the use of CFG in this setting mainly compromises diversity
  and introduces the characteristic, high-contrast look (Fig.~\ref{figBiggerPictureImages}d).

\ifconf\else
\figPerformancePlot
\fi

\subsection{Ablations}
\label{sec:ablations}

\newcommand\AbRow[1]{(row~#1)}

Figure~\ref{figAblations} presents a number of ablation experiments, starting with the Flow-GRPO baseline \AbRow{A}.
The amount of stochasticity is tuned separately for each ablation variant.
Switching to our stochastic sampler in Algorithm~\ref{alg:stochastic} \AbRow{B} improves convergence considerably.
Replacing the Flow-GRPO policy gradients with our finite difference flow optimization \AbRow{C},
  while keeping the number of unique prompts per epoch unchanged,
  results in another major boost.
We then start all trajectories in each group from the same initial noise \AbRow{D}, which improves stability and improves the results further.
Finally, 
we can increase the number of unique prompts considered in each epoch by virtue of needing only two rollouts per prompt, which leads to our full method \AbRow{E}.

The bottom section of the table evaluates some of our specific design choices.
Instead of using stochastic sampling for both trajectories \AbRow{E}, 
  we can equally well use deterministic sampling for one of them \AbRow{F},
  which may be desirable in order to reduce the discrepancy between rollout generation and inference-time sampling.
Using PPO~\cite{Schulman2017PPO} \AbRow{G} in place of SPO~\cite{Xie2025SPO}
  \AbRow{E} yields similar results, although only when
  using the interval stochasticity schedule (Appendix~\refAppSchedules{})%
  \,---\,this was the only experiment where a
  constant $\gamma_i$ schedule was clearly inferior to a more complex schedule.
It is beneficial to normalize $\Delta \boldx$ as described in Section~\ref{sec:implementation}; disabling this
\AbRow{H} is highly detrimental to stability.
Replacing $\DeltaR\Deltax$ with the actual reward gradient obtained by
  backpropagating through the reward model
  \AbRow{I} does not improve the results,
  implying that our method is a good fit regardless
  of whether the reward is differentiable or not.
Further backpropagating this reward gradient through the sampling
  trajectory to determine the update for each sampling step \AbRow{J} is significantly
  worse than our approach of bending all sampling steps towards $\DeltaR\Deltax$.

\ifconf
\figPerformancePlot
\fi

\subsection{Wall-Clock Performance}
\label{sec:wallclock}

Fig.~\ref{figPerformancePlot}a plots convergence as a function of GPU hours,
  comparing our method and Flow-GRPO in baseline (40~sampling steps) and fast (10~steps) configurations.
As can be seen, our method converges significantly faster,
  crossing the highlighted combined reward level 19$\times$~faster in the baseline configuration and 5$\times$~faster
  in the fast configuration.
Because the official Flow-GRPO implementation suffers from significant implementation overheads, we also estimated
  zero-overhead convergence curves (Fig.~\ref{figPerformancePlot}b) to confirm that
  our wall-clock advantage is not merely a result of a more careful implementation.

\subsection{Direct Human Reward Function}
\label{sec:dhrf}

\figDHRF

As with other RL methods, our results depend heavily on the quality of proxy rewards.
In principle, the proxies can be eliminated in favor of direct feedback from a human in the training loop, assuming the optimization is sample efficient enough to converge in reasonable amount of time.
This leads to a method that uses human preferences on \emph{on-policy data}, instead of fine-tuning using a pre-collected dataset~\cite{Wallace2024_DiffusionDPO}.

To prototype this design, we ran a closed-loop generate-label-train procedure
  in which one of the authors provided live pairwise feedback on on-policy generations over $\sim$4 hours (3,200 image pairs generated using HPDv2~\cite{wu2023_hpsv2} prompts).
We used neither reward models nor offline preference data.
Fig.~\ref{figDHRF} shows the model's samples across 50 rounds of reward collection and training (epochs).
This was possible because our method combines three key properties:
  high sample efficiency (4h would have been nowhere near enough for FlowGRPO-like methods),
  pairwise comparison (update direction from a single human-provided preference),
  and support for non-differentiable rewards (human preferences).

\section{Discussion and Future Work}
\label{sec:discussion}

Our method (FDFO) presents a step forward in RL post-training of diffusion models.
It is a direct replacement to Flow-GRPO with faster convergence and higher-quality results. 
By leaving the multi-step MDP formulation, we shorten the reward attribution horizon and expand the algorithmic design space. 

Our evaluation is focused on comparing RL algorithms with as few confounding factors as possible,
  and thus we disabled KL regularization in our main experiments.
However, it remains compatible with our algorithm, as demonstrated in further experiments in Appendix~\refAppResults{}.
KL regularization is typically used to better retain variation in the results,
  but the best way to do this in general remains an open problem.
Finding a way to formulate a VLM-based reward component that targets diversity directly
  could be an interesting avenue for future work.

\ifacks
  \paragraph{Acknowledgments}
  We extend our thanks to Tero Kuosmanen and Samuel Klenberg for maintaining our compute infrastructure.
\fi

{
\ifconf
  \bibliographystyle{splncs04}
\else
  \small
  \bibliographystyle{ieeenat_fullname}
\fi
\bibliography{paper}
}

\ifappendix

\newcommand{\refAlgStochastic}{\ref{alg:stochastic}}
\newcommand{\refSecRewardmax}{\ref{sec:rewardmax}}
\newcommand{\refSecAnalysis}{\ref{sec:analysis}}
\newcommand{\refSecImplementation}{\ref{sec:implementation}}
\newcommand{\refSecWallclock}{\ref{sec:wallclock}}
\newcommand{\refFigPerformancePlot}{\ref{figPerformancePlot}}
\newcommand{\refFigRLWorksPlot}{\ref{figRLWorksPlot}}
\newcommand{\refSecDHRF}{\ref{sec:dhrf}}

\clearpage
\appendix
\input{appendix-figures}

\input{appendix-body}

\fi

\end{document}

%% file: macros.tex
\definecolor{olive}{rgb}{0.5, 0.5, 0.0}
\definecolor{maroon}{rgb}{0.69, 0.19, 0.38}
\definecolor{celestialblue}{rgb}{0.29, 0.59, 0.82}
\definecolor{darkgreen}{rgb}{0.0, 0.6, 0.0}
\definecolor{grey}{rgb}{0.5,0.5,0.5}
\definecolor{darkblue}{rgb}{0.19, 0.19, 0.62}
\definecolor{silver}{rgb}{0.7,0.7,0.7}
\definecolor{darkcyan}{rgb}{0.0, 0.55, 0.55}

\def\clap#1{\hbox to 0pt{\hss #1\hss}}%

\ifconf
  \newsavebox{\foobox}
  \newcommand{\slantbox}[2][.5]{\mbox{\sbox{\foobox}{#2}\hskip\wd\foobox\pdfsave\pdfsetmatrix{1 0 #1 1}\llap{\usebox{\foobox}}\pdfrestore}}
  \let\StandardDelta\Delta
  \renewcommand{\Delta}{\slantbox[-.25]{$\StandardDelta$}}
\fi

\newcommand{\tstrut}{\vphantom{$\hat{A}$}}

\newcommand{\XX}{\boldsymbol{X}}

\newcommand{\boldzero}{\mathbf{0}}
\newcommand{\boldI}{\mathbf{I}}
\newcommand{\boldx}{\mathbf{x}}

\newcommand{\boldv}{\mathbf{v}}

\newcommand{\boldc}{\mathbf{c}}

\newcommand{\boldJ}{\mathbf{J}}

\newcommand{\boldxhat}{\boldsymbol{\hat{\mathbf{x}}}}
\newcommand{\boldvhat}{\boldsymbol{\hat{\mathbf{v}}}}
\newcommand{\boldxtilde}{\boldsymbol{\tilde{\mathbf{x}}}}

\newcommand{\DeltaR}{\Delta\hspace*{-.03em}R}
\newcommand{\Deltax}{\Delta\boldx}
\newcommand{\nDeltax}{\smash{\overline{\Deltax}}}

\newcommand{\hnabla}{\nabla\hspace*{-.05em}}

\newcommand{\getsu}{\gets S \cup}

\newcommand\undefcolumntype[1]{\expandafter\let\csname NC@find@#1\endcsname\relax}

\newcommand{\s}{\hphantom{0}}

\DeclareMathOperator{\argmax}{arg\,max}

\newcommand{\Qx}{}
\newcommand{\Qy}{\hspace*{1.25em}}
\newcommand{\Q}{\Qx}
\newcommand{\QQ}{\Qx\Qy}
\newcommand{\QQQ}{\Qx\Qy\Qy}
\newcommand{\QQQQ}{\Qx\Qy\Qy\Qy}

\newcommand{\vel}{v_\theta}
\newcommand{\velnaked}{v}
\newcommand{\condemb}{\boldc}
\newcommand{\trainingprompts}{C}
\newcommand{\reward}{R}
\newcommand{\Reward}{R}

\newcommand{\flow}{f}
\newcommand{\flowtheta}{\flow_\theta}

\newcommand{\jac}{\mathbf{J}}
\newcommand{\jachi}[3]{\jac\raisebox{#3}{${}^{#1}_{#2}$}}
\newcommand{\jactf}{{\textstyle\jachi{\theta}{\hspace*{-.1em}\flow}{.2ex}}}
\newcommand{\jactv}{{\textstyle\jac^\theta{}_{\hspace*{-.47em}\velnaked}}}
\newcommand{\jacxfi}{\jachi{\boldx}{\!\flow_i}{.2ex}}
\newcommand{\jacxfj}{\jachi{\boldx}{\!\flow_j}{.2ex}}
\newcommand{\noisedraw}{\boldsymbol{\epsilon}}
\newcommand{\pertx}{\boldx'}
\newcommand{\lnormal}{\mathcal{LN}}
\newcommand{\transp}{{}^{\!\top}}

\newcommand{\atphantom}{\vphantom{${}^2$}}

\newcommand{\datalabel}[1]{\scalebox{0.9}{\scriptsize #1}}
\pgfplotsset{tick label style={font=\scriptsize}}
\pgfplotsset{legend style={font=\scriptsize}}
\pgfplotsset{tick style={draw=none}}
\pgfplotsset{major grid style={gray!40}}
\pgfplotsset{every axis plot/.append style={line width=0.6pt, mark size=1.1pt}}
\pgfplotsset{legend image code/.code={\draw[mark repeat=2, mark phase=2] plot coordinates {(0mm, 0.2mm) (1.5mm, 0.2mm) (3mm, 0.2mm)};}}

\definecolor{C0}{rgb}{0.121569, 0.466667, 0.705882}
\definecolor{C1}{rgb}{1.000000, 0.498039, 0.054902}
\definecolor{C2}{rgb}{0.172549, 0.627451, 0.172549}
\definecolor{C3}{rgb}{0.839216, 0.152941, 0.156863}
\definecolor{C4}{rgb}{0.580392, 0.403922, 0.741176}
\definecolor{C5}{rgb}{0.549020, 0.337255, 0.294118}
\definecolor{C6}{rgb}{0.890196, 0.466667, 0.760784}
\definecolor{C7}{rgb}{0.498039, 0.498039, 0.498039}
\definecolor{C8}{rgb}{0.737255, 0.741176, 0.133333}
\definecolor{C9}{rgb}{0.090196, 0.745098, 0.811765}
\definecolor{C10}{rgb}{1,1,1}
\definecolor{C11}{rgb}{0,0,0}

\newcommand{\hs}[1]{\hspace{#1mm}}

\newcommand{\placeholderfig}[3]{\makebox[#1][c]{\begin{tcolorbox}[width=#1,height=#2,colback=gray!10,colframe=gray!30,halign=center,valign=center]\color{gray!70}#3\end{tcolorbox}}}

\newcommand{\thickfbox}[1]{\begingroup\setlength{\fboxrule}{0.1mm}\setlength{\fboxsep}{-\fboxrule}\fbox{#1}\endgroup}
\newcommand{\eval}[2][2]{\pgfkeys{/pgf/fpu}\pgfmathparse{#2}\pgfmathprintnumber[fixed, zerofill, precision=#1, set thousands separator={}]{\pgfmathresult}}
\newcommand{\evaltext}[2][2]{\pgfkeys{/pgf/fpu}\pgfmathparse{#2}\pgfmathprintnumber[fixed, zerofill, precision=#1, set thousands separator={}, verbatim, assume math mode]{\pgfmathresult}}

\newcounter{ArrayIndex}

\newcommand\ArrayDefine[2]{\expandafter\newarray\csname #1\endcsname\readarray{#1}{#2}}

\newcommand\ArrayLength[1]{\csname total@#1\endcsname}

\newcommand\ArrayLookup[3]{\csname check#2\endcsname(#3)\xdef#1{\cachedata}}

\newcommand\ArrayMin[2]{%
  \xdef#1{1e4}%
  \setcounter{ArrayIndex}{0}%
  \whiledo{\value{ArrayIndex}<\ArrayLength{#2}}{%
    \stepcounter{ArrayIndex}%
    \ArrayLookup{\X}{#2}{\theArrayIndex}%
    \ifx\X\empty\else%
      \pgfmathparse{min(#1, \X)}%
      \xdef#1{\pgfmathresult}%
    \fi%
  }%
}

\newcommand\ArrayMax[2]{%
  \xdef#1{-1e4}%
  \setcounter{ArrayIndex}{0}%
  \whiledo{\value{ArrayIndex}<\ArrayLength{#2}}{%
    \stepcounter{ArrayIndex}%
    \ArrayLookup{\X}{#2}{\theArrayIndex}%
    \ifx\X\empty\else%
      \pgfmathparse{max(#1, \X)}%
      \xdef#1{\pgfmathresult}%
    \fi%
  }%
}

\newcommand\ArrayXYCombine[4]{%
  \xdef#1{}%
  \setcounter{ArrayIndex}{0}%
  \whiledo{\value{ArrayIndex}<\ArrayLength{#2}}{%
    \stepcounter{ArrayIndex}%
    \ArrayLookup{\X}{#2}{\theArrayIndex}%
    \ArrayLookup{\Y}{#3}{\theArrayIndex}%
    \ifx\X\empty\else\ifx\Y\empty\else%
      \xappto#1{#4}%
    \fi\fi%
  }%
}

\newcommand\ArrayXYLookup[4]{%
  \xdef#1{}%
  \setcounter{ArrayIndex}{0}%
  \whiledo{\value{ArrayIndex}<\ArrayLength{#2}}{%
    \stepcounter{ArrayIndex}%
    \ArrayLookup{\X}{#2}{\theArrayIndex}%
    \ifx\X#4%
      \ArrayLookup{\Y}{#3}{\theArrayIndex}%
      \xdef#1{\Y}%
    \fi%
  }%
}

\newcommand\PlotCurve[4]{%
  \ArrayXYCombine{\Coords}{#2}{#3}{#4}%
  \addplot[#1] coordinates {\Coords};%
}
\newcommand\PlotCurveDashed[4]{\PlotCurve{#1, line cap=round, dash pattern=on 1pt off 2pt, line width=.8pt}{#2}{#3}{#4}}

\newcommand\PlotDataLabelAtMax[8]{%
  \ArrayMax{#5}{#3}%
  \ArrayXYLookup{#4}{#3}{#2}{#5}%
  \addplot[#1, mark=*, forget plot, nodes near coords align=#7, nodes near coords=#8] coordinates {#6};%
}

\newcommand\PlotDataLabelAtX[8]{%
  \ArrayXYLookup{#5}{#2}{#3}{#4}%
  \addplot[#1, mark=*, forget plot, nodes near coords align=#7, nodes near coords=#8] coordinates {#6};%
}

\newcommand\MinorLine[1]{%
  \addplot[black, opacity=0.08, thin, forget plot] coordinates {#1};%
}

\newcommand\Image[2]{%
  \thickfbox{\includegraphics[width=#1]{images/#2}}%
}

\newcommand\ImageWithLabel[3]{%
  \thickfbox{\includegraphics[width=#1]{images/#2}}%
  \makebox(0,0)[r]{\raisebox{4.5mm}{\makebox[#1][r]{\color{white}\contourlength{0.25mm}\contour{black}{#3\hs{1}}}}}%
}

\newcommand\ImageRowTitle[2]{%
  \rotatebox{90}{\makebox[#1][c]{#2}}\hs{1}%
}

\newcommand{\commbox}{\makebox[29mm][l]}
\ifconf
  \newcommand{\commboxx}{\makebox[35mm][l]}
\else
  \newcommand{\commboxx}{\makebox[27.6mm][l]}
\fi
\newcommand{\commcolor}{\textcolor{gray}}
\newcommand{\commsymbol}{$\triangleright$}
\newcommand{\comm}[1]{\hfill\commbox{\commcolor{\commsymbol{} #1}}}
\newcommand{\comx}[1]{\hfill\commbox{\hphantom{\commsymbol} \commcolor{#1}}}
\newcommand{\commm}[1]{\hfill\commboxx{\commcolor{\commsymbol{} #1}}}
\newcommand{\commx}[1]{\hfill\commboxx{\hphantom{\commsymbol} \commcolor{#1}}}
\newcommand{\odetime}{t}

%% file: data/data-comparison.tex
\ArrayDefine{cmpFDFO::xk}{0&4.32&8.64&12.96&17.28&21.6&25.92&30.24&34.56&38.88&43.2&47.52&51.84&56.16&60.48&64.8&69.12&73.44&77.76&82.08&86.4&90.72&95.04&99.36&103.68&108&112.32&116.64&120.96&125.28&129.6&133.92&138.24&142.56&146.88&151.2&155.52&159.84&164.16&168.48&172.8&177.12&181.44&185.76&190.08&194.4&198.72&203.04&207.36&211.68&216&220.32&224.64&228.96&233.28&237.6&241.92&246.24&250.56&254.88&259.2&263.52&267.84&272.16&276.48&280.8&285.12&289.44&293.76&298.08&302.4&306.72&311.04&315.36&319.68&324&328.32&332.64&336.96&341.28&345.6&349.92&354.24&358.56&362.88&367.2&371.52&375.84&380.16&384.48&388.8&393.12&397.44&401.76&406.08&410.4&414.72&419.04&423.36&427.68&432&436.32&440.64&444.96&449.28&453.6&457.92&462.24&466.56&470.88&475.2&479.52&483.84&488.16&492.48&496.8&501.12&505.44&509.76&514.08&518.4&522.72&527.04&531.36&535.68&540&544.32&548.64&552.96&557.28&561.6&565.92&570.24&574.56&578.88&583.2&587.52&591.84&596.16&600.48&604.8&609.12&613.44&617.76&622.08&626.4&630.72&635.04&639.36&643.68&648&652.32&656.64&660.96&665.28&669.6&673.92&678.24&682.56&686.88&691.2&695.52&699.84&704.16&708.48&712.8&717.12&721.44&725.76&730.08&734.4&738.72&743.04&747.36&751.68&756&760.32&764.64&768.96&773.28&777.6&781.92&786.24&790.56&794.88&799.2&803.52&807.84&812.16&816.48&820.8&825.12&829.44&833.76&838.08&842.4&846.72&851.04&855.36&859.68&864}
\ArrayDefine{cmpFDFO::pickscore}{20.19&21&21.53&21.87&22.04&22.15&22.26&22.36&22.45&22.55&22.62&22.68&22.74&22.79&22.84&22.89&22.95&23&23.05&23.09&23.1&23.12&23.16&23.21&23.23&23.24&23.26&23.29&23.32&23.35&23.37&23.37&23.39&23.43&23.46&23.47&23.46&23.49&23.54&23.55&23.56&23.57&23.59&23.62&23.64&23.66&23.66&23.66&23.59&23.49&23.47&23.48&23.49&23.51&23.52&23.53&23.54&23.57&23.58&23.6&23.62&23.63&23.64&23.64&23.65&23.66&23.64&23.62&23.63&23.66&23.69&23.7&23.68&23.65&23.66&23.69&23.71&23.72&23.72&23.72&23.73&23.73&23.75&23.75&23.74&23.74&23.75&23.74&23.73&23.73&23.74&23.75&23.75&23.76&23.78&23.79&23.78&23.79&23.81&23.82&23.81&23.78&23.77&23.79&23.81&23.81&23.82&23.82&23.82&23.83&23.84&23.85&23.86&23.88&23.9&23.9&23.88&23.87&23.86&23.86&23.87&23.88&23.89&23.91&23.9&23.89&23.89&23.91&23.91&23.9&23.9&23.91&23.92&23.93&23.94&23.94&23.95&23.97&23.97&23.96&23.95&23.95&23.98&24.02&24.03&24.01&24&23.99&23.99&24.02&24.02&24.02&24&23.99&24&24&24.02&24.04&24.04&24.03&24.03&24.03&24.01&24.03&24.04&24.04&24.04&24.04&24.05&24.05&24.04&24.06&24.05&24.05&24.06&24.07&24.08&24.08&24.08&24.07&24.05&24.04&24.05&24.06&24.08&24.09&24.05&24.03&24.05&24.07&24.07&24.05&24.04&24.03&24.03&24.03&24.02&24.04&24.06&24.05&24.05}

\ArrayDefine{cmpFlowGRPOCfg::xk}{0&60&120&180&240&300&360&420}
\ArrayDefine{cmpFlowGRPOCfg::pickscore}{21.72&22.034&22.507&22.821&23.078&23.311&23.41&23.581}

\ArrayDefine{cmpOnlineDPOB100::xk}{0&60&120&180&240&300&360&420}
\ArrayDefine{cmpOnlineDPOB100::pickscore}{21.72&22.409&22.703&22.887&22.9&22.942&23.282&23.085}

\ArrayDefine{cmpFlowGRPONoCfg::xk}{0&4.32&8.64&12.96&17.28&21.6&25.92&30.24&34.56&38.88&43.2&47.52&51.84&56.16&60.48&64.8&69.12&73.44&77.76&82.08&86.4&90.72&95.04&99.36&103.68&108&112.32&116.64&120.96&125.28&129.6&133.92&138.24&142.56&146.88&151.2&155.52&159.84&164.16&168.48&172.8&177.12&181.44&185.76&190.08&194.4&198.72&203.04&207.36&211.68&216&220.32&224.64&228.96&233.28&237.6&241.92&246.24&250.56&254.88&259.2&263.52&267.84&272.16&276.48&280.8&285.12&289.44&293.76&298.08&302.4&306.72&311.04&315.36&319.68&324&328.32&332.64&336.96&341.28&345.6&349.92&354.24&358.56&362.88&367.2&371.52&375.84&380.16&384.48&388.8&393.12&397.44&401.76&406.08&410.4&414.72&419.04&423.36&427.68&432&436.32&440.64&444.96&449.28&453.6&457.92&462.24&466.56&470.88&475.2&479.52&483.84&488.16&492.48&496.8&501.12&505.44&509.76&514.08&518.4&522.72&527.04&531.36&535.68&540&544.32&548.64&552.96&557.28&561.6&565.92&570.24&574.56&578.88&583.2&587.52&591.84&596.16&600.48&604.8&609.12&613.44&617.76&622.08&626.4&630.72&635.04&639.36&643.68&648&652.32&656.64&660.96&665.28&669.6&673.92&678.24&682.56&686.88&691.2&695.52&699.84&704.16&708.48&712.8&717.12&721.44&725.76&730.08&734.4&738.72&743.04&747.36&751.68&756&760.32&764.64&768.96&773.28&777.6&781.92&786.24&790.56&794.88&799.2&803.52&807.84&812.16&816.48&820.8&825.12&829.44&833.76&838.08&842.4&846.72&851.04&855.36&859.68&864}
\ArrayDefine{cmpFlowGRPONoCfg::pickscore}{20.19&20.17&20.3&20.48&20.66&20.82&20.95&21.05&21.14&21.24&21.3&21.36&21.42&21.46&21.53&21.6&21.65&21.7&21.75&21.8&21.83&21.86&21.9&21.96&22.02&22.06&22.09&22.12&22.17&22.22&22.23&22.23&22.25&22.28&22.29&22.28&22.29&22.31&22.34&22.38&22.42&22.45&22.47&22.48&22.5&22.53&22.54&22.56&22.57&22.57&22.57&22.56&22.55&22.55&22.55&22.55&22.59&22.64&22.69&22.71&22.71&22.7&22.73&22.78&22.8&22.77&22.73&22.73&22.75&22.75&22.76&22.79&22.81&22.79&22.8&22.83&22.84&22.87&22.88&22.89&22.89&22.89&22.88&22.85&22.86&22.92&22.98&23&23.02&23.05&23.08&23.07&23.06&23.07&23.08&23.09&23.06&23.01&22.99&22.99&22.99&22.97&22.93&22.93&22.94&22.93&22.92&22.91&22.92&22.95&22.98&23&23.01&23.04&23.07&23.07&23.07&23.07&23.08&23.09&23.09&23.09&23.09&23.12&23.14&23.17&23.19&23.17&23.14&23.13&23.15&23.16&23.17&23.18&23.2&23.21&23.21&23.21&23.2&23.2&23.21&23.21&23.22&23.21&23.19&23.19&23.2&23.24&23.27&23.27&23.25&23.24&23.24&23.26&23.25&23.25&23.27&23.29&23.28&23.27&23.29&23.31&23.33&23.33&23.3&23.24&23.21&23.25&23.32&23.37&23.39&23.38&23.38&23.37&23.37&23.37&23.36&23.36&23.36&23.36&23.38&23.41&23.42&23.41&23.38&23.36&23.36&23.36&23.35&23.36&23.38&23.41&23.42&23.39&23.39&23.4&23.41&23.42&23.43&23.43&23.42}

\ArrayDefine{cmpReFL::xk}{0&23.04&69.12&92.16&115.2&161.28&184.32&207.36&253.44&276.48&299.52&391.68&414.72&437.76&483.84&506.88&529.92&552.96&599.04&622.08&645.12&691.2&714.24&737.28&760.32&783.36&829.44&852.48&875.52&880.704}
\ArrayDefine{cmpReFL::pickscore}{21.72&21.736&21.84&21.918&22.022&22.178&22.23&22.334&22.438&22.464&22.542&22.646&22.698&22.698&22.75&22.724&22.776&22.802&22.802&22.828&22.828&22.88&22.88&22.906&22.906&22.932&22.932&22.958&22.932&22.938}

\ArrayDefine{cmpOfflineDPOB100::xk}{0&60&120&180&240&300&360&420}
\ArrayDefine{cmpOfflineDPOB100::pickscore}{21.72&22.48&22.7&22.77&22.6&22.586&22.545&22.645}

\ArrayDefine{cmpDDPO::xk}{0&69.12&138.24&207.36&276.48&345.6&414.72&468.864}
\ArrayDefine{cmpDDPO::pickscore}{21.72&21.788&21.918&21.996&22.022&21.788&21.918&19.198}

\ArrayDefine{cmpOfflineSFT::xk}{0&60&120&180&240&300&360&420}
\ArrayDefine{cmpOfflineSFT::pickscore}{21.72&21.77&21.785&21.802&21.813&21.811&21.816&21.849}

\ArrayDefine{cmpOfflineRWR::xk}{0&60&120&180&240&300&360&420}
\ArrayDefine{cmpOfflineRWR::pickscore}{21.72&21.588&21.558&21.568&21.581&21.596&21.644&21.573}

\ArrayDefine{cmpOnlineDPOB10::xk}{0&60&120&180}
\ArrayDefine{cmpOnlineDPOB10::pickscore}{21.72&22.263&22.396&22.27}

\ArrayDefine{cmpOnlineSFT::xk}{0&60&120&180&240}
\ArrayDefine{cmpOnlineSFT::pickscore}{21.72&21.764&21.748&21.757&21.804}

%% file: figures.tex
\newcommand{\figArtifacts}{%
\begin{minipage}{1.0\linewidth}%
\gdef\h{0.245\linewidth}%
\makebox[\h]{\small Our method}\hfill%
\makebox[\h]{\small Flow-GRPO}\hfill%
\makebox[\h]{\small Our method}\hfill%
\makebox[\h]{\small Flow-GRPO}\\[.5ex]
\ImageWithLabel{\h}{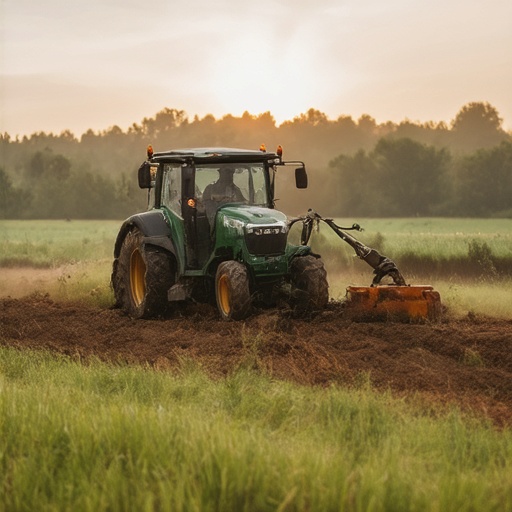}{\small$\mathbf{70}$}\hfill%
\ImageWithLabel{\h}{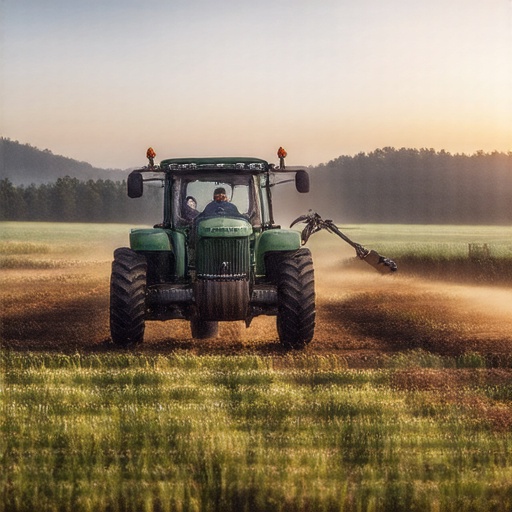}{\small$\mathbf{830}$}\hfill%
\ImageWithLabel{\h}{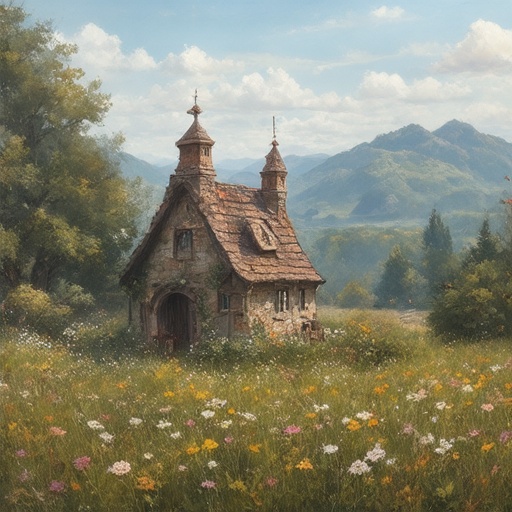}{\small$\mathbf{70}$}\hfill%
\ImageWithLabel{\h}{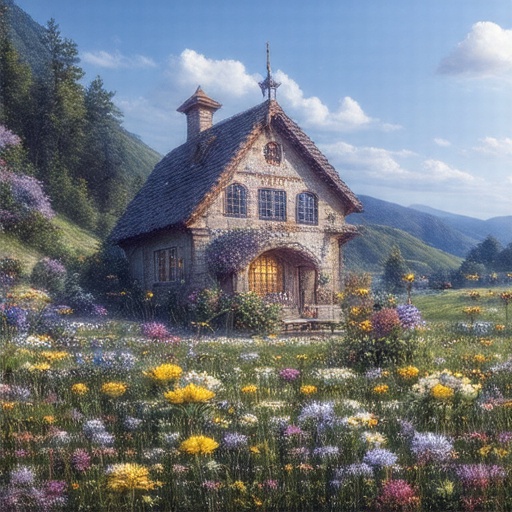}{\small$\mathbf{830}$}\\[-3ex]
\caption{\label{figArtifacts}%
When trained long enough, Flow-GRPO starts exhibiting grid-like artifacts that come and go, but are very disturbing in some epochs (here 830).
Our results do not show these at any stage; here epoch 70 serves as an iso-reward comparison.
Left:
\emph{``A green tractor plowing a field at sunrise.''}
Right:
\emph{``A quaint cottage nestled in a vibrant flower-filled meadow.''}
Combined reward, no CFG or KL regularization.
}%
\end{minipage}
}%

\newcommand{\figRLWorksPlotA}{%
\centering\footnotesize%
\begin{tikzpicture}%
\begin{axis}[
  width={1.17\columnwidth}, height={55mm}, grid={major},
  xmin={0}, xmax={1000}, xmode={linear}, xtick={0, 200, 400, 600, 800, 1000}, xticklabels={$0$, $200$, $400$, $600$, $800$, \hs{-2}$1$k},
  ymin={20}, ymax={24.5}, ymode={linear}, ytick={20, 21, 22, 23, 24}, yticklabels={\raisebox{0.2mm}[0mm][0mm]{$20$}, $21$, $22$, $23$, $24$},
  legend pos={south east}, legend cell align={left}, legend columns={1},
]
\gdef\xmid{epoch}\gdef\xmul{1}
\gdef\ymid{rc02-4096x1::pickscore_score}\gdef\ymul{100}
\gdef\pmx##1{\MinorLine{(##1,20)(##1,24.5)}}\pmx{100}\pmx{300}\pmx{500}\pmx{700}\pmx{900}
\gdef\pmy##1{\MinorLine{(0,##1)(1000,##1)}}\pmy{20.5}\pmy{21.5}\pmy{22.5}\pmy{23.5}

\addlegendentry{Our method}
\gdef\rid{rc09-kA-cA-rI-mH}
\PlotCurve{C0}{\rid::\xmid}{\rid::\ymid}{(\X*\xmul,\Y*\ymul)}

\addlegendentry{Flow-GRPO}
\gdef\rid{rc02-kA-cA-rI-mE}
\PlotCurve{C1}{\rid::\xmid}{\rid::\ymid}{(\X*\xmul,\Y*\ymul)}

\end{axis}
\end{tikzpicture}%
}%

\newcommand{\figRLWorksPlotB}{%
\centering\footnotesize%
\begin{tikzpicture}%
\begin{axis}[
  width={1.17\columnwidth}, height={55mm}, grid={major},
  xmin={0}, xmax={1000}, xmode={linear}, xtick={0, 200, 400, 600, 800, 1000}, xticklabels={$0$, $200$, $400$, $600$, $800$, \hs{-2}$1$k},
  ymin={25}, ymax={85}, ymode={linear}, ytick={30, 40, 50, 60, 70, 80}, yticklabels={$30$, $40$, $50$, $60$, $70$, $80$},
  legend pos={south east}, legend cell align={left}, legend columns={1},
]
\gdef\xmid{epoch}\gdef\xmul{1}
\gdef\ymid{rc02-4096x1::prompt_rB}\gdef\ymul{100}
\gdef\pmx##1{\MinorLine{(##1,25)(##1,85)}}\pmx{100}\pmx{300}\pmx{500}\pmx{700}\pmx{900}
\gdef\pmy##1{\MinorLine{(0,##1)(1000,##1)}}\pmy{35}\pmy{45}\pmy{55}\pmy{65}\pmy{75}

\addlegendentry{Our method}
\gdef\rid{rc05probe-kA-cA-rB-mH}
\PlotCurve{C0}{\rid::\xmid}{\rid::\ymid}{(\X*\xmul,\Y*\ymul)}

\addlegendentry{Flow-GRPO}
\gdef\rid{rc02-kA-cA-rB-mE}
\PlotCurve{C1}{\rid::\xmid}{\rid::\ymid}{(\X*\xmul,\Y*\ymul)}

\end{axis}
\end{tikzpicture}%
}%

\newcommand{\figRLWorksPlotC}{%
\centering\footnotesize%
\begin{tikzpicture}%
\begin{axis}[
  width={1.17\columnwidth}, height={55mm}, grid={major},
  xmin={0}, xmax={1000}, xmode={linear}, xtick={0, 200, 400, 600, 800, 1000}, xticklabels={$0$, $200$, $400$, $600$, $800$, \hs{-2}$1$k},
  ymin={22.5}, ymax={31.3}, ymode={linear}, ytick={23, 24, 25, 26, 27, 28, 29, 30, 31}, yticklabels={$23$, $24$, $25$, $26$, $27$, $28$, $29$, $30$, $31$},
  legend pos={south east}, legend cell align={left}, legend columns={1},
]
\gdef\xmid{epoch}\gdef\xmul{1}
\gdef\ymid{rc02-4096x1::combo_rK}\gdef\ymul{10}
\gdef\pmx##1{\MinorLine{(##1,22.5)(##1,31.3)}}\pmx{100}\pmx{300}\pmx{500}\pmx{700}\pmx{900}

\addlegendentry{Our method}
\gdef\rid{rc05probe-kA-cA-rK-mH}
\PlotCurve{C0}{\rid::\xmid}{\rid::\ymid}{(\X*\xmul,\Y*\ymul)}
\gdef\XOA{30}\PlotDataLabelAtX{C0}{\rid::\xmid}{\rid::\ymid}{\XOA}{\YOA}{(\XOA*\xmul,\YOA*\ymul)}{west}{\datalabel{\raisebox{-2mm}{\eval[1]{\YOA*\ymul}}}}
\gdef\XOB{60}\PlotDataLabelAtX{C0}{\rid::\xmid}{\rid::\ymid}{\XOB}{\YOB}{(\XOB*\xmul,\YOB*\ymul)}{west}{\datalabel{\raisebox{-2mm}{\eval[1]{\YOB*\ymul}}}}
\gdef\XOC{90}\PlotDataLabelAtX{C0}{\rid::\xmid}{\rid::\ymid}{\XOC}{\YOC}{(\XOC*\xmul,\YOC*\ymul)}{west}{\datalabel{\raisebox{-2mm}{\eval[1]{\YOC*\ymul}}}}
\gdef\XOD{600}\PlotDataLabelAtX{C0}{\rid::\xmid}{\rid::\ymid}{\XOD}{\YOD}{(\XOD*\xmul,\YOD*\ymul)}{south}{\datalabel{\eval[1]{\YOD*\ymul}}}

\addlegendentry{Flow-GRPO}
\gdef\rid{rc02-kA-cA-rK-mE}
\PlotCurve{C1}{\rid::\xmid}{\rid::\ymid}{(\X*\xmul,\Y*\ymul)}
\gdef\XFA{245}\PlotDataLabelAtX{C1}{\rid::\xmid}{\rid::\ymid}{\XFA}{\YFA}{(\XFA*\xmul,\YFA*\ymul)}{north}{\hs{1}\datalabel{\eval[1]{\YFA*\ymul}}}
\gdef\XFB{690}\PlotDataLabelAtX{C1}{\rid::\xmid}{\rid::\ymid}{\XFB}{\YFB}{(\XFB*\xmul,\YFB*\ymul)}{north}{\hs{1}\datalabel{\eval[1]{\YFB*\ymul}}}
\gdef\XFC{920}\PlotDataLabelAtX{C1}{\rid::\xmid}{\rid::\ymid}{\XFC}{\YFC}{(\XFC*\xmul,\YFC*\ymul)}{north}{\hs{1}\datalabel{\eval[1]{\YFC*\ymul}}}
\gdef\XFD{0}\PlotDataLabelAtX{C0!50!C1}{\rid::\xmid}{\rid::\ymid}{\XFD}{\YFD}{(\XFD*\xmul,\YFD*\ymul)}{west}{\datalabel{\eval[1]{\YFD*\ymul}}}

\end{axis}
\end{tikzpicture}%
}%

\ArrayMax{\FlowGRPOBest}{rc02-kA-cA-rI-mE::rc02-4096x1::pickscore_score}

\newcommand{\figRLWorksPlot}{%
\begin{minipage}{1.0\linewidth}%
\hspace*{-1mm}%
\begin{subfigure}[b]{0.33\linewidth}\figRLWorksPlotA\end{subfigure}\hfill%
\begin{subfigure}[b]{0.33\linewidth}\figRLWorksPlotB\end{subfigure}\hfill%
\begin{subfigure}[b]{0.33\linewidth}\figRLWorksPlotC\end{subfigure}\\[-.2ex]
\makebox[0.335\linewidth]{\small (a) PickScore reward}\hfill%
\makebox[0.330\linewidth]{\small (b) VLM alignment reward\ \ }\hfill%
\makebox[0.310\linewidth]{\small (c) Combined reward}%
\vspace*{-.5ex}%
\caption{\label{figRLWorksPlot}%
Comparison between our method and Flow-GRPO using different reward functions.
Each plot shows the reward as a function of RL epoch.
KL regularization and CFG are disabled.
Example images for the highlighted checkpoints of combined reward are shown visually in Fig.~\ref{figRLWorksImagesB}.
\textbf{(a)}
PickScore reward.
Flow-GRPO reaches \eval[2]{\FlowGRPOBest*100} after 1000 epochs (2000 training steps).
\textbf{(b)}
Our VLM alignment reward: \textit{Does this image match the caption "$\,\cdots$"? Answer Yes or No.}
\textbf{(c)}
Combined reward: PickScore $+$ VLM alignment $/~10$.
}%
\end{minipage}
}%

\newcommand{\figRLWorksImagesB}{%
\begin{minipage}{1.0\linewidth}%
\gdef\h{0.191\linewidth}%
\gdef\CO##1{\ImageWithLabel{\h}{rc05probe-kA-cA-rK-mH/e##1-p015-s04.jpg}{\small$\mathbf{\eval[0]{##1}}$}}%
\gdef\CF##1{\ImageWithLabel{\h}{rc02-kA-cA-rK-mE/e##1-p015-s04.jpg}{\small$\mathbf{\eval[0]{##1}}$}}%
\ImageRowTitle{\h}{\small Our method}\CO{0000}\hfill\CO{0030}\hfill\CO{0060}\hfill\CO{0090}\hfill\CO{0600}\\[1mm]
\ImageRowTitle{\h}{\small Flow-GRPO}\CF{0000}\hfill\CF{0245}\hfill\CF{0690}\hfill\CF{0920}\hfill\placeholderfig{\h}{\h}{\LARGE N/A}%
\vspace{-.5ex}%
\caption{\label{figRLWorksImagesB}%
Visual results for the combined reward using the prompt
\textit{``A girl with pigtails is holding a giant sunflower.''}
The epoch numbers (bottom-right corners) correspond to the dots in Fig.~\ref{figRLWorksPlot}c, approximately matching the reward between the methods.
The last column: Flow-GRPO is unable to increase the reward enough to match our method.
}%
\end{minipage}
}%

\newcommand{\figMetricsPlotA}{%
\centering\footnotesize%
\begin{tikzpicture}%
\begin{axis}[
  width={1.17\columnwidth}, height={55mm}, grid={major},
  xmin={0}, xmax={1000}, xmode={linear}, xtick={0, 200, 400, 600, 800, 1000}, xticklabels={$0$, $200$, $400$, $600$, $800$, \hs{-2}$1$k},
  ymin={53}, ymax={85}, ymode={linear}, ytick={55, 60, 65, 70, 75, 80, 85}, yticklabels={$55$, $60$, $65$, $70$, $75$, $80$, \raisebox{-1.6mm}[0mm][0mm]{$85$}},
  legend pos={south east}, legend cell align={left}, legend columns={2}, legend style={nodes={scale=0.95, transform shape}, row sep=-1.25pt},
]
\gdef\xmid{epoch}\gdef\xmul{1}
\gdef\ymid{oneigbench-1120x4::alignment}\gdef\ymul{100}
\gdef\pmx##1{\MinorLine{(##1,53)(##1,85)}}\pmx{100}\pmx{300}\pmx{500}\pmx{700}\pmx{900}

\addlegendimage{empty legend}
\addlegendentry{\hspace*{0.5em}Ours}
\addlegendimage{empty legend}
\addlegendentry{\hspace*{-1.5em}Flow-GRPO}

\addlegendentry{combined}
\gdef\rid{rc05probe-kA-cA-rK-mH}
\PlotCurve{C0}{\rid::\xmid}{\rid::\ymid}{(\X*\xmul,\Y*\ymul)}

\addlegendentry{combined$\!\!$}
\gdef\rid{rc02-kA-cA-rK-mE}
\PlotCurve{C1}{\rid::\xmid}{\rid::\ymid}{(\X*\xmul,\Y*\ymul)}

\addlegendentry{VLM}
\gdef\rid{rc05probe-kA-cA-rB-mH}
\PlotCurve{C0!50!white}{\rid::\xmid}{\rid::\ymid}{(\X*\xmul,\Y*\ymul)}

\addlegendentry{VLM}
\gdef\rid{rc02-kA-cA-rB-mE}
\PlotCurve{C1!50!white}{\rid::\xmid}{\rid::\ymid}{(\X*\xmul,\Y*\ymul)}

\addlegendentry{PickScore}
\gdef\rid{rc09-kA-cA-rI-mH}
\PlotCurveDashed{C0!80!white}{\rid::\xmid}{\rid::\ymid}{(\X*\xmul,\Y*\ymul)}

\addlegendentry{PickScore}
\gdef\rid{rc02-kA-cA-rI-mE}
\PlotCurveDashed{C1!80!white}{\rid::\xmid}{\rid::\ymid}{(\X*\xmul,\Y*\ymul)}

\end{axis}
\end{tikzpicture}%
}%

\newcommand{\figMetricsPlotB}{%
\centering\footnotesize%
\begin{tikzpicture}%
\begin{axis}[
  width={1.17\columnwidth}, height={55mm}, grid={major},
  xmin={0}, xmax={1000}, xmode={linear}, xtick={0, 200, 400, 600, 800, 1000}, xticklabels={$0$, $200$, $400$, $600$, $800$, \hs{-2}$1$k},
  ymin={25}, ymax={29.5}, ymode={linear}, ytick={25, 26, 27, 28, 29}, yticklabels={\raisebox{2mm}{$25$}, $26$, $27$, $28$, $29$},
  legend pos={south east}, legend cell align={left}, legend columns={2}, legend style={nodes={scale=0.85, transform shape}},
]
\gdef\xmid{epoch}\gdef\xmul{1}
\gdef\ymid{hpsv2-3200x4::hpsv2_score}\gdef\ymul{100}
\gdef\pmx##1{\MinorLine{(##1,25)(##1,29.5)}}\pmx{100}\pmx{300}\pmx{500}\pmx{700}\pmx{900}
\gdef\pmy##1{\MinorLine{(0,##1)(1000,##1)}}\pmy{25.5}\pmy{26.5}\pmy{27.5}\pmy{28.5}

\gdef\rid{rc05probe-kA-cA-rK-mH}
\PlotCurve{C0}{\rid::\xmid}{\rid::\ymid}{(\X*\xmul,\Y*\ymul)}

\gdef\rid{rc02-kA-cA-rK-mE}
\PlotCurve{C1}{\rid::\xmid}{\rid::\ymid}{(\X*\xmul,\Y*\ymul)}

\gdef\rid{rc05probe-kA-cA-rB-mH}
\PlotCurve{C0!50!white}{\rid::\xmid}{\rid::\ymid}{(\X*\xmul,\Y*\ymul)}

\gdef\rid{rc02-kA-cA-rB-mE}
\PlotCurve{C1!50!white}{\rid::\xmid}{\rid::\ymid}{(\X*\xmul,\Y*\ymul)}

\gdef\rid{rc09-kA-cA-rI-mH}
\PlotCurveDashed{C0!80!white}{\rid::\xmid}{\rid::\ymid}{(\X*\xmul,\Y*\ymul)}

\gdef\rid{rc02-kA-cA-rI-mE}
\PlotCurveDashed{C1!80!white}{\rid::\xmid}{\rid::\ymid}{(\X*\xmul,\Y*\ymul)}

\end{axis}
\end{tikzpicture}%
}%

\newcommand{\figMetricsPlotC}{%
\centering\footnotesize%
\begin{tikzpicture}%
\begin{axis}[
  width={1.17\columnwidth}, height={55mm}, grid={major},
  xmin={0}, xmax={1000}, xmode={linear}, xtick={0, 200, 400, 600, 800, 1000}, xticklabels={$0$, $200$, $400$, $600$, $800$, \hs{-2}$1$k},
  ymin={10}, ymax={53}, ymode={linear}, ytick={10, 20, 30, 40, 50}, yticklabels={\raisebox{2mm}{$10$}, $20$, $30$, $40$, $50$},
  legend pos={south east}, legend cell align={left}, legend columns={2}, legend style={nodes={scale=0.85, transform shape}},
]
\gdef\xmid{epoch}\gdef\xmul{1}
\gdef\ymid{oneigbench-1120x4::diversity}\gdef\ymul{100}
\gdef\pmx##1{\MinorLine{(##1,10)(##1,53)}}\pmx{100}\pmx{300}\pmx{500}\pmx{700}\pmx{900}
\gdef\pmy##1{\MinorLine{(0,##1)(1000,##1)}}\pmy{15}\pmy{25}\pmy{35}\pmy{45}

\gdef\rid{rc05probe-kA-cA-rK-mH}
\PlotCurve{C0}{\rid::\xmid}{\rid::\ymid}{(\X*\xmul,\Y*\ymul)}

\gdef\rid{rc02-kA-cA-rK-mE}
\PlotCurve{C1}{\rid::\xmid}{\rid::\ymid}{(\X*\xmul,\Y*\ymul)}

\gdef\rid{rc05probe-kA-cA-rB-mH}
\PlotCurve{C0!50!white}{\rid::\xmid}{\rid::\ymid}{(\X*\xmul,\Y*\ymul)}

\gdef\rid{rc02-kA-cA-rB-mE}
\PlotCurve{C1!50!white}{\rid::\xmid}{\rid::\ymid}{(\X*\xmul,\Y*\ymul)}

\gdef\rid{rc09-kA-cA-rI-mH}
\PlotCurveDashed{C0!80!white}{\rid::\xmid}{\rid::\ymid}{(\X*\xmul,\Y*\ymul)}

\gdef\rid{rc02-kA-cA-rI-mE}
\PlotCurveDashed{C1!80!white}{\rid::\xmid}{\rid::\ymid}{(\X*\xmul,\Y*\ymul)}

\end{axis}
\end{tikzpicture}%
}%

\newcommand{\figMetricsPlot}{%
\begin{minipage}{1.0\linewidth}%
\hspace*{-1mm}%
\begin{subfigure}[b]{0.33\linewidth}\figMetricsPlotA\end{subfigure}\hfill%
\begin{subfigure}[b]{0.33\linewidth}\figMetricsPlotB\end{subfigure}\hfill%
\begin{subfigure}[b]{0.33\linewidth}\figMetricsPlotC\end{subfigure}\\[-.2ex]
\makebox[0.335\linewidth]{\small (a) OneIG-Bench prompt alignment}\hfill%
\makebox[0.330\linewidth]{\small (b) HPSv2\ \ }\hfill%
\makebox[0.310\linewidth]{\small (c) OneIG-Bench diversity}%
\vspace*{-.5ex}%
\caption{\label{figMetricsPlot}%
Control metrics for our three rewards (combined, VLM, PickScore) as a function of RL epoch.
\textbf{(a)} Independent assessment of prompt alignment.
\textbf{(b)} Independent assessment of human preferences.
\textbf{(c)} Independent assessment of result diversity.
}%
\end{minipage}%
}%

\newcommand{\figGuidancePlotA}{%
\centering\footnotesize%
\begin{tikzpicture}%
\begin{axis}[
  width={1.17\columnwidth}, height={55mm}, grid={major},
  xmin={0}, xmax={1000}, xmode={linear}, xtick={0, 200, 400, 600, 800, 1000}, xticklabels={$0$, $200$, $400$, $600$, $800$, \hs{-2}$1$k},
  ymin={56}, ymax={88}, ymode={linear}, ytick={60, 65, 70, 75, 80, 85}, yticklabels={$60$, $65$, $70$, $75$, $80$, $85$},
  legend pos={south east}, legend cell align={left}, legend columns={2}, legend style={nodes={scale=0.95, transform shape}, row sep=-1.25pt},
]
\gdef\xmid{epoch}\gdef\xmul{1}
\gdef\ymid{oneigbench-1120x4::alignment}\gdef\ymul{100}
\gdef\pmx##1{\MinorLine{(##1,56)(##1,88)}}\pmx{100}\pmx{300}\pmx{500}\pmx{700}\pmx{900}

\addlegendimage{empty legend}
\addlegendentry{\hspace*{-0.6em}Ours}
\addlegendimage{empty legend}
\addlegendentry{\hspace*{-2.2em}Flow-GRPO}

\addlegendentry{$w{=}1$}
\gdef\rid{rc05probe-kA-cA-rK-mH}
\PlotCurve{C0}{\rid::\xmid}{\rid::\ymid}{(\X*\xmul,\Y*\ymul)}
\gdef\XNA{0}\PlotDataLabelAtX{C0}{\rid::\xmid}{\rid::\ymid}{\XNA}{\YNA}{(\XNA*\xmul,\YNA*\ymul)}{west}{}%
\gdef\XNB{80}\PlotDataLabelAtX{C0}{\rid::\xmid}{\rid::\ymid}{\XNB}{\YNB}{(\XNB*\xmul,\YNB*\ymul)}{west}{}%

\addlegendentry{$w{=}1$}
\gdef\rid{rc02-kA-cA-rK-mE}
\PlotCurve{C1}{\rid::\xmid}{\rid::\ymid}{(\X*\xmul,\Y*\ymul)}

\addlegendentry{$w{=}2$}
\gdef\rid{rc05probe-kA-cB-rK-mH}
\PlotCurve{C0!50!white}{\rid::\xmid}{\rid::\ymid}{(\X*\xmul,\Y*\ymul)}

\addlegendentry{$w{=}2$}
\gdef\rid{rc02-kA-cB-rK-mE}
\PlotCurve{C1!50!white}{\rid::\xmid}{\rid::\ymid}{(\X*\xmul,\Y*\ymul)}

\addlegendentry{$w{=}4.5$}
\gdef\rid{rc05probe-kA-cC-rK-mH}
\PlotCurveDashed{C0!80!white}{\rid::\xmid}{\rid::\ymid}{(\X*\xmul,\Y*\ymul)}
\gdef\XCA{0}\PlotDataLabelAtX{C0!70!white}{\rid::\xmid}{\rid::\ymid}{\XCA}{\YCA}{(\XCA*\xmul,\YCA*\ymul)}{west}{}%
\gdef\XCB{80}\PlotDataLabelAtX{C0!70!white}{\rid::\xmid}{\rid::\ymid}{\XCB}{\YCB}{(\XCB*\xmul,\YCB*\ymul)}{west}{}%

\addlegendentry{$w{=}4.5$}
\gdef\rid{rc02-kA-cC-rK-mE}
\PlotCurveDashed{C1!80!white}{\rid::\xmid}{\rid::\ymid}{(\X*\xmul,\Y*\ymul)}

\end{axis}
\end{tikzpicture}%
}%

\newcommand{\figGuidancePlotB}{%
\centering\footnotesize%
\begin{tikzpicture}%
\begin{axis}[
  width={1.17\columnwidth}, height={55mm}, grid={major},
  xmin={0}, xmax={1000}, xmode={linear}, xtick={0, 200, 400, 600, 800, 1000}, xticklabels={$0$, $200$, $400$, $600$, $800$, \hs{-2}$1$k},
  ymin={25}, ymax={30}, ymode={linear}, ytick={25, 26, 27, 28, 29, 30}, yticklabels={\raisebox{2mm}{$25$}, $26$, $27$, $28$, $29$, \raisebox{-1.6mm}[0mm][0mm]{$30$}},
  legend pos={south east}, legend cell align={left}, legend columns={2}, legend style={nodes={scale=0.85, transform shape}},
]
\gdef\xmid{epoch}\gdef\xmul{1}
\gdef\ymid{hpsv2-3200x4::hpsv2_score}\gdef\ymul{100}
\gdef\pmx##1{\MinorLine{(##1,25)(##1,30)}}\pmx{100}\pmx{300}\pmx{500}\pmx{700}\pmx{900}
\gdef\pmy##1{\MinorLine{(0,##1)(1000,##1)}}\pmy{25.5}\pmy{26.5}\pmy{27.5}\pmy{28.5}\pmy{29.5}

\gdef\rid{rc05probe-kA-cA-rK-mH}
\PlotCurve{C0}{\rid::\xmid}{\rid::\ymid}{(\X*\xmul,\Y*\ymul)}
\gdef\XNA{0}\PlotDataLabelAtX{C0}{\rid::\xmid}{\rid::\ymid}{\XNA}{\YNA}{(\XNA*\xmul,\YNA*\ymul)}{west}{}%
\gdef\XNB{80}\PlotDataLabelAtX{C0}{\rid::\xmid}{\rid::\ymid}{\XNB}{\YNB}{(\XNB*\xmul,\YNB*\ymul)}{west}{}%

\gdef\rid{rc02-kA-cA-rK-mE}
\PlotCurve{C1}{\rid::\xmid}{\rid::\ymid}{(\X*\xmul,\Y*\ymul)}

\gdef\rid{rc05probe-kA-cB-rK-mH}
\PlotCurve{C0!50!white}{\rid::\xmid}{\rid::\ymid}{(\X*\xmul,\Y*\ymul)}

\gdef\rid{rc02-kA-cB-rK-mE}
\PlotCurve{C1!50!white}{\rid::\xmid}{\rid::\ymid}{(\X*\xmul,\Y*\ymul)}

\gdef\rid{rc05probe-kA-cC-rK-mH}
\PlotCurveDashed{C0!80!white}{\rid::\xmid}{\rid::\ymid}{(\X*\xmul,\Y*\ymul)}
\gdef\XCA{0}\PlotDataLabelAtX{C0!70!white}{\rid::\xmid}{\rid::\ymid}{\XCA}{\YCA}{(\XCA*\xmul,\YCA*\ymul)}{west}{}%
\gdef\XCB{80}\PlotDataLabelAtX{C0!70!white}{\rid::\xmid}{\rid::\ymid}{\XCB}{\YCB}{(\XCB*\xmul,\YCB*\ymul)}{west}{}%

\gdef\rid{rc02-kA-cC-rK-mE}
\PlotCurveDashed{C1!80!white}{\rid::\xmid}{\rid::\ymid}{(\X*\xmul,\Y*\ymul)}

\end{axis}
\end{tikzpicture}%
}%

\newcommand{\figGuidancePlotC}{%
\centering\footnotesize%
\begin{tikzpicture}%
\begin{axis}[
  width={1.17\columnwidth}, height={55mm}, grid={major},
  xmin={0}, xmax={1000}, xmode={linear}, xtick={0, 200, 400, 600, 800, 1000}, xticklabels={$0$, $200$, $400$, $600$, $800$, \hs{-2}$1$k},
  ymin={10}, ymax={53}, ymode={linear}, ytick={10, 20, 30, 40, 50}, yticklabels={\raisebox{2mm}{$10$}, $20$, $30$, $40$, $50$},
  legend pos={south east}, legend cell align={left}, legend columns={2}, legend style={nodes={scale=0.85, transform shape}},
]
\gdef\xmid{epoch}\gdef\xmul{1}
\gdef\ymid{oneigbench-1120x4::diversity}\gdef\ymul{100}
\gdef\pmx##1{\MinorLine{(##1,10)(##1,53)}}\pmx{100}\pmx{300}\pmx{500}\pmx{700}\pmx{900}
\gdef\pmy##1{\MinorLine{(0,##1)(1000,##1)}}\pmy{15}\pmy{25}\pmy{35}\pmy{45}

\gdef\rid{rc05probe-kA-cA-rK-mH}
\PlotCurve{C0}{\rid::\xmid}{\rid::\ymid}{(\X*\xmul,\Y*\ymul)}
\gdef\XNA{0}\PlotDataLabelAtX{C0}{\rid::\xmid}{\rid::\ymid}{\XNA}{\YNA}{(\XNA*\xmul,\YNA*\ymul)}{west}{}%
\gdef\XNB{80}\PlotDataLabelAtX{C0}{\rid::\xmid}{\rid::\ymid}{\XNB}{\YNB}{(\XNB*\xmul,\YNB*\ymul)}{west}{}%

\gdef\rid{rc02-kA-cA-rK-mE}
\PlotCurve{C1}{\rid::\xmid}{\rid::\ymid}{(\X*\xmul,\Y*\ymul)}

\gdef\rid{rc05probe-kA-cB-rK-mH}
\PlotCurve{C0!50!white}{\rid::\xmid}{\rid::\ymid}{(\X*\xmul,\Y*\ymul)}

\gdef\rid{rc02-kA-cB-rK-mE}
\PlotCurve{C1!50!white}{\rid::\xmid}{\rid::\ymid}{(\X*\xmul,\Y*\ymul)}

\gdef\rid{rc05probe-kA-cC-rK-mH}
\PlotCurveDashed{C0!80!white}{\rid::\xmid}{\rid::\ymid}{(\X*\xmul,\Y*\ymul)}
\gdef\XCA{0}\PlotDataLabelAtX{C0!70!white}{\rid::\xmid}{\rid::\ymid}{\XCA}{\YCA}{(\XCA*\xmul,\YCA*\ymul)}{west}{}%
\gdef\XCB{80}\PlotDataLabelAtX{C0!70!white}{\rid::\xmid}{\rid::\ymid}{\XCB}{\YCB}{(\XCB*\xmul,\YCB*\ymul)}{west}{}%

\gdef\rid{rc02-kA-cC-rK-mE}
\PlotCurveDashed{C1!80!white}{\rid::\xmid}{\rid::\ymid}{(\X*\xmul,\Y*\ymul)}

\end{axis}
\end{tikzpicture}%
}%

\newcommand{\figGuidancePlot}{%
\begin{minipage}{1.0\linewidth}%
\hspace*{-1mm}%
\begin{subfigure}[b]{0.33\linewidth}\figGuidancePlotA\end{subfigure}\hfill%
\begin{subfigure}[b]{0.33\linewidth}\figGuidancePlotB\end{subfigure}\hfill%
\begin{subfigure}[b]{0.33\linewidth}\figGuidancePlotC\end{subfigure}\\[-.2ex]
\makebox[0.335\linewidth]{\small (a) OneIG-Bench prompt alignment}\hfill%
\makebox[0.330\linewidth]{\small (b) HPSv2\ \ }\hfill%
\makebox[0.310\linewidth]{\small (c) OneIG-Bench diversity}%
\vspace*{-.5ex}%
\caption{\label{figGuidancePlot}%
Control metrics without CFG ($w=1$), and with medium ($w=2$) and high ($w=4.5$) guidance strength as a function of RL epoch using the combined reward.
CFG drastically changes the results before post-training (epoch 0) but the gap narrows over time.
Results for the highlighted checkpoints are shown visually in Figure~\ref{figBiggerPictureImages}.
}%
\end{minipage}%
}%

\newcommand{\figBiggerPictureImages}{%
\begin{minipage}{1.0\linewidth}%
\gdef\h{0.49\linewidth}%
\gdef\hh{0.325\linewidth}%
\gdef\Subfig##1##2{\begin{subfigure}[b]{\h}%
  \Image{\hh}{##1-p125-s02.jpg}\hfill%
  \Image{\hh}{##1-p125-s05.jpg}\hfill%
  \Image{\hh}{##1-p125-s06.jpg}\\[-.4ex]%
  \centering\small{##2}
  \end{subfigure}}%
\Subfig{rc05probe-kA-cA-rK-mH/e0000}{(a) Original model, no CFG}\hfill\hfill%
\Subfig{rc05probe-kA-cC-rK-mH/e0000}{(b) Original model, CFG enabled ($w = 4.5$)}\\[.75ex]
\Subfig{rc05probe-kA-cA-rK-mH/e0080}{(c) 80 epochs of post-training, no CFG}\hfill\hfill%
\Subfig{rc05probe-kA-cC-rK-mH/e0080}{(d) 80 epochs of post-training, CFG enabled}%
\vspace{-.5mm}%
\caption{\label{figBiggerPictureImages}%
Visual results for prompt alignment vs.~diversity:
\textit{``A cat wearing ski goggles is exploring in the snow.''}
\textbf{(a)}
Without CFG, the original model has high diversity but poor quality and alignment.
\textbf{(b)}
Enabling CFG flips all these axes, leading to a distinctive low-detail look.
\textbf{(c)}
Our RL post-training achieves similar diversity and alignment with more detail.
\textbf{(d)}
Enabling CFG simplifies images and reduces diversity slightly.
The used snapshots align with the dots in Fig.~\ref{figGuidancePlot}.
}%
\end{minipage}%
}%

\gdef\AbEpoch{200}
\gdef\AbXMid{epoch}\gdef\AbXMul{1}
\gdef\AbYMid{oneigbench-1120x4::alignment}\gdef\AbYMul{100}

\newcommand\DefineAblation[4]{
  \gdef#1{#4}
  \ArrayXYLookup{#2}{#1::\AbXMid}{#1::\AbYMid}{\AbEpoch}
  \ArrayMax{#3}{#1::\AbYMid}
}

\newcommand\AbFmt[1]{\evaltext[2]{#1*\AbYMul}}
\newcommand\AbFmB[1]{\textbf{\evaltext[2]{#1*\AbYMul}}}

\DefineAblation{\AbRidA}{\AbFixA}{\AbMaxA}{rc02-kA-cA-rK-mE}
\DefineAblation{\AbRidB}{\AbFixB}{\AbMaxB}{rcbridge-kA-cA-rK-bA}
\DefineAblation{\AbRidC}{\AbFixC}{\AbMaxC}{rc09-kA-cA-rK-bAEFLowChurn}
\DefineAblation{\AbRidD}{\AbFixD}{\AbMaxD}{rc09-kA-cA-rK-bACDEF}
\DefineAblation{\AbRidE}{\AbFixE}{\AbMaxE}{rc05probe-kA-cA-rK-mH}

\DefineAblation{\AbRidF}{\AbFixF}{\AbMaxF}{rc09-kA-cA-rK-aD}
\DefineAblation{\AbRidG}{\AbFixG}{\AbMaxG}{rcablations-kA-cA-rK-aA}
\DefineAblation{\AbRidH}{\AbFixH}{\AbMaxH}{rc09-kA-cA-rK-aH}
\DefineAblation{\AbRidI}{\AbFixI}{\AbMaxI}{rc09-kA-cA-rK-aJ}
\DefineAblation{\AbRidJ}{\AbFixJ}{\AbMaxJ}{rc09-kA-cA-rK-aK}

\newcommand{\figAblationsTable}{%
\centering\footnotesize%
\resizebox{\linewidth}{!}{%
\newdimen\TabWidth\setlength\TabWidth{2.2mm}%
\gdef\Tab##1{\hspace{##1\TabWidth}}%
\begin{tabu}{|c@{\hs{2}}l@{\hs{2}}|r@{\hs{5.3}}r@{\hs{3.3}}|}
\tabucline{-}
\multicolumn{2}{|l|}{\multirow{2}{*}{\bf Ablations}} & \multicolumn{2}{c|}{\tstrut\bf OneIG alignment} \\
\multicolumn{2}{|l|}{} & \multicolumn{1}{c}{\tstrut \AbEpoch{} epochs} & \multicolumn{1}{c|}{Highest} \\
\tabucline{-}\tstrut%
{\scalebox{.9}{A}} & {Flow-GRPO}                                    & {\AbFmt{\AbFixA}} & {\AbFmt{\AbMaxA}} \\
{\scalebox{.9}{B}} & {+ Our stochastic sampling}                    & {\AbFmt{\AbFixB}} & {\AbFmt{\AbMaxB}} \\
{\scalebox{.9}{C}} & {\Tab{1}+ Finite difference flow optimization} & {\AbFmB{\AbFixC}} & {\AbFmt{\AbMaxC}} \\
{\scalebox{.9}{D}} & {\Tab{2}+ Start from the same noise}           & {\AbFmt{\AbFixD}} & {\AbFmt{\AbMaxD}} \\
{\scalebox{.9}{E}} & {\Tab{3}+ Cover more prompts (= Ours)}         & {\AbFmt{\AbFixE}} & {\AbFmB{\AbMaxE}} \\
\tabucline{-}\tstrut%
{\scalebox{.9}{E}} & {Our method}                                   & {\AbFmt{\AbFixE}} & {\AbFmB{\AbMaxE}} \\
{\scalebox{.9}{F}} & {+ One trajectory is deterministic}            & {\AbFmt{\AbFixF}} & {\AbFmt{\AbMaxF}} \\
{\scalebox{.9}{G}} & {+ PPO clipping instead of SPO}                & {\AbFmt{\AbFixG}} & {\AbFmt{\AbMaxG}} \\
{\scalebox{.9}{H}} & {+ Do not normalize difference vector}         & {\AbFmt{\AbFixH}} & {\AbFmt{\AbMaxH}} \\
{\scalebox{.9}{I}} & {+ Compute true reward gradient}               & {\AbFmt{\AbFixI}} & {\AbFmt{\AbMaxI}} \\
{\scalebox{.9}{J}} & {\Tab{1}+ Backpropagate through SDE steps}     & {\AbFmt{\AbFixJ}} & {\AbFmt{\AbMaxJ}} \\
\tabucline{-}
\end{tabu}%
}%
}%

\newcommand{\figAblationsPlotA}{%
\centering\footnotesize%
\begin{tikzpicture}%
\begin{axis}[
  width={1.20\linewidth}, height={54.6mm}, grid={major},
  xmin={0}, xmax={1001}, xmode={linear}, x coord trafo/.code=\pgfmathparse{##1^0.7}, xtick={0, 100, 200, 400, 600, 800, 1000}, xticklabels={$0$, $100$, $200$, $400$, $600$, $800$, \hs{-2}$1$k},
  ymin={57}, ymax={85}, ymode={linear}, y coord trafo/.code=\pgfmathparse{##1^2.0}, ytick={60, 65, 70, 75, 80, 85}, yticklabels={$60$, $65$, $70$, $75$, $80$, \raisebox{-1.6mm}[0mm][0mm]{$85$}},
  legend pos={south east}, legend cell align={left}, legend columns={6}, legend style={nodes={scale=0.80, transform shape}},
]
\gdef\pmx##1{\MinorLine{(##1,57)(##1,85)}}\pmx{300}\pmx{500}\pmx{700}\pmx{900}

\gdef\rid{\AbRidA}
\PlotCurve{C1}{\rid::\AbXMid}{\rid::\AbYMid}{(\X*\AbXMul,\Y*\AbYMul)}
\PlotDataLabelAtX{C1}{\rid::\AbXMid}{\rid::\AbYMid}{\AbEpoch}{\YY}{(\AbEpoch*\AbXMul,\YY*\AbYMul)}{south}{}
\PlotDataLabelAtMax{C1}{\rid::\AbXMid}{\rid::\AbYMid}{\XX}{\YY}{(\XX*\AbXMul,\YY*\AbYMul)}{south}{}

\gdef\rid{\AbRidB}
\PlotCurve{C2}{\rid::\AbXMid}{\rid::\AbYMid}{(\X*\AbXMul,\Y*\AbYMul)}
\PlotDataLabelAtX{C2}{\rid::\AbXMid}{\rid::\AbYMid}{\AbEpoch}{\YY}{(\AbEpoch*\AbXMul,\YY*\AbYMul)}{south}{}
\PlotDataLabelAtMax{C2}{\rid::\AbXMid}{\rid::\AbYMid}{\XX}{\YY}{(\XX*\AbXMul,\YY*\AbYMul)}{south}{}

\gdef\rid{\AbRidC}
\PlotCurve{C3}{\rid::\AbXMid}{\rid::\AbYMid}{(\X*\AbXMul,\Y*\AbYMul)}
\PlotDataLabelAtX{C3}{\rid::\AbXMid}{\rid::\AbYMid}{\AbEpoch}{\YY}{(\AbEpoch*\AbXMul,\YY*\AbYMul)}{south}{}
\PlotDataLabelAtMax{C3}{\rid::\AbXMid}{\rid::\AbYMid}{\XX}{\YY}{(\XX*\AbXMul,\YY*\AbYMul)}{south}{}

\gdef\rid{\AbRidD}
\PlotCurve{C4}{\rid::\AbXMid}{\rid::\AbYMid}{(\X*\AbXMul,\Y*\AbYMul)}
\PlotDataLabelAtX{C4}{\rid::\AbXMid}{\rid::\AbYMid}{\AbEpoch}{\YY}{(\AbEpoch*\AbXMul,\YY*\AbYMul)}{south}{}
\PlotDataLabelAtMax{C4}{\rid::\AbXMid}{\rid::\AbYMid}{\XX}{\YY}{(\XX*\AbXMul,\YY*\AbYMul)}{south}{}

\gdef\rid{\AbRidE}
\PlotCurve{C0}{\rid::\AbXMid}{\rid::\AbYMid}{(\X*\AbXMul,\Y*\AbYMul)}
\PlotDataLabelAtX{C0}{\rid::\AbXMid}{\rid::\AbYMid}{\AbEpoch}{\YY}{(\AbEpoch*\AbXMul,\YY*\AbYMul)}{south}{}
\PlotDataLabelAtMax{C0}{\rid::\AbXMid}{\rid::\AbYMid}{\XX}{\YY}{(\XX*\AbXMul,\YY*\AbYMul)}{south}{}

\legend{A\,\,\,, B\,\,\,, C\,\,\,, D\,\,\,, E}
\end{axis}
\end{tikzpicture}%
}%

\newcommand{\figAblationsPlotB}{%
\centering\footnotesize%
\begin{tikzpicture}%
\begin{axis}[
  width={1.20\linewidth}, height={54.6mm}, grid={major},
  xmin={0}, xmax={1001}, xmode={linear}, x coord trafo/.code=\pgfmathparse{##1^0.7}, xtick={0, 100, 200, 400, 600, 800, 1000}, xticklabels={$0$, $100$, $200$, $400$, $600$, $800$, \hs{-2}$1$k},
  ymin={57}, ymax={85}, ymode={linear}, y coord trafo/.code=\pgfmathparse{##1^2.0}, ytick={60, 65, 70, 75, 80, 85}, yticklabels={$60$, $65$, $70$, $75$, $80$, \raisebox{-1.6mm}[0mm][0mm]{$85$}},
  legend pos={south east}, legend cell align={left}, legend columns={6}, legend style={nodes={scale=0.80, transform shape}},
]
\gdef\pmx##1{\MinorLine{(##1,57)(##1,85)}}\pmx{300}\pmx{500}\pmx{700}\pmx{900}

\gdef\rid{\AbRidE}
\PlotCurve{C0}{\rid::\AbXMid}{\rid::\AbYMid}{(\X*\AbXMul,\Y*\AbYMul)}
\PlotDataLabelAtX{C0}{\rid::\AbXMid}{\rid::\AbYMid}{\AbEpoch}{\YY}{(\AbEpoch*\AbXMul,\YY*\AbYMul)}{south}{}
\PlotDataLabelAtMax{C0}{\rid::\AbXMid}{\rid::\AbYMid}{\XX}{\YY}{(\XX*\AbXMul,\YY*\AbYMul)}{south}{}

\gdef\rid{\AbRidF}
\PlotCurve{C1}{\rid::\AbXMid}{\rid::\AbYMid}{(\X*\AbXMul,\Y*\AbYMul)}
\PlotDataLabelAtX{C1}{\rid::\AbXMid}{\rid::\AbYMid}{\AbEpoch}{\YY}{(\AbEpoch*\AbXMul,\YY*\AbYMul)}{south}{}
\PlotDataLabelAtMax{C1}{\rid::\AbXMid}{\rid::\AbYMid}{\XX}{\YY}{(\XX*\AbXMul,\YY*\AbYMul)}{south}{}

\gdef\rid{\AbRidG}
\PlotCurve{C2}{\rid::\AbXMid}{\rid::\AbYMid}{(\X*\AbXMul,\Y*\AbYMul)}
\PlotDataLabelAtX{C2}{\rid::\AbXMid}{\rid::\AbYMid}{\AbEpoch}{\YY}{(\AbEpoch*\AbXMul,\YY*\AbYMul)}{south}{}
\PlotDataLabelAtMax{C2}{\rid::\AbXMid}{\rid::\AbYMid}{\XX}{\YY}{(\XX*\AbXMul,\YY*\AbYMul)}{south}{}

\gdef\rid{\AbRidH}
\PlotCurve{C3}{\rid::\AbXMid}{\rid::\AbYMid}{(\X*\AbXMul,\Y*\AbYMul)}
\PlotDataLabelAtX{C3}{\rid::\AbXMid}{\rid::\AbYMid}{\AbEpoch}{\YY}{(\AbEpoch*\AbXMul,\YY*\AbYMul)}{south}{}
\PlotDataLabelAtMax{C3}{\rid::\AbXMid}{\rid::\AbYMid}{\XX}{\YY}{(\XX*\AbXMul,\YY*\AbYMul)}{south}{}

\gdef\rid{\AbRidI}
\PlotCurve{C4}{\rid::\AbXMid}{\rid::\AbYMid}{(\X*\AbXMul,\Y*\AbYMul)}
\PlotDataLabelAtX{C4}{\rid::\AbXMid}{\rid::\AbYMid}{\AbEpoch}{\YY}{(\AbEpoch*\AbXMul,\YY*\AbYMul)}{south}{}
\PlotDataLabelAtMax{C4}{\rid::\AbXMid}{\rid::\AbYMid}{\XX}{\YY}{(\XX*\AbXMul,\YY*\AbYMul)}{south}{}

\gdef\rid{\AbRidJ}
\PlotCurve{C5}{\rid::\AbXMid}{\rid::\AbYMid}{(\X*\AbXMul,\Y*\AbYMul)}
\PlotDataLabelAtX{C5}{\rid::\AbXMid}{\rid::\AbYMid}{\AbEpoch}{\YY}{(\AbEpoch*\AbXMul,\YY*\AbYMul)}{south}{}
\PlotDataLabelAtMax{C5}{\rid::\AbXMid}{\rid::\AbYMid}{\XX}{\YY}{(\XX*\AbXMul,\YY*\AbYMul)}{south}{}

\legend{E\,\,\,, F\,\,\,, G\,\,\,, H\,\,\,, I\,\,\,, J}
\end{axis}
\end{tikzpicture}%
}%

\newcommand{\figAblations}{%
\begin{figure*}[t]%
\gdef\h{0.440\linewidth}%
\gdef\hh{0.278\linewidth}%
\gdef\vv{42.5mm}%
\makebox(\h,\vv)[b]{\begin{minipage}{\h}\figAblationsTable\end{minipage}}\hfill%
\makebox(\hh,\vv)[b]{\begin{minipage}{\hh}\figAblationsPlotA\end{minipage}}%
\makebox(\hh,\vv)[b]{\begin{minipage}{\hh}\figAblationsPlotB\end{minipage}}%
\vspace*{-.5ex}%
\caption{\label{figAblations}%
OneIG-Bench alignment score for ablations of our method using the combined reward.
The plots correspond to the top and bottom section of the table, respectively.
We report the alignment score at \AbEpoch{} epochs as well as the highest achieved score, annotated in the plots.
}%
\end{figure*}
}%

\newcommand{\figPerformancePlotA}{%
\hspace*{-1mm}%
\begin{tikzpicture}%
\begin{axis}[
  width={1.10\columnwidth}, height={55mm}, grid={major},
  xmin={0}, xmax={750}, xmode={linear}, x coord trafo/.code=\pgfmathparse{##1^0.4},
  xtick={0, 1, 2, 5, 10, 20, 40, 100, 200, 400, 700}, xticklabels={$0$, {}, $2$, {}, $10$, {}, $40$, $100$, $200$, $400$, $\!\!700\!\!$},
  ymin={22.5}, ymax={31}, ymode={linear}, ytick={23, 24, 25, 26, 27, 28, 29, 30, 31}, yticklabels={$23$, $24$, $25$, $26$, $27$, $28$, $29$, $30$, \raisebox{-1.6mm}[0mm][0mm]{$31$}},
  ylabel={\small Combined reward},
  legend pos={south east}, legend cell align={left}, legend columns={2}, legend style={nodes={scale=0.95, transform shape}, row sep=-1pt},
]
\gdef\xmid{epoch}
\gdef\ymid{rc02-4096x1::combo_rK}\gdef\ymul{10}
\gdef\pmx##1{\MinorLine{(##1,22.5)(##1,31)}}
\pmx{0.1}\pmx{0.2}\pmx{0.3}\pmx{0.4}\pmx{0.5}\pmx{0.6}\pmx{0.7}\pmx{0.8}\pmx{0.9}
\pmx{3}\pmx{4}\pmx{6}\pmx{7}\pmx{8}\pmx{9}
\pmx{30}\pmx{40}\pmx{50}\pmx{60}\pmx{70}\pmx{80}\pmx{90}
\addplot[black, opacity=0.3, thin, forget plot] coordinates {(0,28)(750,28)};

\addlegendentry{Ours, 10 steps\hspace*{.75em}}
\gdef\rid{rc09-kA-cA-rK-mJ}
\gdef\xmulC{0.13317}
\PlotCurve{C2}{\rid::\xmid}{\rid::\ymid}{(\X*\xmulC,\Y*\ymul)}
\gdef\XC{130}\PlotDataLabelAtX{C2}{\rid::\xmid}{\rid::\ymid}{\XC}{\YC}{(\XC*\xmulC,\YC*\ymul)}{east}{\raisebox{2.5mm}{\datalabel{\eval[1]{\XC*\xmulC}}}}

\addlegendentry{Flow-GRPO, 10 steps}
\gdef\rid{rc02-kA-cA-rK-mD}
\gdef\xmulD{0.21344}
\PlotCurve{C3}{\rid::\xmid}{\rid::\ymid}{(\X*\xmulD,\Y*\ymul)}
\gdef\XD{420}\PlotDataLabelAtX{C3}{\rid::\xmid}{\rid::\ymid}{\XD}{\YD}{(\XD*\xmulD,\YD*\ymul)}{west}{\hs{-0.5}\raisebox{-4.5mm}{\datalabel{\eval[1]{\XD*\xmulD}}}}

\addlegendentry{Ours, 40 steps}
\gdef\rid{rc05probe-kA-cA-rK-mH}
\gdef\xmulA{0.45153}
\PlotCurve{C0}{\rid::\xmid}{\rid::\ymid}{(\X*\xmulA,\Y*\ymul)}
\gdef\XA{65}\PlotDataLabelAtX{C0}{\rid::\xmid}{\rid::\ymid}{\XA}{\YA}{(\XA*\xmulA,\YA*\ymul)}{south}{\hs{1}\datalabel{\eval[1]{\XA*\xmulA}}}

\addlegendentry{Flow-GRPO, 40 steps}
\gdef\rid{rc02-kA-cA-rK-mE}
\gdef\xmulB{0.76019}
\PlotCurve{C1}{\rid::\xmid}{\rid::\ymid}{(\X*\xmulB,\Y*\ymul)}
\gdef\XB{740}\PlotDataLabelAtX{C1}{\rid::\xmid}{\rid::\ymid}{\XB}{\YB}{(\XB*\xmulB,\YB*\ymul)}{north}{\hs{2.5}\raisebox{-1.5mm}{\datalabel{\eval[1]{\XB*\xmulB}}}}

\end{axis}
\end{tikzpicture}%
}%

\newcommand{\figPerformancePlotB}{%
\begin{tikzpicture}%
\begin{axis}[
  width={1.10\columnwidth}, height={55mm}, grid={major},
  xmin={0}, xmax={450}, xmode={linear}, x coord trafo/.code=\pgfmathparse{##1^0.4},
  xtick={0, 1, 2, 5, 10, 20, 40, 100, 200, 300, 400}, xticklabels={$0$, {}, $2$, {}, $10$, {}, $40$, $100$, $200$, {}, $400$},
  ymin={22.5}, ymax={31}, ymode={linear}, ytick={23, 24, 25, 26, 27, 28, 29, 30, 31}, yticklabels={$23$, $24$, $25$, $26$, $27$, $28$, $29$, $30$, \raisebox{-1.6mm}[0mm][0mm]{$31$}},
  ylabel={\small Combined reward},
  legend pos={south east}, legend cell align={left}, legend columns={2}, legend style={nodes={scale=0.95, transform shape}, row sep=-1pt},
]
\gdef\xmid{epoch}
\gdef\ymid{rc02-4096x1::combo_rK}\gdef\ymul{10}
\gdef\pmx##1{\MinorLine{(##1,22.5)(##1,31)}}
\pmx{0.1}\pmx{0.2}\pmx{0.3}\pmx{0.4}\pmx{0.5}\pmx{0.6}\pmx{0.7}\pmx{0.8}\pmx{0.9}
\pmx{3}\pmx{4}\pmx{6}\pmx{7}\pmx{8}\pmx{9}
\pmx{30}\pmx{40}\pmx{50}\pmx{60}\pmx{70}\pmx{80}\pmx{90}
\addplot[black, opacity=0.3, thin, forget plot] coordinates {(0,28)(450,28)};

\addlegendentry{Ours, 10 steps\hspace*{.75em}}
\gdef\rid{rc09-kA-cA-rK-mJ}
\gdef\xmulC{0.12194}
\PlotCurve{C2}{\rid::\xmid}{\rid::\ymid}{(\X*\xmulC,\Y*\ymul)}
\gdef\XC{130}\PlotDataLabelAtX{C2}{\rid::\xmid}{\rid::\ymid}{\XC}{\YC}{(\XC*\xmulC,\YC*\ymul)}{east}{\raisebox{2.5mm}{\datalabel{\eval[1]{\XC*\xmulC}}}}

\addlegendentry{Flow-GRPO, 10 steps}
\gdef\rid{rc02-kA-cA-rK-mD}
\gdef\xmulD{0.11448}
\PlotCurve{C3}{\rid::\xmid}{\rid::\ymid}{(\X*\xmulD,\Y*\ymul)}
\gdef\XD{420}\PlotDataLabelAtX{C3}{\rid::\xmid}{\rid::\ymid}{\XD}{\YD}{(\XD*\xmulD,\YD*\ymul)}{west}{\hs{-0.5}\raisebox{-4.5mm}{\datalabel{\eval[1]{\XD*\xmulD}}}}

\addlegendentry{Ours, 40 steps}
\gdef\rid{rc05probe-kA-cA-rK-mH}
\gdef\xmulA{0.44090}
\PlotCurve{C0}{\rid::\xmid}{\rid::\ymid}{(\X*\xmulA,\Y*\ymul)}
\gdef\XA{65}\PlotDataLabelAtX{C0}{\rid::\xmid}{\rid::\ymid}{\XA}{\YA}{(\XA*\xmulA,\YA*\ymul)}{south}{\hs{1}\datalabel{\eval[1]{\XA*\xmulA}}}

\addlegendentry{Flow-GRPO, 40 steps}
\gdef\rid{rc02-kA-cA-rK-mE}
\gdef\xmulB{0.43344}
\PlotCurve{C1}{\rid::\xmid}{\rid::\ymid}{(\X*\xmulB,\Y*\ymul)}
\gdef\XB{740}\PlotDataLabelAtX{C1}{\rid::\xmid}{\rid::\ymid}{\XB}{\YB}{(\XB*\xmulB,\YB*\ymul)}{north}{\hs{2.5}\raisebox{-1.5mm}{\datalabel{\eval[1]{\XB*\xmulB}}}}

\end{axis}
\end{tikzpicture}%
}%

\newcommand{\figPerformancePlot}{%
\begin{figure*}[t]%
\centering\footnotesize%
\figPerformancePlotA{}\hfill%
\figPerformancePlotB{}\\[.3ex]
\makebox[0.525\linewidth]{\small (a) Measured GPU hours}\hfill%
\makebox[0.427\linewidth]{\small (b) Ideal GPU hours}
\vspace*{-.5ex}%
\caption{\label{figPerformancePlot}%
Wall-clock time comparison (NVIDIA H200 hours).
\textbf{(a)} Actual elapsed time.
\textbf{(b)} Assuming ideal, zero-overhead implementation of both methods.
Note the different horizontal scales.
}%
\end{figure*}
}%

\definecolor{TC0}{rgb}{0, 0, 0}
\definecolor{TC1}{rgb}{0.1, 0.2, 1.0}
\definecolor{TC2}{rgb}{1.0, 0.0, 0.0}
\definecolor{TC3}{rgb}{0.0, 0.6, 0.2}
\pgfplotsset{
  legend image with pair/.style={
    legend image code/.code={%
      \draw[##1, densely dotted] (0mm, .6mm) -- (3mm, .6mm);%
      \draw[##1, thick] (0mm,-.2mm) -- (3mm,-.2mm);%
    }
  },
  legend image with pairb/.style={
    legend image code/.code={%
      \draw[##1, line width=0.8pt, dash pattern=on 0.8pt off 1.1pt] (0mm, .6mm) -- (3mm, .6mm);%
      \draw[##1, thick] (0mm,-.2mm) -- (3mm,-.2mm);%
    }
  },
}

\newcommand{\figTrajectoriesRewardPlotBegin}[1]{%
\begin{tikzpicture}%
\begin{axis}[
  xmin={0}, xmax={1}, ymin={\figTrajectoriesPlotxmin}, ymax={\figTrajectoriesPlotxmax},
  xtick={0, 0.5, 1}, xticklabels={\rlap{\vphantom{$1$}}, \raisebox{-1.46ex}[0ex][0ex]{$#1$}, \llap{\vphantom{$0$}}},
  ytick={-1, 0, .5, 1, 2}, yticklabels={$-1$, $0$, $\boldx\ $, $1$, $2$},
  width={25mm}, height={60mm},
]
}%
\newcommand{\figTrajectoriesRewardPlotEnd}{%
\end{axis}%
\end{tikzpicture}\hspace*{-3mm}%
}%
\newcommand{\figTrajectoriesCommonBegin}{%
\begin{tikzpicture}%
\begin{axis}[
  width={85mm}, height={60mm},
  xmin={\figTrajectoriesPlottmin}, xmax={\figTrajectoriesPlottmax}, ymin={\figTrajectoriesPlotxmin}, ymax={\figTrajectoriesPlotxmax},
  xtick={0, 0.2, 0.4, 0.5, 0.6, 0.8, 1.0}, xticklabels={\hphantom{$0$}$0$, $0.2$, $0.4$, \raisebox{-2ex}[0ex][0ex]{$t$}, $0.6$, $0.8$, $1$\hphantom{$1$}},
  ytick={}, yticklabels={},
]
\draw [draw=none] (\figTrajectoriesPlottmin, \figTrajectoriesPlotxmin) coordinate (a) -- (\figTrajectoriesPlottmax, \figTrajectoriesPlotxmax) coordinate (b);
\begin{scope}[on background layer]
\path let \p1=(a), \p2=(b) in
  node[inner sep=0pt, anchor=south west] at (\x1,\y1) {\includegraphics[width={\x2-\x1},height={\y2-\y1}]{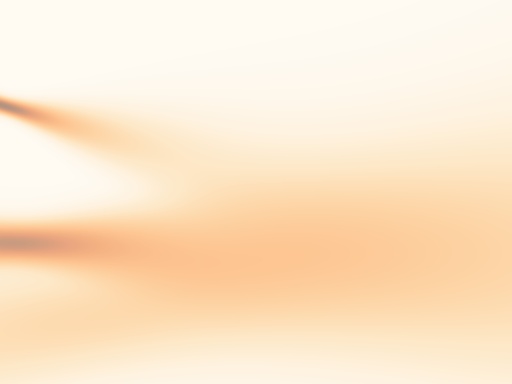}};
\end{scope}%
\foreach \bgidx in {\figTrajectoriesBgFirst,...,\figTrajectoriesBgLast}
  \addplot[black, opacity=0.2, forget plot] table [x index=0, y index=\bgidx] {data/trajectories.txt};
}%
\newcommand{\figTrajectoriesCommonEnd}{%
\end{axis}%
\end{tikzpicture}%
}%

\newcommand{\algStochFM}{%
\begin{algorithm}[t]
\footnotesize
\captionof{algorithm}{\atphantom\ \ Stochastic flow matching sampler}
\Q \textbf{procedure} \textsc{StochasticFlowSampler}$(\theta, \noisedraw, \boldc, \gamma_{0:T})$\\
\QQ $\boldx_0 \gets \noisedraw$\commm{Start from given noise}\\
\QQ \textbf{for each} $i \in \{0, \ldots, T-1\}$ \textbf{do}\commm{Take $T$ sampling steps}\\
\QQQ $\tilde\odetime_{i+1} \gets \odetime_{i+1}/(1-\gamma_i \odetime_{i+1} + \gamma_i)$\commm{Overshoot}\\
\QQQ $\boldv_i \gets \vel(\boldx_i; t_i, \boldc)$\commm{Evaluate velocity}\\
\QQQ $\boldxtilde_{i+1} \gets \boldx_i + (\tilde\odetime_{i+1} - \odetime_i) \boldv_i$
  \commm{Step from $\odetime_i$ to $\tilde\odetime_{i+1}$}\\
\QQQ {\bf sample} $\noisedraw_i \sim \mathcal{N}(\boldzero,\boldI)$\commm{Draw fresh noise}\\[1.5ex]
\vphantom{t}\commm{Add noise to land}\\
\vphantom{t}\commx{at $\odetime_{i+1}$, correct scale}\\[-7ex]
\QQQ $\boldx_{i+1} \gets \displaystyle\frac
  {\boldxtilde_{i+1} + \tilde\odetime_{i+1} \sqrt{\gamma_i^2+2\gamma_i} \cdot \boldsymbol{\epsilon}_i}
  {\gamma_i \tilde\odetime_{i+1} + 1}$\\[.5ex]
\QQ \textbf{return} $\boldx_{0:T},\boldv_{0:T-1}\vphantom{\big{(}}$
\label{alg:stochastic}
\end{algorithm}%
}%

\newcommand{\figDHRF}{%
\begin{figure*}[t]%
\gdef\h{0.49\linewidth}%
\gdef\hh{0.325\linewidth}%
\gdef\Subfig##1{\begin{subfigure}[b]{\h}%
  \ImageWithLabel{\hh}{##1-ep0.jpg}{\small$\mathbf{0}$}\hfill%
  \ImageWithLabel{\hh}{##1-ep25.jpg}{\small$\mathbf{25}$}\hfill%
  \ImageWithLabel{\hh}{##1-ep50.jpg}{\small$\mathbf{50}$}%
  \end{subfigure}}%
\Subfig{dhrf/bird}\hfill\hfill%
\Subfig{dhrf/cat}\\[-3.5ex]%
\caption{\label{figDHRF}%
Qualitative progress across 50 post-training epochs with human rewards.
Left:
\emph{``A small baby bird on a piece of metal.''}
Right:
\emph{``A cat floating inside the international space station.''}
}%
\end{figure*}
}%

%% file: data/trajectories_def.tex
\gdef\figTrajectoriesBgFirst{1}%
\gdef\figTrajectoriesBgLast{30}%
\gdef\figTrajectoriesGrpoGroupFirst{31}%
\gdef\figTrajectoriesGrpoGroupLast{35}%
\gdef\figTrajectoriesGrpoBase{36}%
\gdef\figTrajectoriesMatchingBase{37}%
\gdef\figTrajectoriesMatchingChurned{38}%
\newcommand{\figTrajectoriesMatchingUpdatePoints}[2]{
#1 (0.194,-0.828) #2;
#1 (0.420,-0.298) #2;
#1 (0.677,-0.493) #2;
#1 (0.902,-1.028) #2;
}%
\newcommand{\figTrajectoriesMatchingUpdatePointsB}[2]{
#1 (0.194,0.436) #2;
#1 (0.420,0.512) #2;
#1 (0.677,0.364) #2;
}%
\newcommand{\figTrajectoriesMatchingUpdateVelocities}[3]{
#1 (0.194,-0.828) #2 (0.044,-0.963) #3;
#1 (0.420,-0.298) #2 (0.270,-0.318) #3;
#1 (0.677,-0.493) #2 (0.527,-0.448) #3;
#1 (0.902,-1.028) #2 (0.752,-0.873) #3;
}%
\newcommand{\figTrajectoriesMatchingUpdateVelocitiesB}[3]{
#1 (0.194,0.436) #2 (0.044,0.284) #3;
#1 (0.420,0.512) #2 (0.270,0.518) #3;
#1 (0.677,0.364) #2 (0.527,0.332) #3;
}%
\newcommand{\figTrajectoriesMatchingUpdatePseudogradient}[3]{
#1 (0.010,-0.752) #2 (0.010,0.171) #3;
}%
\newcommand{\figTrajectoriesMatchingUpdateAdded}[3]{
#1 (0.044,-0.963) #2 (0.044,-0.686) #3;
#1 (0.270,-0.318) #2 (0.270,-0.041) #3;
#1 (0.527,-0.448) #2 (0.527,-0.171) #3;
#1 (0.752,-0.873) #2 (0.752,-0.596) #3;
}%
\newcommand{\figTrajectoriesMatchingUpdateAddedB}[3]{
#1 (0.044,0.284) #2 (0.044,0.561) #3;
#1 (0.270,0.518) #2 (0.270,0.795) #3;
#1 (0.527,0.332) #2 (0.527,0.609) #3;
}%
\newcommand{\figTrajectoriesMatchingUpdateVelocitiesUpdated}[3]{
#1 (0.194,-0.828) #2 (0.044,-0.686) #3;
#1 (0.420,-0.298) #2 (0.270,-0.041) #3;
#1 (0.677,-0.493) #2 (0.527,-0.171) #3;
#1 (0.902,-1.028) #2 (0.752,-0.596) #3;
}%
\newcommand{\figTrajectoriesMatchingUpdateVelocitiesUpdatedB}[3]{
#1 (0.194,0.436) #2 (0.044,0.561) #3;
#1 (0.420,0.512) #2 (0.270,0.795) #3;
#1 (0.677,0.364) #2 (0.527,0.609) #3;
}%
\newcommand{\figTrajectoriesGrpoUpdatePts}[2]{
#1 (0.194,0.081) #2;
#1 (0.420,0.125) #2;
#1 (0.677,-0.009) #2;
#1 (0.902,-0.870) #2;
}%
\newcommand{\figTrajectoriesGrpoUpdateVelocities}[3]{
#1 (0.194,0.081) #2 (0.044,0.093) #3;
#1 (0.420,0.125) #2 (0.270,0.087) #3;
#1 (0.677,-0.009) #2 (0.527,-0.011) #3;
#1 (0.902,-0.870) #2 (0.752,-0.739) #3;
}%
\newcommand{\figTrajectoriesGrpoUpdateStepdirs}[3]{
#1 (0.194,0.081) #2 (0.044,0.270) #3;
#1 (0.420,0.125) #2 (0.270,-0.701) #3;
#1 (0.677,-0.009) #2 (0.527,1.044) #3;
#1 (0.902,-0.870) #2 (0.752,-1.109) #3;
}%
\newcommand{\figTrajectoriesGrpoUpdateAdded}[3]{
#1 (0.044,0.093) #2 (0.044,0.143) #3;
#1 (0.270,0.087) #2 (0.270,-0.133) #3;
#1 (0.527,-0.011) #2 (0.527,0.270) #3;
#1 (0.752,-0.739) #2 (0.752,-0.802) #3;
}%
\newcommand{\figTrajectoriesGrpoUpdateVelocitiesUpdated}[3]{
#1 (0.194,0.081) #2 (0.044,0.143) #3;
#1 (0.420,0.125) #2 (0.270,-0.133) #3;
#1 (0.677,-0.009) #2 (0.527,0.270) #3;
#1 (0.902,-0.870) #2 (0.752,-0.802) #3;
}%
\newcommand{\figTrajectoriesGrpoBaseStart}{(0.999,-1.0)}
\newcommand{\figTrajectoriesGrpoBaseEnd}{(0.001,0.08597690551338072)}
\newcommand{\figTrajectoriesGrpoBaseEndX}{0.08597690551338072}
\newcommand{\figTrajectoriesGrpoBaseR}{0.2544532301386452}
\newcommand{\figTrajectoriesMatchingBaseStart}{(0.999,-1.0)}
\newcommand{\figTrajectoriesMatchingBaseEnd}{(0.001,-0.7519093752317716)}
\newcommand{\figTrajectoriesMatchingBaseEndX}{-0.7519093752317716}
\newcommand{\figTrajectoriesMatchingBaseR}{0.8426200539172329}
\newcommand{\figTrajectoriesMatchingChurnedStart}{(0.999,-1.0)}
\newcommand{\figTrajectoriesMatchingChurnedEnd}{(0.001,0.17117263641433153)}
\newcommand{\figTrajectoriesMatchingChurnedEndX}{0.17117263641433153}
\newcommand{\figTrajectoriesMatchingChurnedR}{0.21992901590433767}
\newcommand{\figTrajectoriesPolygonRewPos}[1]{#1(0.500,-0.256)--(0.489,-0.256)--(0.451,-0.211)--(0.416,-0.167)--(0.381,-0.122)--(0.350,-0.078)--(0.320,-0.033)--(0.293,0.011)--(0.269,0.056)--(0.248,0.100)--(0.230,0.144)--(0.214,0.189)--(0.201,0.233)--(0.191,0.278)--(0.182,0.322)--(0.176,0.367)--(0.172,0.411)--(0.170,0.456)--(0.170,0.500)--(0.171,0.544)--(0.173,0.589)--(0.176,0.633)--(0.180,0.678)--(0.185,0.722)--(0.192,0.767)--(0.198,0.811)--(0.206,0.856)--(0.214,0.900)--(0.224,0.944)--(0.234,0.989)--(0.245,1.033)--(0.257,1.078)--(0.270,1.122)--(0.284,1.167)--(0.299,1.211)--(0.316,1.256)--(0.334,1.300)--(0.353,1.344)--(0.373,1.389)--(0.395,1.433)--(0.417,1.478)--(0.441,1.522)--(0.466,1.567)--(0.491,1.611)--(0.500,1.611)--cycle;}%
\newcommand{\figTrajectoriesPolygonRewNegA}[1]{#1(0.500,-1.500)--(0.599,-1.500)--(0.622,-1.456)--(0.647,-1.411)--(0.672,-1.367)--(0.697,-1.322)--(0.723,-1.278)--(0.747,-1.233)--(0.771,-1.189)--(0.794,-1.144)--(0.814,-1.100)--(0.831,-1.056)--(0.845,-1.011)--(0.855,-0.967)--(0.861,-0.922)--(0.863,-0.878)--(0.860,-0.833)--(0.853,-0.789)--(0.840,-0.744)--(0.823,-0.700)--(0.802,-0.656)--(0.776,-0.611)--(0.747,-0.567)--(0.715,-0.522)--(0.680,-0.478)--(0.643,-0.433)--(0.605,-0.389)--(0.566,-0.344)--(0.527,-0.300)--(0.500,-0.300)--cycle;}%
\newcommand{\figTrajectoriesPolygonRewNegB}[1]{#1(0.500,1.656)--(0.516,1.656)--(0.541,1.700)--(0.566,1.744)--(0.591,1.789)--(0.615,1.833)--(0.637,1.878)--(0.658,1.922)--(0.678,1.967)--(0.695,2.011)--(0.711,2.056)--(0.724,2.100)--(0.735,2.144)--(0.743,2.189)--(0.749,2.233)--(0.753,2.278)--(0.754,2.322)--(0.753,2.367)--(0.750,2.411)--(0.745,2.456)--(0.738,2.500)--(0.729,2.544)--(0.719,2.589)--(0.708,2.633)--(0.696,2.678)--(0.684,2.722)--(0.670,2.767)--(0.657,2.811)--(0.643,2.856)--(0.630,2.900)--(0.500,2.900)--cycle;}%
\newcommand{\figTrajectoriesPolygonRewFull}[1]{#1(1.000,-1.500)--(0.599,-1.500)--(0.622,-1.456)--(0.647,-1.411)--(0.672,-1.367)--(0.697,-1.322)--(0.723,-1.278)--(0.747,-1.233)--(0.771,-1.189)--(0.794,-1.144)--(0.814,-1.100)--(0.831,-1.056)--(0.845,-1.011)--(0.855,-0.967)--(0.861,-0.922)--(0.863,-0.878)--(0.860,-0.833)--(0.853,-0.789)--(0.840,-0.744)--(0.823,-0.700)--(0.802,-0.656)--(0.776,-0.611)--(0.747,-0.567)--(0.715,-0.522)--(0.680,-0.478)--(0.643,-0.433)--(0.605,-0.389)--(0.566,-0.344)--(0.527,-0.300)--(0.489,-0.256)--(0.451,-0.211)--(0.416,-0.167)--(0.381,-0.122)--(0.350,-0.078)--(0.320,-0.033)--(0.293,0.011)--(0.269,0.056)--(0.248,0.100)--(0.230,0.144)--(0.214,0.189)--(0.201,0.233)--(0.191,0.278)--(0.182,0.322)--(0.176,0.367)--(0.172,0.411)--(0.170,0.456)--(0.170,0.500)--(0.171,0.544)--(0.173,0.589)--(0.176,0.633)--(0.180,0.678)--(0.185,0.722)--(0.192,0.767)--(0.198,0.811)--(0.206,0.856)--(0.214,0.900)--(0.224,0.944)--(0.234,0.989)--(0.245,1.033)--(0.257,1.078)--(0.270,1.122)--(0.284,1.167)--(0.299,1.211)--(0.316,1.256)--(0.334,1.300)--(0.353,1.344)--(0.373,1.389)--(0.395,1.433)--(0.417,1.478)--(0.441,1.522)--(0.466,1.567)--(0.491,1.611)--(0.516,1.656)--(0.541,1.700)--(0.566,1.744)--(0.591,1.789)--(0.615,1.833)--(0.637,1.878)--(0.658,1.922)--(0.678,1.967)--(0.695,2.011)--(0.711,2.056)--(0.724,2.100)--(0.735,2.144)--(0.743,2.189)--(0.749,2.233)--(0.753,2.278)--(0.754,2.322)--(0.753,2.367)--(0.750,2.411)--(0.745,2.456)--(0.738,2.500)--(0.729,2.544)--(0.719,2.589)--(0.708,2.633)--(0.696,2.678)--(0.684,2.722)--(0.670,2.767)--(0.657,2.811)--(0.643,2.856)--(0.630,2.900)--(1.000,2.900)--cycle;}%
\gdef\figTrajectoriesPlottmin{0}%
\gdef\figTrajectoriesPlottmax{1}%
\gdef\figTrajectoriesPlotxmin{-1.5}%
\gdef\figTrajectoriesPlotxmax{2.9}%

%% file: appendix-figures.tex
\newcommand{\figAppFlowGRPOWeaknesses}{%
\begin{figure*}[t]%
\small\centering%
\gdef\h{0.1615\linewidth}%
\gdef\vv{0mm}
\gdef\ColTitle##1{\makebox[\h][c]{\bf##1}}%
\gdef\RowTitle##1{\rotatebox{90}{\makebox[\h][c]{\small##1}}\hs{0.3}}%
\gdef\Row##1{%
  \RowTitle{Epoch \eval[0]{##1}}%
  \hfill\Image{\h}{rc02-kA-cA-rK-mE/e##1-p015-s06.jpg}%
  \hfill\Image{\h}{rc02-kA-cA-rK-mE/e##1-p043-s07.jpg}%
  \hfill\Image{\h}{rc02-kA-cA-rK-mE/e##1-p094-s01.jpg}%
  \hfill\Image{\h}{rc02-kA-cA-rK-mE/e##1-p100-s05.jpg}%
  \hfill\Image{\h}{rc02-kA-cA-rK-mE/e##1-p134-s04.jpg}%
  \hfill\Image{\h}{rc02-kA-cA-rK-mE/e##1-p144-s00.jpg}%
}%
\hphantom{\RowTitle{E}}\hfill\ColTitle{(a)}\hfill\ColTitle{(b)}\hfill\ColTitle{(c)}\hfill\ColTitle{(d)}\hfill\ColTitle{(e)}\hfill\ColTitle{(f)}\\[1mm]
\Row{0000}\\[\vv]
\Row{0230}\\[\vv]
\Row{0600}\\[\vv]
\Row{0670}\\[\vv]
\Row{0830}\\[\vv]
\Row{0835}\\[\vv]
\Row{1000}%
\caption{\label{figAppFlowGRPOWeaknesses}%
Flow-GRPO results on selected prompts and epochs. Combined reward, no CFG, no KL.
\textbf{(a)} \textit{``A girl with pigtails is holding a giant sunflower.''}
\textbf{(b)} \textit{``Lunch in Bavaria - oil painting''}
\textbf{(c)} \textit{``A cat dressed as a wizard in broad daylight.''}
\textbf{(d)} \textit{``Budapest as a beautiful flowerpunk city, flowerpunk, hyper realistic, high quality, 8k''}
\textbf{(e)} \textit{``Kids race their bikes down the hill as their friends cheer from the sidelines, and a kite flutters in the breeze above them.''}
\textbf{(f)} \textit{``A quiet suburban cul-de-sac, where children play in its enclosed street.''}
}%
\end{figure*}
}%

\newcommand{\figAppProgressA}{%
\begin{figure*}[t]%
\small\centering%
\gdef\h{0.162\linewidth}%
\gdef\vv{-0.8mm}%
\gdef\ImageRow##1{%
  \Image{\h}{\rid/e0000-##1.jpg}%
  \hfill\Image{\h}{\rid/e0050-##1.jpg}%
  \hfill\Image{\h}{\rid/e0100-##1.jpg}%
  \hfill\Image{\h}{\rid/e0200-##1.jpg}%
  \hfill\Image{\h}{\rid/e0500-##1.jpg}%
  \hfill\Image{\h}{\rid/e1000-##1.jpg}%
}%
\gdef\Stats##1{%
  \begin{minipage}{\h}\centering\footnotesize\begin{align*}%
  \gdef\Epoch{##1}\text{Epoch:}~&\eval[0]{\Epoch}\\[\vv]
  \ArrayXYLookup{\Value}{\rid::epoch}{\rid::rc02-4096x1::combo_rK}{\Epoch}\text{Reward:}~&\eval[2]{\Value*10}\\[\vv]
  \ArrayXYLookup{\Value}{\rid::epoch}{\rid::oneigbench-1120x4::alignment}{\Epoch}\text{Alignment:}~&\eval[2]{\Value*100}\\[\vv]
  \ArrayXYLookup{\Value}{\rid::epoch}{\rid::oneigbench-1120x4::diversity}{\Epoch}\text{Diversity:}~&\eval[2]{\Value*100}\\[\vv]
  \end{align*}\end{minipage}%
}%
\gdef\rid{rc05probe-kA-cA-rK-mH}%
{\normalsize\bf Our method without CFG}\\[1.5mm]
\ImageRowTitle{\h}{Seed A}\ImageRow{p095-s01}\\
\ImageRowTitle{\h}{Seed B}\ImageRow{p095-s06}\\
\ImageRowTitle{\h}{Seed C}\ImageRow{p095-s04}\\[-3mm]
\ImageRowTitle{0mm}{\vphantom{A}}\Stats{0}\hfill\Stats{50}\hfill\Stats{100}\hfill\Stats{200}\hfill\Stats{500}\hfill\Stats{1000}\\
\gdef\rid{rc02-kA-cA-rK-mE}%
{\normalsize\bf Flow-GRPO without CFG}\\[1.5mm]
\ImageRowTitle{\h}{Seed A}\ImageRow{p095-s01}\\
\ImageRowTitle{\h}{Seed B}\ImageRow{p095-s06}\\
\ImageRowTitle{\h}{Seed C}\ImageRow{p095-s04}\\[-3mm]
\ImageRowTitle{0mm}{\vphantom{A}}\Stats{0}\hfill\Stats{50}\hfill\Stats{100}\hfill\Stats{200}\hfill\Stats{500}\hfill\Stats{1000}\\
\vspace{-1mm}%
\caption{\label{figAppProgressA}%
Qualitative comparison without classifier-free guidance, using prompt
\emph{``A cat-dragon hybrid. Photograph.''}
}%
\end{figure*}
}%

\newcommand{\figAppProgressC}{%
\begin{figure*}[t]%
\small\centering%
\gdef\h{0.162\linewidth}%
\gdef\vv{-0.8mm}%
\gdef\ImageRow##1{%
  \Image{\h}{\rid/e0000-##1.jpg}%
  \hfill\Image{\h}{\rid/e0050-##1.jpg}%
  \hfill\Image{\h}{\rid/e0100-##1.jpg}%
  \hfill\Image{\h}{\rid/e0200-##1.jpg}%
  \hfill\Image{\h}{\rid/e0500-##1.jpg}%
  \hfill\Image{\h}{\rid/e1000-##1.jpg}%
}%
\gdef\Stats##1{%
  \begin{minipage}{\h}\centering\footnotesize\begin{align*}%
  \gdef\Epoch{##1}\text{Epoch:}~&\eval[0]{\Epoch}\\[\vv]
  \ArrayXYLookup{\Value}{\rid::epoch}{\rid::rc02-4096x1::combo_rK}{\Epoch}\text{Reward:}~&\eval[2]{\Value*10}\\[\vv]
  \ArrayXYLookup{\Value}{\rid::epoch}{\rid::oneigbench-1120x4::alignment}{\Epoch}\text{Alignment:}~&\eval[2]{\Value*100}\\[\vv]
  \ArrayXYLookup{\Value}{\rid::epoch}{\rid::oneigbench-1120x4::diversity}{\Epoch}\text{Diversity:}~&\eval[2]{\Value*100}\\[\vv]
  \end{align*}\end{minipage}%
}%
\gdef\rid{rc05probe-kA-cC-rK-mH}%
{\normalsize\bf Our method with CFG}\\[1.5mm]
\ImageRowTitle{\h}{Seed A}\ImageRow{p095-s01}\\
\ImageRowTitle{\h}{Seed B}\ImageRow{p095-s06}\\
\ImageRowTitle{\h}{Seed C}\ImageRow{p095-s04}\\[-3mm]
\ImageRowTitle{0mm}{\vphantom{A}}\Stats{0}\hfill\Stats{50}\hfill\Stats{100}\hfill\Stats{200}\hfill\Stats{500}\hfill\Stats{1000}\\
\gdef\rid{rc02-kA-cC-rK-mE}%
{\normalsize\bf Flow-GRPO with CFG}\\[1.5mm]
\ImageRowTitle{\h}{Seed A}\ImageRow{p095-s01}\\
\ImageRowTitle{\h}{Seed B}\ImageRow{p095-s06}\\
\ImageRowTitle{\h}{Seed C}\ImageRow{p095-s04}\\[-3mm]
\ImageRowTitle{0mm}{\vphantom{A}}\Stats{0}\hfill\Stats{50}\hfill\Stats{100}\hfill\Stats{200}\hfill\Stats{500}\hfill\Stats{1000}\\
\vspace{-1mm}%
\caption{\label{figAppProgressC}%
Qualitative comparison with classifier-free guidance, using prompt
\emph{``A cat-dragon hybrid. Photograph.''}
}%
\end{figure*}
}%

\newcommand{\figAppGuidance}{%
\begin{figure*}[t]%
\small\centering%
\vspace{13mm}%
\gdef\h{0.159\linewidth}%
\gdef\hh{0.4mm}%
\gdef\hhh{-0.6mm}%
\gdef\vv{0mm}%
\gdef\Title##1{\ImageRowTitle{\h}{\footnotesize##1}}%
\hfill{\makebox[0.483\linewidth][c]{\normalsize\bf Our method (without CFG)}}%
\hfill{\makebox[0.483\linewidth][c]{\normalsize\bf CFG (applied to the original model)}}%
\\[0.8mm]
\gdef\Images##1##2{%
  \Image{\h}{rc05probe-kA-##1-rK-mH/##2-p121-s01.jpg}\hspace{\hh}%
  \Image{\h}{rc05probe-kA-##1-rK-mH/##2-p121-s03.jpg}\hspace{\hh}%
  \Image{\h}{rc05probe-kA-##1-rK-mH/##2-p121-s04.jpg}%
}%
\Title{Epoch 0}\hspace{\hhh}\Images{cA}{e0000}\hfill\Title{$w = 1$}\Images{cA}{e0000}\\[\vv]
\Title{Epoch 45}\hspace{\hhh}\Images{cA}{e0045}\hfill\Title{$w = 2$}\Images{cB}{e0000}\\[\vv]
\Title{Epoch 90}\hspace{\hhh}\Images{cA}{e0090}\hfill\Title{$w = 4.5$}\Images{cC}{e0000}\\[-1mm]
\emph{``A soft, fabric teddy bear sitting on a child's wooden chair, under the warm glow of a brass lamp.''}\\[1.5mm]
\gdef\Images##1##2{%
  \Image{\h}{rc05probe-kA-##1-rK-mH/##2-p102-s00.jpg}\hspace{\hh}%
  \Image{\h}{rc05probe-kA-##1-rK-mH/##2-p102-s03.jpg}\hspace{\hh}%
  \Image{\h}{rc05probe-kA-##1-rK-mH/##2-p102-s02.jpg}%
}%
\Title{Epoch 0}\hspace{\hhh}\Images{cA}{e0000}\hfill\Title{$w = 1$}\Images{cA}{e0000}\\[\vv]
\Title{Epoch 45}\hspace{\hhh}\Images{cA}{e0045}\hfill\Title{$w = 2$}\Images{cB}{e0000}\\[\vv]
\Title{Epoch 90}\hspace{\hhh}\Images{cA}{e0090}\hfill\Title{$w = 4.5$}\Images{cC}{e0000}\\[-1mm]
\hfill\resizebox{0.985\linewidth}{!}{\emph{``outdoor full body shot on Canon DS of a toddler dressed as a medieval emperor, unforgettable dress, intricate details, insane details, v 5,''}}\\[1.5mm]
\vspace{-1mm}%
\caption{\label{figAppGuidance}%
Qualitative comparison between our RL post-training and classifier-free guidance (CFG).
On each row, the OneIG-Bench diversity matches between the two methods.
}%
\vspace{13mm}%
\end{figure*}
}%

\newcommand{\figAppCraftingRewards}{%
\begin{figure*}[t]%
\small\centering%
\gdef\h{0.197\linewidth}%
\gdef\Epoch{e0200}%
\gdef\Title##1{\makebox[\h][c]{\bf##1}}
\gdef\Titlf##1{\raisebox{-1.25ex}[0pt][0pt]{\makebox[\h][c]{\bf##1}}}
\gdef\Prompt##1##2{%
  \Image{\h}{rc09-kA-cA-rI-mH/\Epoch-##1.jpg}%
  \hfill\Image{\h}{rc05probe-kA-cA-rK-mH/\Epoch-##1.jpg}%
  \hfill\Image{\h}{rc05probe-kA-cA-rB-mH/\Epoch-##1.jpg}%
  \hfill\Image{\h}{rc09-kA-cA-rG-mH/\Epoch-##1.jpg}%
  \hfill\Image{\h}{rc09-kA-cA-rM-mH/\Epoch-##1.jpg}\\[-1mm]
  \emph{``##2''}\\[1.5mm]
}
\Title{(a)}\hfill            \Title{(b)}\hfill             \Title{(c)}\hfill                \Title{(d)}\hfill                 \Title{(e)}\\[.2ex]
\Titlf{PickScore only}\hfill \Title{PickScore $+\!$}\hfill \Titlf{VLM alignment only}\hfill \Title{VLM alignment $+\!$}\hfill \Title{VLM alignment $+\!$}\\[0mm]
\Title{}\hfill               \Title{VLM alignment}\hfill   \Title{}\hfill                   \Title{VLM quality}\hfill         \Title{VLM photorealism}\\[1mm]
\Prompt{p133-s01}{A beach scene where the sandcastle appears taller than the nearby cooler.}
\Prompt{p144-s00}{A quiet suburban cul-de-sac, where children play in its enclosed street.}
\Prompt{p145-s02}{A single rose growing through a crack.}
\Prompt{p043-s04}{Lunch in Bavaria - oil painting}
\Prompt{p122-s02}{A sphinx talking with travelers in front of the Pyramids at sunset.}
\vspace{-2mm}%
\caption{\label{figAppCraftingRewards}%
Effect of different reward functions on the visual look.
Each column was post-trained for 200 epochs using our method without KL regularization or CFG.
}%
\end{figure*}
}%

\newcommand{\figAppDHRF}{%
\begin{figure*}[t]%
\small\centering%
\vspace*{5mm}%
\color{black}%
\gdef\h{0.162\linewidth}%
\gdef\DHRFAppImages##1{%
  \ImageWithLabel{\h}{dhrf_appendix/##1/ep00.jpg}{0}\hfill%
  \ImageWithLabel{\h}{dhrf_appendix/##1/ep10.jpg}{10}\hfill%
  \ImageWithLabel{\h}{dhrf_appendix/##1/ep20.jpg}{20}\hfill%
  \ImageWithLabel{\h}{dhrf_appendix/##1/ep30.jpg}{30}\hfill%
  \ImageWithLabel{\h}{dhrf_appendix/##1/ep40.jpg}{40}\hfill%
  \ImageWithLabel{\h}{dhrf_appendix/##1/ep50.jpg}{50}%
}%
\gdef\DHRFAppPrompt##1{\makebox[\linewidth][c]{\emph{``##1''}}}%
\gdef\DHRFAppRow##1##2{%
  \DHRFAppImages{##1}\\[-0.7mm]%
  \DHRFAppPrompt{##2}\\[1.2mm]%
}%
\DHRFAppRow{chick}{A fluffy chick is nested in an antique coffee cup.}%
\DHRFAppRow{bird}{A bird that is sitting in the rim of a tire.}%
\DHRFAppRow{giraffe}{A tall giraffe.}%
\DHRFAppRow{prince}{The Little Prince talking to the fox in an animated style.}%
\DHRFAppRow{owl}{A disheveled owl is perched on a pine tree.}%
\DHRFAppImages{flowers}\\[-0.7mm]%
\DHRFAppPrompt{A vase of flowers.}%
\vspace{-1mm}%
\caption{\label{figAppDHRF}%
Additional examples of qualitative progress across 50 post-training epochs with direct human rewards.
}%
\vspace*{5mm}%
\end{figure*}
}%

\newcommand{\MetWidth}{1.15\columnwidth}
\newcommand{\MetHeight}{56mm}

\newcommand{\MetCurves}{%
\gdef\xmid{epoch}\gdef\xmul{1}
\gdef\rid{rc05probe-kA-cA-rK-mH}\PlotCurve{C0}{\rid::\xmid}{\rid::\ymid}{(\X*\xmul,\Y*\ymul)}
\gdef\rid{rc02-kA-cA-rK-mE}\PlotCurve{C1}{\rid::\xmid}{\rid::\ymid}{(\X*\xmul,\Y*\ymul)}
\gdef\rid{rc05probe-kA-cC-rK-mH}\PlotCurve{C2}{\rid::\xmid}{\rid::\ymid}{(\X*\xmul,\Y*\ymul)}
\gdef\rid{rc02-kA-cC-rK-mE}\PlotCurve{C3}{\rid::\xmid}{\rid::\ymid}{(\X*\xmul,\Y*\ymul)}
}

\newcommand{\MetLegend}{%
\addlegendentry{Ours, $w{=}1$}
\addlegendentry{Flow-GRPO, $w{=}1$}
\addlegendentry{Ours, $w{=}4.5$}
\addlegendentry{Flow-GRPO, $w{=}4.5$}
}

\newcommand{\MetReward}{%
\centering\footnotesize\begin{tikzpicture}\begin{axis}[
  width={\MetWidth}, height={\MetHeight}, grid={major},
  xmin={0}, xmax={1000}, xmode={linear}, xtick={0, 200, 400, 600, 800, 1000}, xticklabels={$0$, $200$, $400$, $600$, $800$, \hs{-2}$1$k},
  ymin={22.5}, ymax={31.5}, ymode={linear}, ytick={23, 24, 25, 26, 27, 28, 29, 30, 31}, yticklabels={$23$, $24$, $25$, $26$, $27$, $28$, $29$, $30$, $31$},
  legend pos={south east}, legend cell align={left}, legend columns={2}, legend style={nodes={scale=0.95, transform shape}},
]
\gdef\ymid{rc02-4096x1::combo_rK}\gdef\ymul{10}
\gdef\pmx##1{\MinorLine{(##1,22.5)(##1,31.5)}}\pmx{100}\pmx{300}\pmx{500}\pmx{700}\pmx{900}
\MetCurves\MetLegend
\end{axis}\end{tikzpicture}%
}%

\newcommand{\MetPickScore}{%
\centering\footnotesize\begin{tikzpicture}\begin{axis}[
  width={\MetWidth}, height={\MetHeight}, grid={major},
  xmin={0}, xmax={1000}, xmode={linear}, xtick={0, 200, 400, 600, 800, 1000}, xticklabels={$0$, $200$, $400$, $600$, $800$, \hs{-2}$1$k},
  ymin={20}, ymax={23.5}, ymode={linear}, ytick={20, 21, 22, 23}, yticklabels={\raisebox{0.2mm}[0mm][0mm]{$20$}, $21$, $22$, $23$},
]
\gdef\ymid{rc02-4096x1::pickscore_score}\gdef\ymul{100}
\gdef\pmx##1{\MinorLine{(##1,20)(##1,23.5)}}\pmx{100}\pmx{300}\pmx{500}\pmx{700}\pmx{900}
\gdef\pmy##1{\MinorLine{(0,##1)(1000,##1)}}\pmy{20.5}\pmy{21.5}\pmy{22.5}
\MetCurves
\end{axis}\end{tikzpicture}%
}%

\newcommand{\MetVLMAlignment}{%
\centering\footnotesize\begin{tikzpicture}\begin{axis}[
  width={\MetWidth}, height={\MetHeight}, grid={major},
  xmin={0}, xmax={1000}, xmode={linear}, xtick={0, 200, 400, 600, 800, 1000}, xticklabels={$0$, $200$, $400$, $600$, $800$, \hs{-2}$1$k},
  ymin={25}, ymax={85}, ymode={linear}, ytick={30, 40, 50, 60, 70, 80}, yticklabels={$30$, $40$, $50$, $60$, $70$, $80$},
]
\gdef\ymid{rc02-4096x1::prompt_rB}\gdef\ymul{100}
\gdef\pmx##1{\MinorLine{(##1,25)(##1,85)}}\pmx{100}\pmx{300}\pmx{500}\pmx{700}\pmx{900}
\gdef\pmy##1{\MinorLine{(0,##1)(1000,##1)}}\pmy{35}\pmy{45}\pmy{55}\pmy{65}\pmy{75}
\MetCurves
\end{axis}\end{tikzpicture}%
}%

\newcommand{\MetOneIGAlignment}{%
\centering\footnotesize\begin{tikzpicture}\begin{axis}[
  width={\MetWidth}, height={\MetHeight}, grid={major},
  xmin={0}, xmax={1000}, xmode={linear}, xtick={0, 200, 400, 600, 800, 1000}, xticklabels={$0$, $200$, $400$, $600$, $800$, \hs{-2}$1$k},
  ymin={56}, ymax={88}, ymode={linear}, ytick={60, 65, 70, 75, 80, 85}, yticklabels={$60$, $65$, $70$, $75$, $80$, $85$},
]
\gdef\ymid{oneigbench-1120x4::alignment}\gdef\ymul{100}
\gdef\pmx##1{\MinorLine{(##1,56)(##1,88)}}\pmx{100}\pmx{300}\pmx{500}\pmx{700}\pmx{900}
\MetCurves
\end{axis}\end{tikzpicture}%
}%

\newcommand{\MetOneIGReasoning}{%
\centering\footnotesize\begin{tikzpicture}\begin{axis}[
  width={\MetWidth}, height={\MetHeight}, grid={major},
  xmin={0}, xmax={1000}, xmode={linear}, xtick={0, 200, 400, 600, 800, 1000}, xticklabels={$0$, $200$, $400$, $600$, $800$, \hs{-2}$1$k},
  ymin={17}, ymax={31.5}, ymode={linear}, ytick={18, 20, 22, 24, 26, 28, 30}, yticklabels={$18$, $20$, $22$, $24$, $26$, $28$, $30$},
]
\gdef\ymid{oneigbench-1120x4::reasoning}\gdef\ymul{100}
\gdef\pmx##1{\MinorLine{(##1,17)(##1,31.5)}}\pmx{100}\pmx{300}\pmx{500}\pmx{700}\pmx{900}
\MetCurves
\end{axis}\end{tikzpicture}%
}%

\newcommand{\MetOneIGStyle}{%
\centering\footnotesize\begin{tikzpicture}\begin{axis}[
  width={\MetWidth}, height={\MetHeight}, grid={major},
  xmin={0}, xmax={1000}, xmode={linear}, xtick={0, 200, 400, 600, 800, 1000}, xticklabels={$0$, $200$, $400$, $600$, $800$, \hs{-2}$1$k},
  ymin={18}, ymax={38}, ymode={linear}, ytick={20, 25, 30, 35}, yticklabels={$20$, $25$, $30$, $35$},
]
\gdef\ymid{oneigbench-1120x4::style}\gdef\ymul{100}
\gdef\pmx##1{\MinorLine{(##1,18)(##1,38)}}\pmx{100}\pmx{300}\pmx{500}\pmx{700}\pmx{900}
\MetCurves
\end{axis}\end{tikzpicture}%
}%

\newcommand{\MetOneIGText}{%
\centering\footnotesize\begin{tikzpicture}\begin{axis}[
  width={\MetWidth}, height={\MetHeight}, grid={major},
  xmin={0}, xmax={1000}, xmode={linear}, xtick={0, 200, 400, 600, 800, 1000}, xticklabels={$0$, $200$, $400$, $600$, $800$, \hs{-2}$1$k},
  ymin={0}, ymax={55}, ymode={linear}, ytick={0, 10, 20, 30, 40, 50}, yticklabels={$0$, $10$, $20$, $30$, $40$, $50$},
]
\gdef\ymid{oneigbench-1120x4::text}\gdef\ymul{100}
\gdef\pmx##1{\MinorLine{(##1,0)(##1,55)}}\pmx{100}\pmx{300}\pmx{500}\pmx{700}\pmx{900}
\MetCurves
\end{axis}\end{tikzpicture}%
}%

\newcommand{\MetOneIGDiversity}{%
\centering\footnotesize\begin{tikzpicture}\begin{axis}[
  width={\MetWidth}, height={\MetHeight}, grid={major},
  xmin={0}, xmax={1000}, xmode={linear}, xtick={0, 200, 400, 600, 800, 1000}, xticklabels={$0$, $200$, $400$, $600$, $800$, \hs{-2}$1$k},
  ymin={8}, ymax={53}, ymode={linear}, ytick={10, 20, 30, 40, 50}, yticklabels={$10$, $20$, $30$, $40$, $50$},
]
\gdef\ymid{oneigbench-1120x4::diversity}\gdef\ymul{100}
\gdef\pmx##1{\MinorLine{(##1,8)(##1,53)}}\pmx{100}\pmx{300}\pmx{500}\pmx{700}\pmx{900}
\gdef\pmy##1{\MinorLine{(0,##1)(1000,##1)}}\pmy{15}\pmy{25}\pmy{35}\pmy{45}
\MetCurves
\end{axis}\end{tikzpicture}%
}%

\newcommand{\MetDreamSimDiversity}{%
\centering\footnotesize\begin{tikzpicture}\begin{axis}[
  width={\MetWidth}, height={\MetHeight}, grid={major},
  xmin={0}, xmax={1000}, xmode={linear}, xtick={0, 200, 400, 600, 800, 1000}, xticklabels={$0$, $200$, $400$, $600$, $800$, \hs{-2}$1$k},
  ymin={0.15}, ymax={1.05}, ymode={linear}, ytick={0.2, 0.4, 0.6, 0.8, 1.0}, yticklabels={$0.2$, $0.4$, $0.6$, $0.8$, $1.0$},
]
\gdef\ymid{hpsv2-3200x4::dreamsim_diversity}\gdef\ymul{1}
\gdef\pmx##1{\MinorLine{(##1,0.15)(##1,1.05)}}\pmx{100}\pmx{300}\pmx{500}\pmx{700}\pmx{900}
\gdef\pmy##1{\MinorLine{(0,##1)(1000,##1)}}\pmy{0.3}\pmy{0.5}\pmy{0.7}\pmy{0.9}
\MetCurves
\end{axis}\end{tikzpicture}%
}%

\newcommand{\MetHPS}{%
\centering\footnotesize\begin{tikzpicture}\begin{axis}[
  width={\MetWidth}, height={\MetHeight}, grid={major},
  xmin={0}, xmax={1000}, xmode={linear}, xtick={0, 200, 400, 600, 800, 1000}, xticklabels={$0$, $200$, $400$, $600$, $800$, \hs{-2}$1$k},
  ymin={25}, ymax={30}, ymode={linear}, ytick={25, 26, 27, 28, 29, 30}, yticklabels={\raisebox{0.2mm}{$25$}, $26$, $27$, $28$, $29$, \raisebox{-1.6mm}[0mm][0mm]{$30$}},
]
\gdef\ymid{hpsv2-3200x4::hpsv2_score}\gdef\ymul{100}
\gdef\pmx##1{\MinorLine{(##1,25)(##1,30)}}\pmx{100}\pmx{300}\pmx{500}\pmx{700}\pmx{900}
\MetCurves
\end{axis}\end{tikzpicture}%
}%

\newcommand{\MetClipH}{%
\centering\footnotesize\begin{tikzpicture}\begin{axis}[
  width={\MetWidth}, height={\MetHeight}, grid={major},
  xmin={0}, xmax={1000}, xmode={linear}, xtick={0, 200, 400, 600, 800, 1000}, xticklabels={$0$, $200$, $400$, $600$, $800$, \hs{-2}$1$k},
  ymin={30}, ymax={39}, ymode={linear}, ytick={30, 32, 34, 36, 38}, yticklabels={\raisebox{0.2mm}{$30$}, $32$, $34$, $36$, $38$},
]
\gdef\ymid{hpsv2-3200x4::clip_h14_score}\gdef\ymul{100}
\gdef\pmx##1{\MinorLine{(##1,30)(##1,39)}}\pmx{100}\pmx{300}\pmx{500}\pmx{700}\pmx{900}
\gdef\pmy##1{\MinorLine{(0,##1)(1000,##1)}}\pmy{31}\pmy{33}\pmy{35}\pmy{37}
\MetCurves
\end{axis}\end{tikzpicture}%
}%

\newcommand{\MetClipL}{%
\centering\footnotesize\begin{tikzpicture}\begin{axis}[
  width={\MetWidth}, height={\MetHeight}, grid={major},
  xmin={0}, xmax={1000}, xmode={linear}, xtick={0, 200, 400, 600, 800, 1000}, xticklabels={$0$, $200$, $400$, $600$, $800$, \hs{-2}$1$k},
  ymin={24}, ymax={31.5}, ymode={linear}, ytick={24, 26, 28, 30}, yticklabels={\raisebox{0.2mm}{$24$}, $26$, $28$, $30$},
]
\gdef\ymid{hpsv2-3200x4::clip_l14_score}\gdef\ymul{100}
\gdef\pmx##1{\MinorLine{(##1,24)(##1,31.5)}}\pmx{100}\pmx{300}\pmx{500}\pmx{700}\pmx{900}
\gdef\pmy##1{\MinorLine{(0,##1)(1000,##1)}}\pmy{25}\pmy{27}\pmy{29}\pmy{31}
\MetCurves
\end{axis}\end{tikzpicture}%
}%

\newcommand{\figAppMetrics}{%
\begin{figure*}[t]
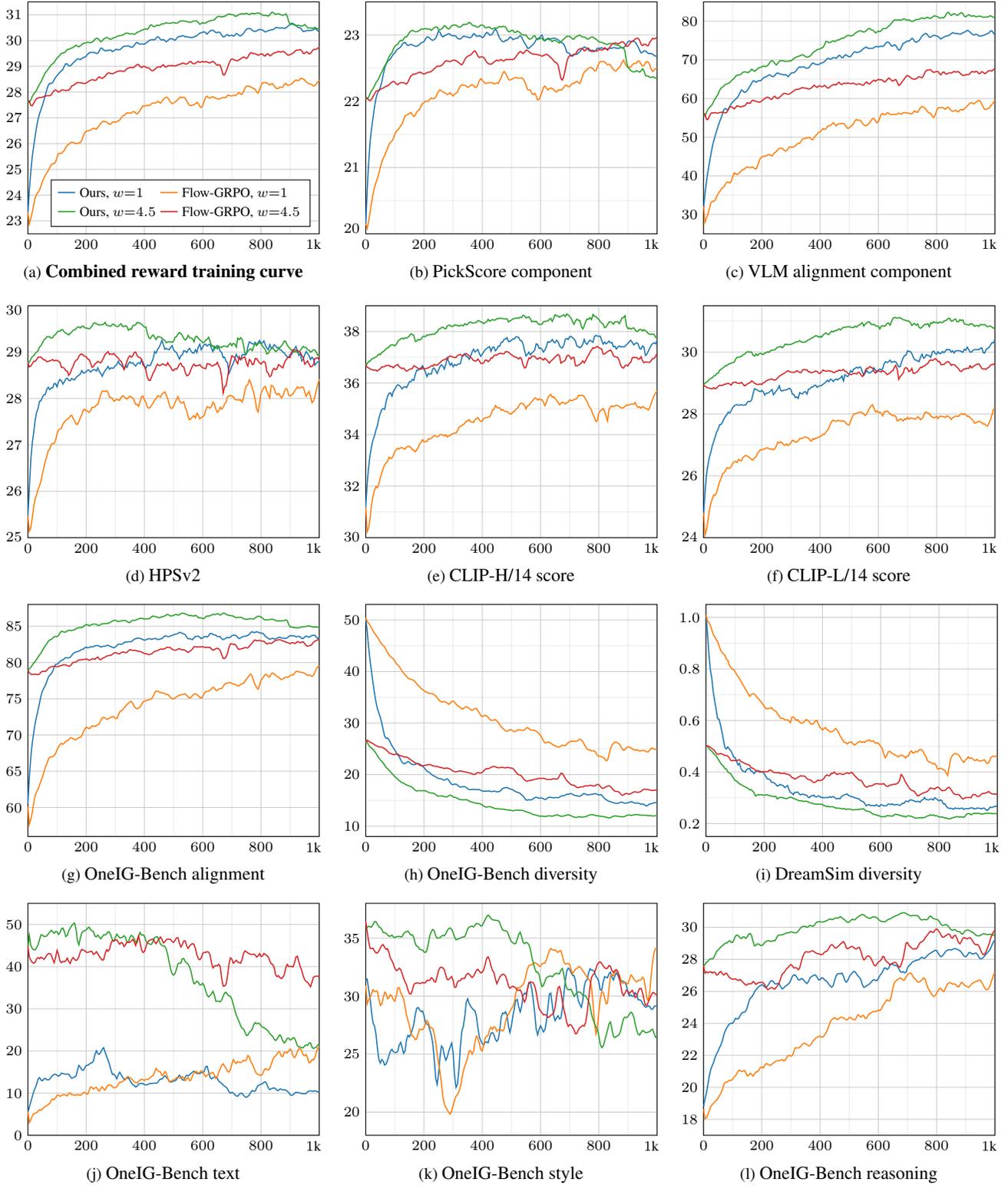
%
\gdef\h{0.33\linewidth}%
\gdef\vv{0.5mm}%
\gdef\vvv{2.5mm}%
\begin{subfigure}[b]{\h}\MetReward\vspace{\vv}\caption{\small \bf Combined reward training curve}\end{subfigure}\hfill%
\begin{subfigure}[b]{\h}\MetPickScore\vspace{\vv}\caption{\small PickScore component}\end{subfigure}\hfill%
\begin{subfigure}[b]{\h}\MetVLMAlignment\vspace{\vv}\caption{\small VLM alignment component}\end{subfigure}\\[\vvv]
\begin{subfigure}[b]{\h}\MetHPS\vspace{\vv}\caption{\small HPSv2}\end{subfigure}\hfill%
\begin{subfigure}[b]{\h}\MetClipH\vspace{\vv}\caption{\small CLIP-H/14 score}\end{subfigure}\hfill%
\begin{subfigure}[b]{\h}\MetClipL\vspace{\vv}\caption{\small CLIP-L/14 score}\end{subfigure}\\[\vvv]%
\begin{subfigure}[b]{\h}\MetOneIGAlignment\vspace{\vv}\caption{\small OneIG-Bench alignment}\end{subfigure}\hfill%
\begin{subfigure}[b]{\h}\MetOneIGDiversity\vspace{\vv}\caption{\small OneIG-Bench diversity}\end{subfigure}\hfill%
\begin{subfigure}[b]{\h}\MetDreamSimDiversity\vspace{\vv}\caption{\small DreamSim diversity}\end{subfigure}\\[\vvv]
\begin{subfigure}[b]{\h}\MetOneIGText\vspace{\vv}\caption{\small OneIG-Bench text}\end{subfigure}\hfill%
\begin{subfigure}[b]{\h}\MetOneIGStyle\vspace{\vv}\caption{\small OneIG-Bench style}\end{subfigure}\hfill%
\begin{subfigure}[b]{\h}\MetOneIGReasoning\vspace{\vv}\caption{\small OneIG-Bench reasoning}\end{subfigure}\\%
\vspace{-2mm}%
\caption{\label{figAppMetrics}%
A wide range of benchmarks for four models post-trained using the combined reward. 
}%
\end{figure*}
}%

\newcommand{\figAppKLwithoutCfgA}{%
\centering\footnotesize%
\begin{tikzpicture}%
\begin{axis}[
  width={1.17\columnwidth}, height={60mm}, grid={major},
  xmin={0}, xmax={1000}, xmode={linear}, xtick={0, 200, 400, 600, 800, 1000}, xticklabels={$0$, $200$, $400$, $600$, $800$, \hs{-2}$1$k},
  ymin={56}, ymax={88}, ymode={linear}, ytick={60, 65, 70, 75, 80, 85}, yticklabels={$60$, $65$, $70$, $75$, $80$, $85$},
  legend pos={south east}, legend cell align={left}, legend columns={2}, legend style={nodes={scale=0.95, transform shape}},
]
\gdef\xmid{epoch}\gdef\xmul{1}
\gdef\ymid{oneigbench-1120x4::alignment}\gdef\ymul{100}
\gdef\pmx##1{\MinorLine{(##1,56)(##1,88)}}\pmx{100}\pmx{300}\pmx{500}\pmx{700}\pmx{900}

\gdef\rid{rc05probe-kA-cA-rK-mH}\PlotCurve{C0}{\rid::\xmid}{\rid::\ymid}{(\X*\xmul,\Y*\ymul)}\addlegendentry{Ours}
\gdef\rid{rc02-kA-cA-rK-mE}\PlotCurve{C1}{\rid::\xmid}{\rid::\ymid}{(\X*\xmul,\Y*\ymul)}\addlegendentry{Flow-GRPO}
\gdef\rid{rc09-kB-cA-rK-mH}\PlotCurveDashed{C0}{\rid::\xmid}{\rid::\ymid}{(\X*\xmul,\Y*\ymul)}\addlegendentry{Ours$+$KL}
\gdef\rid{rc02-kB-cA-rK-mE}\PlotCurveDashed{C1}{\rid::\xmid}{\rid::\ymid}{(\X*\xmul,\Y*\ymul)}\addlegendentry{Flow-GRPO$+$KL}

\end{axis}
\end{tikzpicture}%
}%

\newcommand{\figAppKLwithoutCfgB}{%
\centering\footnotesize%
\begin{tikzpicture}%
\begin{axis}[
  width={1.17\columnwidth}, height={60mm}, grid={major},
  xmin={0}, xmax={1000}, xmode={linear}, xtick={0, 200, 400, 600, 800, 1000}, xticklabels={$0$, $200$, $400$, $600$, $800$, \hs{-2}$1$k},
  ymin={25}, ymax={29.5}, ymode={linear}, ytick={25, 26, 27, 28, 29}, yticklabels={\raisebox{2mm}{$25$}, $26$, $27$, $28$, $29$},
  legend pos={south east}, legend cell align={left}, legend columns={2}, legend style={nodes={scale=0.95, transform shape}},
]
\gdef\xmid{epoch}\gdef\xmul{1}
\gdef\ymid{hpsv2-3200x4::hpsv2_score}\gdef\ymul{100}
\gdef\pmx##1{\MinorLine{(##1,25)(##1,29.5)}}\pmx{100}\pmx{300}\pmx{500}\pmx{700}\pmx{900}
\gdef\pmy##1{\MinorLine{(0,##1)(1000,##1)}}\pmy{25.5}\pmy{26.5}\pmy{27.5}\pmy{28.5}

\gdef\rid{rc05probe-kA-cA-rK-mH}\PlotCurve{C0}{\rid::\xmid}{\rid::\ymid}{(\X*\xmul,\Y*\ymul)}%
\gdef\rid{rc02-kA-cA-rK-mE}\PlotCurve{C1}{\rid::\xmid}{\rid::\ymid}{(\X*\xmul,\Y*\ymul)}%
\gdef\rid{rc09-kB-cA-rK-mH}\PlotCurveDashed{C0}{\rid::\xmid}{\rid::\ymid}{(\X*\xmul,\Y*\ymul)}%
\gdef\rid{rc02-kB-cA-rK-mE}\PlotCurveDashed{C1}{\rid::\xmid}{\rid::\ymid}{(\X*\xmul,\Y*\ymul)}%

\end{axis}
\end{tikzpicture}%
}%

\newcommand{\figAppKLwithoutCfgC}{%
\centering\footnotesize%
\begin{tikzpicture}%
\begin{axis}[
  width={1.17\columnwidth}, height={60mm}, grid={major},
  xmin={0}, xmax={1000}, xmode={linear}, xtick={0, 200, 400, 600, 800, 1000}, xticklabels={$0$, $200$, $400$, $600$, $800$, \hs{-2}$1$k},
  ymin={10}, ymax={53}, ymode={linear}, ytick={10, 20, 30, 40, 50}, yticklabels={\raisebox{2mm}{$10$}, $20$, $30$, $40$, $50$},
  legend pos={south east}, legend cell align={left}, legend columns={2}, legend style={nodes={scale=0.95, transform shape}},
]
\gdef\xmid{epoch}\gdef\xmul{1}
\gdef\ymid{oneigbench-1120x4::diversity}\gdef\ymul{100}
\gdef\pmx##1{\MinorLine{(##1,10)(##1,53)}}\pmx{100}\pmx{300}\pmx{500}\pmx{700}\pmx{900}
\gdef\pmy##1{\MinorLine{(0,##1)(1000,##1)}}\pmy{15}\pmy{25}\pmy{35}\pmy{45}

\gdef\rid{rc05probe-kA-cA-rK-mH}\PlotCurve{C0}{\rid::\xmid}{\rid::\ymid}{(\X*\xmul,\Y*\ymul)}%
\gdef\rid{rc02-kA-cA-rK-mE}\PlotCurve{C1}{\rid::\xmid}{\rid::\ymid}{(\X*\xmul,\Y*\ymul)}%
\gdef\rid{rc09-kB-cA-rK-mH}\PlotCurveDashed{C0}{\rid::\xmid}{\rid::\ymid}{(\X*\xmul,\Y*\ymul)}%
\gdef\rid{rc02-kB-cA-rK-mE}\PlotCurveDashed{C1}{\rid::\xmid}{\rid::\ymid}{(\X*\xmul,\Y*\ymul)}%

\end{axis}
\end{tikzpicture}%
}%

\newcommand{\figAppKLwithoutCfg}{%
\begin{figure*}[t]
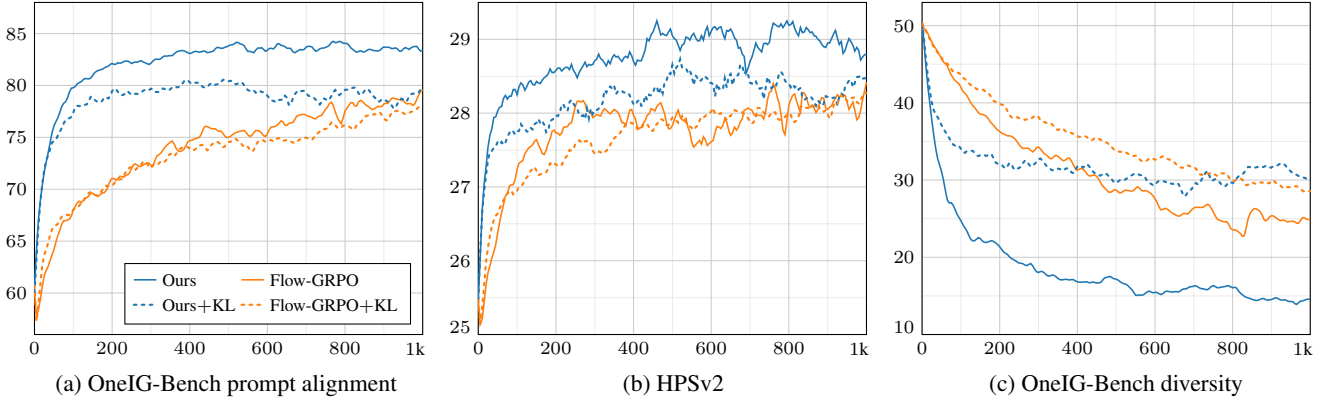
%
\hspace*{-1mm}%
\begin{subfigure}[b]{0.33\linewidth}\figAppKLwithoutCfgA\end{subfigure}\hfill%
\begin{subfigure}[b]{0.33\linewidth}\figAppKLwithoutCfgB\end{subfigure}\hfill%
\begin{subfigure}[b]{0.33\linewidth}\figAppKLwithoutCfgC\end{subfigure}\\[-.2ex]
\makebox[0.335\linewidth]{\small (a) OneIG-Bench prompt alignment}\hfill%
\makebox[0.330\linewidth]{\small (b) HPSv2\ \ }\hfill%
\makebox[0.310\linewidth]{\small (c) OneIG-Bench diversity}%
\vspace{-.5ex}%
\caption{\label{figAppKLwithoutCfg}%
Effect of KL regularization on prompt alignment, human preferences, and diversity.
These results are without CFG and with KL ratio $\beta = 0.04$, which is the default in Flow-GRPO.
}%
\end{figure*}
}%

\gdef\ScEpoch{200}
\gdef\ScXMid{epoch}\gdef\ScXMul{1}
\gdef\ScYMid{oneigbench-1120x4::alignment}\gdef\ScYMul{100}

\newcommand\DefineAblationAppendix[4]{
  \gdef#1{#4}
  \ArrayXYLookup{#2}{#1::\ScXMid}{#1::\ScYMid}{\ScEpoch}
  \ArrayMax{#3}{#1::\ScYMid}
}

\newcommand\ScFmt[1]{\evaltext[2]{#1*\ScYMul}}
\newcommand\ScFmB[1]{\textbf{\evaltext[2]{#1*\ScYMul}}}

\DefineAblationAppendix{\ScRidU}{\ScFixU}{\ScMaxU}{rc05probe-kA-cA-rK-mH}
\DefineAblationAppendix{\ScRidUL}{\ScFixUL}{\ScMaxUL}{rc05-kA-cA-rK-sUL}
\DefineAblationAppendix{\ScRidUH}{\ScFixUH}{\ScMaxUH}{rc05-kA-cA-rK-sUH}
\DefineAblationAppendix{\ScRidI}{\ScFixI}{\ScMaxI}{rc05-kA-cA-rK-sMM}
\DefineAblationAppendix{\ScRidIL}{\ScFixIL}{\ScMaxIL}{rc05-kA-cA-rK-sML}
\DefineAblationAppendix{\ScRidIH}{\ScFixIH}{\ScMaxIH}{rc05-kA-cA-rK-sMH}
\DefineAblationAppendix{\ScRidP}{\ScFixP}{\ScMaxP}{rc05-kA-cA-rK-sPM}
\DefineAblationAppendix{\ScRidPL}{\ScFixPL}{\ScMaxPL}{rc05-kA-cA-rK-sPL}
\DefineAblationAppendix{\ScRidPH}{\ScFixPH}{\ScMaxPH}{rc05-kA-cA-rK-sPH}

\newcommand{\figAppSchedulesTable}{%
\centering\small%
\resizebox{\linewidth}{!}{%
\newdimen\TabWidth\setlength\TabWidth{2.2mm}%
\gdef\Tab##1{\hspace{##1\TabWidth}}%
\begin{tabu}{|l|c@{\hs{2}}c|r@{\hs{5.3}}r@{\hs{3.3}}|}
\tabucline{2-}
\multicolumn{1}{c|}{} & \multicolumn{2}{c|}{\bf Schedule} & \multicolumn{2}{c|}{\tstrut\bf OneIG alignment} \\
\multicolumn{1}{c|}{} & {Stochasticity} & {Weighting} & \multicolumn{1}{c}{\tstrut \ScEpoch{} epochs} & \multicolumn{1}{c|}{Highest} \\
\tabucline{-}\tstrut%
{\scalebox{.9}{U}}  & {Uniform}  & {--}   & {\ScFmt{\ScFixU}}  & {\ScFmt{\ScMaxU}} \\
{\scalebox{.9}{UL}} & {Uniform}  & {Low}  & {\ScFmt{\ScFixUL}} & {\ScFmt{\ScMaxUL}} \\
{\scalebox{.9}{UH}} & {Uniform}  & {High} & {\ScFmt{\ScFixUH}} & {\ScFmB{\ScMaxUH}} \\
\tabucline{-}\tstrut%
{\scalebox{.9}{I}}  & {Interval} & {--}   & {\ScFmB{\ScFixI}}  & {\ScFmt{\ScMaxI}} \\
{\scalebox{.9}{IL}} & {Interval} & {Low}  & {\ScFmt{\ScFixIL}} & {\ScFmt{\ScMaxIL}} \\
{\scalebox{.9}{IH}} & {Interval} & {High} & {\ScFmt{\ScFixIH}} & {\ScFmt{\ScMaxIH}} \\
\tabucline{-}\tstrut%
{\scalebox{.9}{P}}  & {Prior}    & {--}   & {\ScFmt{\ScFixP}}  & {\ScFmt{\ScMaxP}} \\
{\scalebox{.9}{PL}} & {Prior}    & {Low}  & {\ScFmt{\ScFixPL}} & {\ScFmt{\ScMaxPL}} \\
{\scalebox{.9}{PH}} & {Prior}    & {High} & {\ScFmt{\ScFixPH}} & {\ScFmt{\ScMaxPH}} \\
\tabucline{-}
\end{tabu}%
}%
}%

\newcommand{\figAppSchedulesPlotU}{%
\centering\footnotesize%
\begin{tikzpicture}%
\begin{axis}[
  width={1.292\linewidth}, height={52.5mm}, grid={major},
  xmin={0}, xmax={1001}, xmode={linear}, x coord trafo/.code=\pgfmathparse{##1^0.7}, xtick={0, 100, 300, 600, 900}, xticklabels={$0$, $100$, $300$, $600$, \hs{0}$900$},
  ymin={57}, ymax={85.7}, ymode={linear}, y coord trafo/.code=\pgfmathparse{##1^2.0}, ytick={60, 65, 70, 75, 80, 85}, yticklabels={$60$, $65$, $70$, $75$, $80$, \raisebox{-0.7mm}[0mm][0mm]{$85$}},
  legend pos={south east}, legend cell align={left},
]
\gdef\pmx##1{\MinorLine{(##1,57)(##1,85.7)}}\pmx{200}\pmx{400}\pmx{500}\pmx{700}\pmx{800}

\gdef\rid{\ScRidU}
\PlotCurve{C0}{\rid::\ScXMid}{\rid::\ScYMid}{(\X*\ScXMul,\Y*\ScYMul)}
\PlotDataLabelAtX{C0}{\rid::\ScXMid}{\rid::\ScYMid}{\ScEpoch}{\YY}{(\ScEpoch*\ScXMul,\YY*\ScYMul)}{south}{}
\PlotDataLabelAtMax{C0}{\rid::\ScXMid}{\rid::\ScYMid}{\XX}{\YY}{(\XX*\ScXMul,\YY*\ScYMul)}{south}{}

\gdef\rid{\ScRidUL}
\PlotCurve{C1}{\rid::\ScXMid}{\rid::\ScYMid}{(\X*\ScXMul,\Y*\ScYMul)}
\PlotDataLabelAtX{C1}{\rid::\ScXMid}{\rid::\ScYMid}{\ScEpoch}{\YY}{(\ScEpoch*\ScXMul,\YY*\ScYMul)}{south}{}
\PlotDataLabelAtMax{C1}{\rid::\ScXMid}{\rid::\ScYMid}{\XX}{\YY}{(\XX*\ScXMul,\YY*\ScYMul)}{south}{}

\gdef\rid{\ScRidUH}
\PlotCurve{C2}{\rid::\ScXMid}{\rid::\ScYMid}{(\X*\ScXMul,\Y*\ScYMul)}
\PlotDataLabelAtX{C2}{\rid::\ScXMid}{\rid::\ScYMid}{\ScEpoch}{\YY}{(\ScEpoch*\ScXMul,\YY*\ScYMul)}{south}{}
\PlotDataLabelAtMax{C2}{\rid::\ScXMid}{\rid::\ScYMid}{\XX}{\YY}{(\XX*\ScXMul,\YY*\ScYMul)}{south}{}

\legend{U, UL, UH}
\end{axis}
\end{tikzpicture}%
}%

\newcommand{\figAppSchedulesPlotI}{%
\centering\footnotesize%
\begin{tikzpicture}%
\begin{axis}[
  width={1.292\linewidth}, height={52.5mm}, grid={major},
  xmin={0}, xmax={1001}, xmode={linear}, x coord trafo/.code=\pgfmathparse{##1^0.7}, xtick={0, 100, 300, 600, 900}, xticklabels={$0$, $100$, $300$, $600$, \hs{0}$900$},
  ymin={57}, ymax={85.7}, ymode={linear}, y coord trafo/.code=\pgfmathparse{##1^2.0}, ytick={60, 65, 70, 75, 80, 85}, yticklabels={$60$, $65$, $70$, $75$, $80$, \raisebox{-0.7mm}[0mm][0mm]{$85$}},
  legend pos={south east}, legend cell align={left},
]
\gdef\pmx##1{\MinorLine{(##1,57)(##1,85.7)}}\pmx{200}\pmx{400}\pmx{500}\pmx{700}\pmx{800}

\gdef\rid{\ScRidI}
\PlotCurve{C0}{\rid::\ScXMid}{\rid::\ScYMid}{(\X*\ScXMul,\Y*\ScYMul)}
\PlotDataLabelAtX{C0}{\rid::\ScXMid}{\rid::\ScYMid}{\ScEpoch}{\YY}{(\ScEpoch*\ScXMul,\YY*\ScYMul)}{south}{}
\PlotDataLabelAtMax{C0}{\rid::\ScXMid}{\rid::\ScYMid}{\XX}{\YY}{(\XX*\ScXMul,\YY*\ScYMul)}{south}{}

\gdef\rid{\ScRidIL}
\PlotCurve{C1}{\rid::\ScXMid}{\rid::\ScYMid}{(\X*\ScXMul,\Y*\ScYMul)}
\PlotDataLabelAtX{C1}{\rid::\ScXMid}{\rid::\ScYMid}{\ScEpoch}{\YY}{(\ScEpoch*\ScXMul,\YY*\ScYMul)}{south}{}
\PlotDataLabelAtMax{C1}{\rid::\ScXMid}{\rid::\ScYMid}{\XX}{\YY}{(\XX*\ScXMul,\YY*\ScYMul)}{south}{}

\gdef\rid{\ScRidIH}
\PlotCurve{C2}{\rid::\ScXMid}{\rid::\ScYMid}{(\X*\ScXMul,\Y*\ScYMul)}
\PlotDataLabelAtX{C2}{\rid::\ScXMid}{\rid::\ScYMid}{\ScEpoch}{\YY}{(\ScEpoch*\ScXMul,\YY*\ScYMul)}{south}{}
\PlotDataLabelAtMax{C2}{\rid::\ScXMid}{\rid::\ScYMid}{\XX}{\YY}{(\XX*\ScXMul,\YY*\ScYMul)}{south}{}

\legend{I, IL, IH}
\end{axis}
\end{tikzpicture}%
}%

\newcommand{\figAppSchedulesPlotP}{%
\centering\footnotesize%
\begin{tikzpicture}%
\begin{axis}[
  width={1.292\linewidth}, height={52.5mm}, grid={major},
  xmin={0}, xmax={1001}, xmode={linear}, x coord trafo/.code=\pgfmathparse{##1^0.7}, xtick={0, 100, 300, 600, 900}, xticklabels={$0$, $100$, $300$, $600$, \hs{0}$900$},
  ymin={57}, ymax={85.7}, ymode={linear}, y coord trafo/.code=\pgfmathparse{##1^2.0}, ytick={60, 65, 70, 75, 80, 85}, yticklabels={$60$, $65$, $70$, $75$, $80$, \raisebox{-0.7mm}[0mm][0mm]{$85$}},
  legend pos={south east}, legend cell align={left},
]
\gdef\pmx##1{\MinorLine{(##1,57)(##1,85.7)}}\pmx{200}\pmx{400}\pmx{500}\pmx{700}\pmx{800}

\gdef\rid{\ScRidP}
\PlotCurve{C0}{\rid::\ScXMid}{\rid::\ScYMid}{(\X*\ScXMul,\Y*\ScYMul)}
\PlotDataLabelAtX{C0}{\rid::\ScXMid}{\rid::\ScYMid}{\ScEpoch}{\YY}{(\ScEpoch*\ScXMul,\YY*\ScYMul)}{south}{}
\PlotDataLabelAtMax{C0}{\rid::\ScXMid}{\rid::\ScYMid}{\XX}{\YY}{(\XX*\ScXMul,\YY*\ScYMul)}{south}{}

\gdef\rid{\ScRidPL}
\PlotCurve{C1}{\rid::\ScXMid}{\rid::\ScYMid}{(\X*\ScXMul,\Y*\ScYMul)}
\PlotDataLabelAtX{C1}{\rid::\ScXMid}{\rid::\ScYMid}{\ScEpoch}{\YY}{(\ScEpoch*\ScXMul,\YY*\ScYMul)}{south}{}
\PlotDataLabelAtMax{C1}{\rid::\ScXMid}{\rid::\ScYMid}{\XX}{\YY}{(\XX*\ScXMul,\YY*\ScYMul)}{south}{}

\gdef\rid{\ScRidPH}
\PlotCurve{C2}{\rid::\ScXMid}{\rid::\ScYMid}{(\X*\ScXMul,\Y*\ScYMul)}
\PlotDataLabelAtX{C2}{\rid::\ScXMid}{\rid::\ScYMid}{\ScEpoch}{\YY}{(\ScEpoch*\ScXMul,\YY*\ScYMul)}{south}{}
\PlotDataLabelAtMax{C2}{\rid::\ScXMid}{\rid::\ScYMid}{\XX}{\YY}{(\XX*\ScXMul,\YY*\ScYMul)}{south}{}

\legend{P, PL, PH}
\end{axis}
\end{tikzpicture}%
}%

\newcommand{\figAppSchedulesPlots}{%
\begin{figure*}[t]
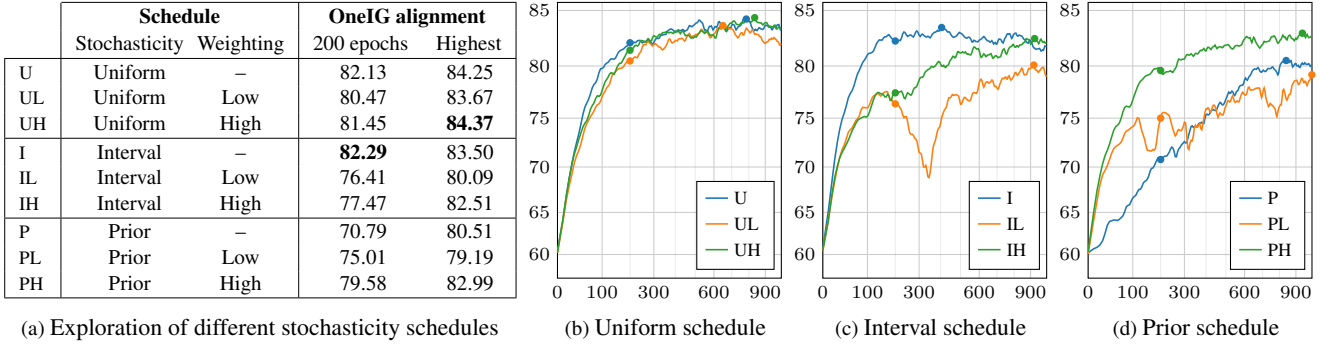
%
\gdef\h{0.390\linewidth}%
\gdef\hh{0.202\linewidth}%
\gdef\vv{39.4mm}%
\gdef\vvv{1ex}%
\begin{subfigure}[b]{\h}%
  \makebox(\linewidth,\vv)[t]{\begin{minipage}{\linewidth}\figAppSchedulesTable\end{minipage}}%
  \vspace{\vvv}\caption{\small Exploration of different stochasticity schedules}%
\end{subfigure}\hfill%
\begin{subfigure}[b]{\hh}%
  \makebox(\linewidth,\vv)[t]{\begin{minipage}{\linewidth}\figAppSchedulesPlotU\end{minipage}}%
  \vspace{\vvv}\caption{\small Uniform schedule \hspace*{-1.2em}}%
\end{subfigure}%
\begin{subfigure}[b]{\hh}%
  \makebox(\linewidth,\vv)[t]{\begin{minipage}{\linewidth}\figAppSchedulesPlotI\end{minipage}}%
  \vspace{\vvv}\caption{\small Interval schedule \hspace*{-1.2em}}%
\end{subfigure}%
\begin{subfigure}[b]{\hh}%
  \makebox(\linewidth,\vv)[t]{\begin{minipage}{\linewidth}\figAppSchedulesPlotP\end{minipage}}%
  \vspace{\vvv}\caption{\small Prior schedule \hspace*{-1.2em}}%
\end{subfigure}%
\vspace{-1mm}%
\caption{\label{figAppSchedulesPlots}%
\textbf{(a)}
Results for our three stochasticity schedules using the combined reward, with optional gradient weighting.
We report the OneIG-Bench alignment at \ScEpoch{} epochs as well as the highest achieved score.
Our main method corresponds to the first row.
\textbf{(b--d)}~%
Convergence of OneIG-Bench alignment for each schedule, annotated with dots that correspond to the numbers shown in the table.
}%
\end{figure*}
}%

\newcommand{\figAppSchedulesImages}{%
\begin{figure*}[t]%
\small\centering%
\gdef\h{0.165\linewidth}%
\gdef\Epoch{e0200}%
\gdef\Title##1{\makebox[\h][c]{##1}}
\gdef\Prompt##1##2{%
  \Image{\h}{\ScRidU/\Epoch-##1.jpg}%
  \hfill\Image{\h}{\ScRidUH/\Epoch-##1.jpg}%
  \hfill\Image{\h}{\ScRidI/\Epoch-##1.jpg}%
  \hfill\Image{\h}{\ScRidIH/\Epoch-##1.jpg}%
  \hfill\Image{\h}{\ScRidP/\Epoch-##1.jpg}%
  \hfill\Image{\h}{\ScRidPH/\Epoch-##1.jpg}\\[-1mm]
  \emph{``##2''}\\[1.5mm]
}
\Title{{\bf U}: Uniform}%
\hfill\Title{{\bf UH}: Uniform $+$ High}%
\hfill\Title{{\bf I}: Interval}%
\hfill\Title{{\bf IH}: Interval $+$ High}%
\hfill\Title{{\bf P}: Prior}%
\hfill\Title{{\bf PH}: Prior $+$ High}\\[1mm]
\Prompt{p001-s04}{snail}
\Prompt{p043-s04}{Lunch in Bavaria - oil painting}
\Prompt{p101-s06}{victorian constable, influenced by warthog features, located in a dark and gloomy victorian England,\\human-like, corrupt, smug, fat, in the london ghetto, anime realism}
\Prompt{p131-s03}{A sailboat drifts lazily along the river, with swans paddling gently beside it and willows weeping at its banks.}
\Prompt{p135-s05}{A bridge arching over a sparkling river, connecting two bustling districts.}
\Prompt{p142-s00}{A broken schoolbag with orange and green colors.}
\vspace{-1mm}%
\caption{\label{figAppSchedulesImages}%
Effect of different stochasticity schedules (from Fig.~\ref{figAppSchedulesPlots}) on the visual look of different prompts.
Each column was post-trained for 200 epochs using our method with the combined reward without CFG.
}%
\end{figure*}
}%

\newcommand{\algOurs}{%
\begin{algorithm}[t]
\footnotesize
\captionof{algorithm}{\atphantom\ \ Our training algorithm}
\Q \textbf{procedure} \textsc{SampleRollouts}$(\theta)$\\
\QQ $C \gets \text{SamplePrompts}(n_{\text{prompts}})$\comm{Draw a set of prompts}\\
\QQ $S\gets\emptyset$ \comm{Initialize set of samples}\\
\QQ \textbf{for each} $\boldc \in C$ \textbf{do} \comm{Loop over prompts}\\
\QQQ \textbf{sample} $\noisedraw \sim \mathcal{N} ( \boldzero,\boldI )$ \comm{Sample initial noise}\\
\QQQ $      \gamma_{0:T} \gets \text{SampleStochSched}(T)$ \comm{Draw random}\\
\QQQ $\hat{\gamma}_{0:T} \gets \text{SampleStochSched}(T)$ \comx{stochasticity schedules}\\
\QQQ $\boldx_{0:T},\boldv_{0:T-1} \gets \text{StochasticFlowSampler}(\theta, \noisedraw, \boldc, \gamma_{0:T})$\\
\QQQ $\boldxhat_{0:T},\boldvhat_{0:T-1} \gets \text{StochasticFlowSampler}(\theta, \noisedraw, \boldc, \hat{\gamma}_{0:T})$\\
\QQQ $\DeltaR \gets R(\boldxhat_T, \boldc) - R(\boldx_T, \boldc)$ \comm{Reward difference}\\
\QQQ $\Deltax \gets \boldxhat_T - \boldx_T$ \comm{Image difference}\\
\QQQ $\nDeltax \gets \Deltax / \smash{\big( \|\Delta \boldx\|_\mathrm{RMS}^2 + 10^{-6} \big)}$ \comm{Normalize $\Deltax$}\\
\QQQ \textbf{for each} $t \in \{0,\ldots,T-1\}$ \textbf{do}\\
\QQQQ $S \getsu ( \boldx_t, \boldv_t, t, \boldc, \DeltaR, \nDeltax)$\comm{Save all samples}\\
\QQQQ $S \getsu ( \boldxhat_t, \boldvhat_t, t, \boldc, \DeltaR, \nDeltax)$\comx{from both trajectories}\\
\QQ \textbf{return} $S$\\[.8ex]
\Q \textbf{procedure} \textsc{TrainEpoch}$(\theta)$\\
\QQ $\text{samples} \gets \text{SampleRollouts}(\theta)$\comm{Sample rollouts}\\
\QQ $\text{batches} \gets \text{RandomSplit}(\text{samples}, n_{\text{batches}})$\comm{Split into batches}\\
\QQ \textbf{for each} $\text{batch} \in \text{batches}$ \textbf{do}\comm{Loop over batches}\\
\QQQ $g \gets 0$\comm{Gradient accumulator}\\
\QQQ \textbf{for each} $\text{sample} \in \text{batch}$ \textbf{do}\\
\QQQQ $( \boldx, \boldv_\mathit{\!ref}, t, \boldc, \DeltaR, \nDeltax) \gets \text{sample}$\\
\QQQQ $\boldv_\mathit{\!cur} \gets \vel(\boldx; t, \boldc)$\\
\QQQQ $L \gets \text{SPO}(\boldv_\mathit{\!cur}, \boldv_\mathit{\!ref}, \DeltaR, \nDeltax)$\\
\QQQQ $g \gets g + \nabla_{\!\theta}L$\comm{Accumulate gradients}\\
\QQQ $\theta \gets \text{AdamWUpdate}(\theta,g)$\comm{Apply gradients}\\
\QQ \textbf{return} $\theta$\\[.8ex]
\Q \textbf{procedure} \textsc{SPO}$(\boldv_\mathit{\!cur}, \boldv_\mathit{\!ref}, \DeltaR, \nDeltax)$\comm{Adapted from \cite{Xie2025SPO}}\\
\QQ $\boldv_\mathit{\!target} \gets \boldv_\mathit{\!ref} - \nDeltax\vphantom{\big{(}}$\\
\QQ $\rho\gets\exp\!\big(
  \|\boldv_\mathit{\!target}-\boldv_\mathit{\!ref}\|_2^2 - 
  \|\boldv_\mathit{\!target}-\boldv_\mathit{\!cur}\|_2^2
  \big)$\\
\QQ $L \gets {-}\DeltaR \cdot \rho - \big(|\Delta R| / (2\epsilon)\big) \cdot (\rho - 1)^2$\\
\QQ \textbf{return} $L\vphantom{\big{(}}$
\label{alg:ours}
\end{algorithm}%
}%

\newcommand{\topstrut}{\vphantom{$\tilde{A}$}}
\newcommand{\TabHyperparameters}{%
\begin{table}[t]%
\centering%
\footnotesize%
\begin{tabular}{l@{\hspace*{3em}}c}
\toprule\addlinespace[.3ex]
\textbf{Hyperparameter} & \textbf{Value} \\
\hline\multicolumn{2}{c}{\topstrut\textit{Model Settings}}\\\hline
LoRA rank                     & 32\topstrut{} \\
LoRA alpha                    & 64    \\
Number format                 & fp16 (+ scaling) \\
\hline\multicolumn{2}{c}{\topstrut\textit{Training Settings}}\\\hline
Prompts/pairs per epoch       & 432\topstrut{} \\
Sampling trajectories per epoch & 864   \\
Denoising steps per trajectory    & 40    \\
Denoising steps per epoch     & 40$\times$864 ${=}$ 34,560 \\
Training batch size           & 216   \\
Training batches per epoch    & 160   \\
Gradient accumulation         & 40 batches    \\
Model weight updates per epoch & 4     \\
Clip range                    & 3$\cdot$10$^{\text{\textminus{}2}}$ \\
Learning rate                 & 3$\cdot$10$^{\text{\textminus{}5}}$ \\
Weight decay                  & 1$\cdot$10$^{\text{\textminus{}4}}$ \\
Gradient clip norm            & 1  \\
Ratio clipping                & SPO \cite{Xie2025SPO} \\
Advantage/reward clipping     & None  \\
KL early stopping             & None  \\
\hline\multicolumn{2}{c}{\topstrut\textit{Evaluation Settings}}\\\hline
ODE step count                & 40\topstrut{} \\
ODE step type                 & Euler, deterministic \\
ODE step sizes                & Logit-normal (SD3.5 \cite{StableDiffusion35}) \\
\addlinespace[.3ex]\bottomrule
\end{tabular}
\caption{\label{TabHyperparameters}%
RL training hyperparameters for our method.
}%
\end{table}%
}%

\newcommand{\tabAppWallclock}{%
\begin{table*}[t]%
\centering\footnotesize%
\begin{tabu}{|l|cc|cc|cc|cc|cc|}
\tabucline{-}
\multirow{2}{*}{\bf Item} & \multicolumn{2}{c|}{\tstrut\bf H200 benchmark}
  & \multicolumn{2}{c|}{\bf Ours, 40 steps}
  & \multicolumn{2}{c|}{\bf Flow-GRPO, 40 steps}
  & \multicolumn{2}{c|}{\bf Ours, 10 steps}
  & \multicolumn{2}{c|}{\bf Flow-GRPO, 10 steps} \\
& {ms/eval} & {batch}
  & {eval/epoch} & {s/epoch}
  & {eval/epoch} & {s/epoch}
  & {eval/epoch} & {s/epoch}
  & {eval/epoch} & {s/epoch} \\
\tabucline{-}\tstrut%
{Denoiser fwd}  & {13.2} & {32}   & {69120}   & {\s912.4}     & {68256}   & {\s901.0}     & {17280}   & {228.1}      & {16416}   & {216.7}       \\
{Denoiser bwd}  & {17.9} & {32}   & {34560}   & {\s618.6}     & {33696}   & {\s603.2}     & {\s8640}  & {154.7}      & {\s7776}  & {139.2}       \\
{VAE decoder}     & {11.0} & {32}   & {\s\s864} & {\s\s\s9.5}   & {\s\s864} & {\s\s\s9.5}   & {\s\s864} & {\s\s9.5}    & {\s\s864} & {\s\s9.5}     \\
{PickScore}       & {18.8} & {32}   & {\s\s864} & {\s\s16.2}    & {\s\s864} & {\s\s16.2}    & {\s\s864} & {\s16.2}     & {\s\s864} & {\s16.2}      \\
{Qwen-7B}         & {35.3} & {26}   & {\s\s864} & {\s\s30.5}    & {\s\s864} & {\s\s30.5}    & {\s\s864} & {\s30.5}     & {\s\s864} & {\s30.5}      \\
\tabucline{-}\tstrut%
{Estimated total} & {}     & {}     & {}        & {\bf 1587.2}  & {}        & {\bf 1560.4}  & {}        & {\bf 439.0}  & {}        & {\bf 412.1}   \\
{Measured total}  & {}     & {}     & {}        & {1625.5}      & {}        & {2736.7}      & {}        & {479.4}      & {}        & {768.4}       \\
{Overhead}        & {}     & {}     & {}        & {\bf +2.4\%}  & {}        & {\bf +75.4\%} & {}        & {\bf +9.2\%} & {}        & {\bf +86.5\%} \\
\tabucline{-}
\end{tabu}%
\vspace{-1mm}%
\caption{\label{tabAppWallclock}%
Details of the wall-clock performance results presented in Section~\refSecWallclock{}.
}%
\end{table*}%
}%

\newcommand{\figComparisonPlot}{%
\begin{figure}[t]%
\centering\footnotesize%
\color{black}%
\hspace*{-.5mm}%
\begin{tikzpicture}%
\begin{axis}[
  width={1.08\columnwidth}, height={70mm}, grid={major},
  xmin={0}, xmax={400}, xmode={linear}, xtick={0, 100, 200, 300, 400}, xticklabels={$0$, $100$k, $200$k, $300$k, \hs{-4}$400$k},
  ymin={19.6}, ymax={24.2}, ymode={linear}, ytick={20, 21, 22, 23, 24}, yticklabels={$20$, $21$, $22$, $23$, $24$},
  xlabel={Generated training images}, ylabel={PickScore},
  legend pos={south east}, legend cell align={left}, legend columns={2}, legend style={nodes={scale=0.95, transform shape}, row sep=-1pt},
]
\gdef\pmx##1{\MinorLine{(##1,19.6)(##1,24.2)}}%
\pmx{50}\pmx{150}\pmx{250}\pmx{350}%
\gdef\pmy##1{\MinorLine{(0,##1)(400,##1)}}%
\pmy{20.5}\pmy{21.5}\pmy{22.5}\pmy{23.5}%

\addlegendentry{FDFO (w/o CFG)}
\PlotCurve{C0}{cmpFDFO::xk}{cmpFDFO::pickscore}{(\X,\Y)}

\addlegendentry{Flow-GRPO (w/ CFG)}
\PlotCurve{C3}{cmpFlowGRPOCfg::xk}{cmpFlowGRPOCfg::pickscore}{(\X,\Y)}

\addlegendentry{Online DPO, $\beta{=}100$}
\PlotCurve{C2}{cmpOnlineDPOB100::xk}{cmpOnlineDPOB100::pickscore}{(\X,\Y)}

\addlegendentry{Flow-GRPO (w/o CFG)}
\PlotCurve{C1}{cmpFlowGRPONoCfg::xk}{cmpFlowGRPONoCfg::pickscore}{(\X,\Y)}

\addlegendentry{ReFL, approx}
\PlotCurveDashed{C0}{cmpReFL::xk}{cmpReFL::pickscore}{(\X,\Y)}

\addlegendentry{Offline DPO, $\beta{=}100$}
\PlotCurveDashed{C2}{cmpOfflineDPOB100::xk}{cmpOfflineDPOB100::pickscore}{(\X,\Y)}

\addlegendentry{DDPO, approx}
\PlotCurveDashed{C3}{cmpDDPO::xk}{cmpDDPO::pickscore}{(\X,\Y)}

\addlegendentry{Offline SFT}
\PlotCurveDashed{C7}{cmpOfflineSFT::xk}{cmpOfflineSFT::pickscore}{(\X,\Y)}

\addlegendentry{Offline RWR}
\PlotCurveDashed{C7!60!black}{cmpOfflineRWR::xk}{cmpOfflineRWR::pickscore}{(\X,\Y)}

\addlegendentry{Online DPO, $\beta{=}10$}
\PlotCurve{C2!70!black}{cmpOnlineDPOB10::xk}{cmpOnlineDPOB10::pickscore}{(\X,\Y)}

\addlegendentry{Online SFT}
\PlotCurve{C7}{cmpOnlineSFT::xk}{cmpOnlineSFT::pickscore}{(\X,\Y)}

\end{axis}
\end{tikzpicture}%
\vspace{-1mm}%
\caption{\label{figComparisonPlot}%
Additional comparison methods evaluated using the PickScore reward as a function of the number of generated training images.
For ReFL and DDPO, which are reported only versus training steps, 
  we estimate the image count, so those comparisons should be considered approximate.
}%
\end{figure}
}%

%% file: appendix-body.tex
\setcounter{figure}{10}
\setcounter{equation}{2}
\setcounter{table}{0}
\setcounter{algorithm}{1}
\setcounter{footnote}{0}

\makeatletter\ifx\c@linenumber\undefined\else
\setcounter{linenumber}{596}
\fi\makeatother

\section{Additional results}
\label{app:results}

\vspace{1mm}%
\paragraph{Training progression and artifacts}
\figAppProgressA
\figAppProgressC
\figAppFlowGRPOWeaknesses
\figAppGuidance
Fig.~\ref{figAppProgressA} shows examples of RL post-training progression (without classifier-free guidance) for our method and Flow-GRPO~\cite{Liu2025_FlowGRPO}.
Our method converges much faster and its subjective image quality peaks around 200 epochs.
Training much longer than that results in loss of diversity and monotonically progressing over-saturation of colors.

Meanwhile, Flow-GRPO slowly improves for approximately 500 epochs,
  and after that, its characteristic artifacts start to appear.
These include grid-like noisy patterns that fade in and out during training, as well as random changes to image styles.
Fig.~\ref{figAppFlowGRPOWeaknesses} shows vertical stripes at epoch~600 and particularly strong horizontal stripes at epoch~830.
The artifacts are tied to the particular epochs and show up in almost all prompts.
The presence of artifacts can vary quickly.
For example, the artifacts in epoch~830 have again disappeared just~5 epochs later at epoch~835. 
Similarly, the style of the images can change quickly.
Epochs~670 and~1000 prefer simplistic cartoony style across most prompts.

\vspace{1mm}%
\paragraph{Comparison to classifier-free guidance (CFG)}
Fig.~\ref{figAppProgressC} shows RL post-training progressions with CFG enabled for our method and Flow-GRPO.
The starting points are of vastly higher quality than without CFG (Fig.~\ref{figAppProgressA}), and RL post-training has only a limited effect, mainly increasing the amount of background detail slightly.

Fig.~\ref{figAppGuidance} shows a direct comparison between our RL post-training and CFG.
RL post-training leads to significantly more detailed images at the same level of diversity.

\vspace{1mm}%
\paragraph{Combining rewards}

\figAppCraftingRewards

We find VLM-based rewards to be a promising way to guide RL post-training.
They offer considerable freedom in designing the visual look without collecting additional data,
  and are likely to keep improving in the future as more capable off-the-shelf VLMs become available.

Here, we experiment with various combinations of VLM-based rewards and the more targeted PickScore
  reward~\cite{kirstain2023} that is designed to mimic human preferences.
Fig.~\ref{figAppCraftingRewards} shows the effect of different post-training reward functions
  on the visual look using a number of prompts.
Details of the reward functions are as follows:
\begin{enumerate}[\bf a), itemsep=4pt, topsep=4pt]
\item PickScore as the only reward function
\item PickScore $+$ VLM alignment$\times0.1$\\
      \emph{\small``Does this image match the caption \texttt{"}$\hspace*{.1em}\cdots\!\hspace*{.1em}$\texttt{"}? Answer Yes or No.''}
\item VLM alignment as the only reward function
\item VLM alignment $+$ VLM quality$\times0.4$\\
      \emph{\small``Is this image of professional quality? Answer with Yes or No.''}
\item VLM alignment $+$ VLM photo\-realism$\times0.4$\\
      \emph{\small``Does this image look photorealistic? Answer Yes or No.''}
\end{enumerate}

Using PickScore reward alone produces a somewhat muted and recognizable AI-generated look,
  which is improved with the addition of the VLM-based prompt alignment goal.
However, VLM prompt alignment alone does not define a consistent style and suffers from obvious quality issues.
Adding a goal that asks for professional quality often leads to non-photorealistic styles,
  whereas asking for photorealism leads to mostly realistic lighting and a consistent visual look.

We note that computing the VLM reward components as separate additive terms seemed to work much better
  in our tests than combining multiple objectives in a single prompt.
It appeared that in the latter case,
  the objective that was easiest to quantify tended to dominate the overall assessment of the image,
  so that the less clear-cut objectives became largely neglected during post-training.

\vspace{1mm}%
\paragraph{Direct human rewards}
\figAppDHRF
Fig.~\ref{figAppDHRF} shows additional post-training progressions using our method with direct on-policy human feedback as a reward (Section~\refSecDHRF).

\vspace{1mm}%
\paragraph{Extended baseline comparison}
\figComparisonPlot
Fig.~\ref{figComparisonPlot} compares our method against Flow-GRPO~\cite{Liu2025_FlowGRPO}, ReFL~\cite{Xu2023_ImageReward_ReFL}, DDPO~\cite{Black2023_DDPO}, 
  Online/Offline DPO~\cite{Wallace2024_DiffusionDPO}, Online/Offline SFT, and Offline RWR~\cite{Black2023_DDPO} 
  on PickScore using Stable Diffusion 3.5 Medium~\cite{StableDiffusion35},
  following the evaluation protocol of Liu et al. \cite{Liu2024_AlignmentDiffusionSurvey}.

The scores are plotted against the number of generated training images.
For methods in the Flow-GRPO family, this is simply the number of training prompts times the group size.
For ReFL and DDPO, we estimate the image count based on the number of training steps.

Our method achieves the best PickScore metric, reaching 23.76 at 400k generated images.
The gap against Online Flow-DPO ($\beta=100$), the strongest DPO-style baseline in this comparison, is 0.61.
PickScore is also the reward on which our method shows the smallest improvement over baselines (Fig.~\refFigRLWorksPlot{}),
  so this represents the conservative case for our gains.

\vspace{1mm}%
\paragraph{Additional metrics}
\figAppMetrics
Fig.~\ref{figAppMetrics} provides a range of benchmark results for our method and Flow-GRPO, using models post-trained using the combined reward. A number of observations can be made.

The constituent parts of the combined reward (PickScore and VLM alignment) are both successfully optimized, rising quickly for us and slower for Flow-GRPO. For our method PickScore starts to drop in the over-training regime where colors are exceedingly saturated.

HPSv2~\cite{wu2023_hpsv2} is an alternative human-preference model, and it tells a very similar story compared to PickScore. CLIP models \cite{Radford2021} do not directly represent human preferences, but also tell a substantially similar story, albeit with a twist that CFG is more strongly preferred.

OneIG-Bench alignment~\cite{chang2025oneig} is highly correlated with our VLM alignment, and OneIG-Bench diversity closely matches the widely used DreamSim diversity~\cite{fu2023dreamsim}.
OneIG-Bench text and style are only tangentially related to the combined reward, and thus weakly affected by the optimization. OneIG-Bench reasoning is another benchmark that seems to prefer CFG.

See Appendix~\ref{app:evaluation} for further details on the evaluation protocol.

\paragraph{KL regularization}
\figAppKLwithoutCfg
Fig.~\ref{figAppKLwithoutCfg} illustrates that our method is compatible with KL regularization.
The primary goal of KL is to preserve more of the diversity, and we can see that it clearly achieves this.
The results depend strongly on the regularization strength, which acts as a tradeoff between the retained diversity and prompt/human preference alignment. 
The tradeoff is qualitatively similar between our method and Flow-GRPO, although our method would need a lower regularization strength to reach its optimal tradeoff.

We implement KL regularization as an L2 distance penalty between the velocity predicted by the base model and model being finetuned with RL. This matches the implementation in prior works \cite{Liu2025_FlowGRPO,Fan2025_Online}.

\section{Implementation details}
\label{app:implementation}

\algOurs

In the following,
  we go into further detail on our stochastic sampler and training algorithms.
In the full training pseudocode, presented in Algorithm~\ref{alg:ours},
  the stochastic sampler (Algorithm~\refAlgStochastic{}) is called twice in the rollout sampling loop
  to construct the pair of trajectories from a common initial noise.
Details of the stochastic sampler are discussed in Section~\ref{app:stochasticity}.
Calls to the ``SampleStochSched'' procedure in Algorithm~\ref{alg:ours}
  correspond to drawing random stochasticity schedules for the trajectories.
These are discussed in Section~\ref{app:schedules},
  followed by further notes on implementation details and evaluation in
  Sections~\ref{app:hyperparams}--\ref{app:wallclock}.

\subsection{Stochastic sampling}
\label{app:stochasticity}

Traditional Euler--Maruyama SDE solvers (e.g., \cite{Song2021sde}) come with drawbacks when applied to denoising diffusion.
They effectively alternate between
  (1) removing noise by an ODE-like diffusion solver step, and 
  (2) adding a fraction of the removed noise back as freshly drawn noise.
However, the exact amount of noise removed and added in these steps is slightly off-sync due to discretization errors,
  leading to a discrepancy that compounds over multiple sampling steps
  (see Karras et~al. \cite{Karras2022elucidating} for a detailed discussion).
Furthermore, the SDE-centric approach binds the noise addition rate to the underlying diffusion time
  and other incidental parameterization details, rather than to the steps themselves.
Some steps might experience very large noise injections that are numerically problematic
  as they extend the length of the associated ODE-like step,
  while others get vanishingly small noises.

We build our stochastic flow matching sampler based on the EDM stochastic sampler \cite{Karras2022elucidating}. 
This sampler departs from strict adherence to an SDE
  and explicitly implements the alternation between the two sub-steps as noise addition and an ODE solver step.
The key insight is to interpret the noise addition sub-step as a genuine jump to a higher value of the time parameter,
  maintaining the correct marginal distribution.
This ensures that the correct information about the noise level, signal scale, and the time parameter are passed to the
  network at the ODE solver sub-step.
Furthermore, the noise injections are scaled in proportion to the existing noise level in the sample.

Specifically, the EDM sampler works by overshooting the ODE step to a lower noise level
  than the deterministic sampler step schedule would indicate
  (roughly, by a fraction $\gamma_i$ of the noise level, where $\gamma_i$ is a per-step hyperparameter),
  and then adding just the right amount of fresh noise to hit the target.

\paragraph{Adaptation to flow matching}

When using unit Gaussian noise as the latent distribution, flow matching can be seen as a special case of a diffusion model.
The usual derivation (e.g., \cite{Liu2022flow}) focuses on the deterministic ODE sampling procedure.
Extensions to an SDE have since been constructed (e.g., \cite{Xue2025_DanceGRPO,Liu2025_FlowGRPO})
  but they suffer from the practical shortcoming discussed above.
We shall thus apply the idea behind EDM sampler directly to flow matching.
The same procedure can be used to derive an EDM-style stochastic sampler for any diffusion schedule
  (e.g., DDPM \cite{Ho2020}, DDIM \cite{Song2020ddim}, VP-SDE and VE-SDE \cite{Song2021sde}, etc.)
  expressed as $\sigma(t)$ and $s(t)$ by following the same steps.

\paragraph{EDM parameterization}

We first note that flow matching can be expressed in the general
  EDM parametrization (see Table~1 in \cite{Karras2022elucidating}) as having:

\begin{itemize}[itemsep=4pt, topsep=4pt]
    \item Scale schedule \mbox{$s(t) = 1-t$}, 
    corresponding to the signal linearly fading out over the unit time interval from $[0,1]$.
    \item Noise schedule \mbox{$\sigma(t) = t/(1-t)$},
    corresponding to the ``noise-to-signal-ratio'' rising from 0 to $\infty$ over the same interval.
\end{itemize}

\noindent
As such, the apparent standard deviation of the noise in the intermediate noisy images, after scaling by $s(t)$,
  increases linearly with $t$, as expected for flow matching.

Let us construct an EDM-like stochastic step from time $t_i$ to $t_{i+1}$ so that
  we replicate the meaning of the hyperparameter $\gamma_i$ from EDM.
For practical implementation reasons related to the\texttt{ diffusers }library abstractions,
  we take the noise reduction step first and add noise second,
  while EDM takes these sub-steps in the opposite order.
We also use a 1\textsuperscript{st}-order Euler ODE solver instead of the 2\textsuperscript{nd}-order
  Heun scheme \cite{Ascher1998} originally used in EDM.

\paragraph{Noise reduction sub-step} The deterministic ODE sampler would simply take an Euler step from time $t_i$ and $t_{i+1}$ and proceed to the next step. In the stochastic version, we overshoot the target time $t_{i+1}$ to a less noisy time $\tilde t_{i+1}$. For this, we must calculate the corrected target noise level.

The EDM sub-step calls for dividing the target noise level by \mbox{$1+\gamma_i$}, where \mbox{$\gamma_i \ge 0$}.
Substituting this into the schedule, this gives the overshoot target noise level
\begin{equation}
\tilde\sigma_{i+1} ~=~ \frac{\sigma(t_{i+1})}{1+\gamma_i}
                   ~=~ \frac{t_{i+1}}{\gamma_i - \gamma_i t_{1+1} - t_{1+1} + 1}\text{.}
\end{equation}
Then, to find the corresponding ODE time, we invert the noise schedule formula, giving $t(\sigma) = \sigma / (\sigma+1)$. Substituting $\tilde \sigma_{i+1}$, the overshoot time becomes
\begin{equation}
\tilde t_{i+1} = \frac{t_{i+1}}{1 - \gamma_i t_{i+1} + \gamma_i}\text{.}
\end{equation}

The first half of Algorithm~\refAlgStochastic{} uses this calculation to take an Euler step for the current image $\boldx_i$ from time $t_i$ to $\tilde t_{i+1}$. Following the ODE velocity for this interval reduces the noise level and scales up the signal accordingly, yielding the intermediate image $\tilde \boldx_{i+1}$.

\paragraph{Noise addition sub-step}

Next, we determine the amount of noise to add in order to reach the noise level corresponding to the endpoint of the solver step, i.e., $\sigma(t_{i+1})$. Because of the non-uniform scale schedule employed by flow matching, we also need to manually re-scale the resulting mixture of signal and noise to match the expected scale at $t_{i+1}$.
Given the intermediate image $\tilde \boldx_{i+1}$ and freshly drawn
  unit Gaussian noise \mbox{$\noisedraw_i\sim\mathcal{N}(\boldzero,\boldI)$},
  the target image is a linear mixture of these with weights that we shall now derive.

The non-uniform scale schedule employed by flow matching makes direct calculation cumbersome.
Let us temporarily remove it by dividing out the scale of the current sample as $\tilde \boldx_{i+1} / s(\tilde t_{i+1})$.
Then, the noise level $\sigma(\tilde t_{i+1})$ directly indicates the standard deviation of the noise present in this image.
To reach the noise level $\sigma(t_{i+1})$, we must add enough noise to cover the difference in the variance (square)
  of the two noise levels, as variance between independent noises adds linearly. 

The gap in variance to cover is \mbox{$\sigma^2(t_{i+1}) - \sigma^2(\tilde t_{i+1})$}, 
  and the standard deviation of the noise needed to do so is the square root of this quantity.
Thus, the unscaled sample with the added noise is {$\tilde \boldx_{i+1} / s(\tilde t_{i+1}) + \sqrt{\sigma^2(t_{i+1}) - \sigma^2(\tilde t_{i+1})} \noisedraw_i$}.
Finally, we reinstate the flow matching scaling at the target noise level by multiplying this mixture by $s(t_{i+1})$.
Substituting everything,
  the formula for the noise addition and scaling comes out as
\begin{equation}
\boldx_{i+1} = \frac
  {\boldxtilde_{i+1} + \tilde\odetime_{i+1} \sqrt{\gamma_i^2+2\gamma_i} \cdot \boldsymbol{\epsilon}_i}
  {\gamma_i \tilde\odetime_{i+1} + 1}\text{,}
\end{equation}
which corresponds to the second half of our stochastic sampling loop (Algorithm~\refAlgStochastic{}).

\subsection{Stochasticity schedules}
\label{app:schedules}

\newcommand{\IntervalS}{\textbf{Interval}}
\newcommand{\UniformS}{\textbf{Uniform}}
\newcommand{\PriorS}{\textbf{Prior}}

\figAppSchedulesPlots
\newcommand\ScColumn[1]{(column~#1)}
\newcommand\ScColumnFig[1]{(Fig.~\ref{figAppSchedulesImages}, column~#1)}

The main role of stochasticity in our method is to generate limited perturbations to an image during the pairwise sampling process.
Given the flexibility of specifying the per-step stochasticity strength parameter $\gamma_i$
  (i.e., the stochasticity schedule)
  and the possibility of randomizing this choice between rollouts,
  there are several ways to achieve this. 
As it is not obvious which approach is preferable, we experimented with three families of randomized stochasticity schedules:

\begin{itemize}[itemsep=4pt, topsep=4pt]
\item \UniformS:
  Specify a uniform, fixed value of $\gamma_i$ used at all time steps for all rollouts. We use the value 0.0025 in all our experiments.
\item \IntervalS:
  For each rollout, pick a random center and width of a smooth interval in $t$. 
  This results in both weak and strong perturbations, depending on the randomly chosen interval width.
  The perturbations affect the image details relevant to the noise levels within the chosen interval.
\item \PriorS:
  Make a random perturbation of randomized magnitude at the initial noise level, i.e., at the prior distribution,
  and run the rest of the sampling trajectory deterministically.
\end{itemize}

\noindent
We also experimented with applying the training gradients using non-uniform weights per time step, as opposed to weighting them equally at all time steps.
The hypothesis was that it might be beneficial to focus the updates primarily to
some specific range of steps.
We tried three variants for each schedule: uniform weighting, focusing low noise levels, and focusing high noise levels.
Specifically, focusing low noise levels assigns a weight for each $t$ according 
  to density function $\lnormal(t; -0.3, 1.0)$ and focusing high noise levels according to
  $\lnormal(t; 0.3, 1.0)$, where $\lnormal(x;\mu,\sigma)$ is the 
  logit-normal distribution:
\begin{equation}
\label{eq:logitnormal}
\lnormal(x; \mu, \sigma) := 
  \frac{1}{\sqrt{2\pi} \sigma x (1\,{-}\,x)} \exp\!\Bigg[\!-\frac{(\mathrm{logit}(x) - \mu)^2}{2 \sigma^2}\Bigg]\text{,}
\end{equation}
where
\begin{equation}
\mathrm{logit}(x) := \log \frac{x}{1-x}
\end{equation}
is the inverse function of a sigmoid.
After evaluating the gradient weights for each $t$, they are normalized to sum to 1.

Fig.~\ref{figAppSchedulesPlots} shows prompt alignment and diversity measurements for the three stochasticity schedules
with the three gradient weighting options. 
The uniform schedule is largely unaffected by gradient weighting
  and is the best option overall.
The other two schedules are quite sensitive to gradient weighting
  and yield inferior results.

Fig.~\ref{figAppSchedulesImages} shows a small set of representative visual results for selected configurations. Our observations below about the subjective image quality for each configuration are based on the results at large, not only to the images included in the figure.
Our baseline method, the uniform schedule without gradient weighting \ScColumn{U}, results in good image quality, variation, and tonal balance.
Enabling gradient weighting \ScColumn{UH} causes the images to be consistently too dark, with somewhat reduced contrast.

\newcommand{\coma}[1]{\hfill\makebox[31.0mm][l]{\commcolor{\commsymbol{} #1}}}
\newcommand{\comax}[1]{\hfill\makebox[31.0mm][l]{\hphantom{\commsymbol} \commcolor{#1}}}

\paragraph{Interval schedule}

The interval schedule is configured by hyperparameters $\mu_\text{center}$ and $\sigma_\text{center}$ for specifying the randomization of the interval center position, 
$\sigma_\text{int}$ for interval width, and $w_\text{int}$ for overall strength.
After the interval center has been determined,
  $\gamma_i$ are assigned using the logit-normal distribution density 
  (Eq.~\ref{eq:logitnormal}).
Denoting the vector of timesteps $[t_0,t_1,\ldots,t_{T-1}]$ as $\mathbf{t}$ and 
  the vector of outputs $[\gamma_0,\gamma_1,\ldots,\gamma_{T-1}]$ as $\boldsymbol{\gamma}$,
the per-rollout stochasticity schedule is drawn as follows.
\begin{algorithm}
\footnotesize
\captionof{algorithm}{\atphantom\ \ Interval stochasticity schedule}
\QQ \textbf{sample} $c \sim \mathcal{N}(\mu_\text{center}, \sigma_\text{center})$\coma{Random interval center} \\
\QQ $\boldsymbol{\gamma} \gets \lnormal(\mathbf{t}; c, \sigma_\text{int})$\coma{Evaluate LN density} \\
\QQ $\boldsymbol{\gamma} \gets \boldsymbol{\gamma} / \mathrm{sum}(\boldsymbol{\gamma})$\coma{Normalize} \\
\QQ $\boldsymbol{\gamma} \gets \mathrm{exp}(w_\text{int}\cdot\boldsymbol{\gamma}) - 1$\coma{Apply constant weight} \\
\QQ \textbf{return} $\boldsymbol{\gamma}$
\end{algorithm}%

\noindent
The normalization ensures that $\gamma_i$ sum to~1 prior to applying the overall strength
  regardless of, e.g., spacing of the time steps within the interval.
The exponentiation and subtraction of~1 on the last line contribute a minor correction that equalizes the total impact of stochastic randomization when it is distributed between several steps.

In our experiments we use hyperparameter values
\mbox{$\mu_\text{center} = 1.3$},
\mbox{$\sigma_\text{center} = 1.5$},
\mbox{$\sigma_\text{int} = 0.25$}, and
\mbox{$w_\text{int} = 3$}.
In a separate experiment, we tried matching the gradient weights to the drawn stochasticity interval, but found this to perform significantly worse.

The interval schedule without gradient weighting \ScColumnFig{I} yields good subjective quality and variation, and excellent contrast. It produces cartoony style more often that the uniform schedule. 
The high quality is also confirmed by the high average alignment scores in Fig.~\ref{figAppSchedulesPlots}a, row~I.
With gradient weighting \ScColumnFig{IH} the images become clearly over-exposed with simplified details, overdoing the cartoony look.

\paragraph{Prior schedule} 
The prior schedule is configured via hyperparameters
  $\mu_{\log w}$ and $\sigma_{\log w}$
  that specify the distribution of the strength of stochasticity in each rollout.
Using log-normal distribution here ensures that the final
  applied strength is always positive.

\begin{algorithm}
\footnotesize
\captionof{algorithm}{\atphantom\ \ Prior stochasticity schedule}
\QQ \textbf{sample} $W\! \sim \mathcal{N}(\mu_{\log w}, \sigma_{\log w})$\coma{Sample log of weight}\\
\QQ $w \gets \exp W$\coma{Random positive weight}\\
\QQ $\boldsymbol{\gamma} \gets \mathbf{0}$\coma{Zero stochasticity} \\
\QQ $\gamma_0 \gets \mathrm{exp}(w) - 1$\coma{.. except at first time step} \\
\QQ \textbf{return} $\boldsymbol{\gamma}$
\end{algorithm}%

In our experiments,
we use values \mbox{$\mu_{\log w} = \log 0.1$} and \mbox{$\sigma_{\log w} = 1.0$}.

Prior schedule without gradient weighting \ScColumnFig{P} produces low-detail images with water color look to them. The contrast and exposure are excessively high.
This is also confirmed by the low average alignment scores in Fig.~\ref{figAppSchedulesPlots}a, row~P.
Enabling gradient weighting \ScColumn{PH} improves the results significantly, yielding a model that has decent image quality and variation. That said, the images are slightly under-exposed, and there is a tendency to exaggerate background details as well.

\figAppSchedulesImages

\subsection{Hyperparameters and practical considerations}
\label{app:hyperparams}

We use the hyperparameters specified in Table~\ref{TabHyperparameters}
  across our main experiments and ablations when running variants of our method.
As explained in Section~\refSecImplementation{},
  we effectively process four training batches of 8640 samples/batch per epoch. 
In practice, we do this by running 40 sub-batches with 216 samples each 
  to conserve memory, accumulating gradients until the batch is complete,
  after which the gradients are used to update the model weights.

\TabHyperparameters{}

\tabAppWallclock

\paragraph{Ratio clipping}

An important detail is how we implement PPO-style ratio clipping. We cannot use tractable likelihoods in our algorithm, as these are generally inaccessible in diffusion and flow models, so we compute a proxy ratio for clipping. Our gradient for the velocity takes the form $\DeltaR\nDeltax$, where \mbox{$\nDeltax = \text{normalize}(\boldxhat_T -\boldx_T)$} is the image difference and \mbox{$\DeltaR = \Reward(\boldxhat_T) - \Reward(\boldx_T)$} is the reward difference. This is analogous to an advantage estimate ($\DeltaR$) times the gradient of a log likelihood ($\nDeltax$) in a policy gradient framework.

We compute a target velocity from each velocity prediction saved during rollouts, \mbox{$\boldv_\mathit{\!target} \gets \boldv_\mathit{\!ref} - \nDeltax\vphantom{\big{(}}$},
  and compose an L2 loss between this target velocity and the model's current velocity prediction, which is analogous to the log-likelihood in Gaussian PPO or the conditional flow matching loss in FPO \cite{mcallister2025flowmatchingpolicygradients}.
We compute our ratio as the exponentiated difference of this loss with the old policy parameters and the current policy parameters, i.e., \mbox{$\exp(\|\boldv_\mathit{\!target}-\boldv_\mathit{\!ref}\|_2^2 - 
  \|\boldv_\mathit{\!target}-\boldv_\mathit{\!cur}\|_2^2)$}. This is a ratio in the same form as the PPO/SPO likelihood ratio, so we can apply clipping in an identical manner. 
The ``SPO'' procedure in Algorithm~\ref{alg:ours} provides a pseudocode implementation.

\subsection{Evaluation protocol}
\label{app:evaluation}

All metrics reported in this paper were calculated \emph{post hoc} based on LoRA checkpoints exported during the corresponding RL training runs.
Specifically, our results never include reward values seen during training, as these could be biased by, e.g., the perturbations to the sampling trajectories.

For the reward and its individual components (e.g., Fig.~\ref{figAppMetrics}\{a,b,c\}), we sample 4,096 random prompts from the Pick-a-Pic~\cite{kirstain2023} training set, which we process the same way as the official Flow-GRPO implementation to yield a total of 25,432 prompts.
To avoid selection bias, we re-randomize the choice of prompts and the per-image random seeds on a per-checkpoint basis.
For consistency with our VLM rewards, we scale the raw value of PickScore by 100.

For HPSv2, CLIP-H/14, CLIP-L/14, and DreamSim diversity (Fig.~\ref{figAppMetrics}\{d,e,f,i\}), we run the official models%
\footnote{\scalebox{0.86}{\url{https://huggingface.co/adams-story/HPSv2-hf}}}%
\footnote{\scalebox{0.86}{\url{https://huggingface.co/laion/CLIP-ViT-H-14-laion2B-s32B-b79K}}}%
\footnote{\scalebox{0.86}{\url{https://huggingface.co/openai/clip-vit-large-patch14}}}%
\footnote{\scalebox{0.86}{\url{https://github.com/ssundaram21/dreamsim}}}
with all 3,200 prompts from the HPDv2 dataset~\cite{wu2023_hpsv2}, generating 4 images for each prompt using different random seeds to improve accuracy.
In line with previous work, we define DreamSim diversity of a pair of images as the sum of squares between their normalized DreamSim embeddings, and average the results over every unique pair of images that were generated using the same prompt.

For OneIG-Bench (e.g., Fig.~\ref{figAppMetrics}\{g,h,j,k,l\}), we use the official implementation%
\footnote{\scalebox{0.86}{\url{https://github.com/OneIG-Bench/OneIG-Benchmark}}}
with all 1,120 prompts and 4 images per prompt.
In terms of reporting, we multiply the raw metric values by 100 for convenience.

\subsection{Wall-clock performance}
\label{app:wallclock}

In Section~\refSecWallclock{}, we compare the wall-clock performance of our method to Flow-GRPO.
As our goal is to compare the two methods, not their implementations,
  we disregard implementation-related overheads and report ideal performance estimates assuming zero overhead.
In practice, our implementation matches these ideal estimates closely, whereas the official implementation of Flow-GRPO is considerably less efficient.
Comparison based on true wall-clock time would thus be needlessly unfair to Flow-GRPO.

The details of our performance estimation procedure are shown in Table~\ref{tabAppWallclock}.
We first benchmark each major computational component in isolation on NVIDIA H200 using representative inputs and reasonably sized batches.
For example, running the denoiser (``Denoiser fwd'') for a batch of 32 noisy latent images,
  corresponding to 512$\times$512 resolution in terms of RGB pixels, takes approximately 13.2\,ms per image on average,
  and backpropagating the resulting gradients to the LoRA weights (``Denoiser bwd'') takes a further 17.9\,ms per image.
A similar measurement is carried out for the remaining components, including the evaluation of the reward functions.

We then instrument each implementation to report the number of times each underlying component is evaluated per epoch.
Tallying everything together, we arrive at the numbers shown in the ``Estimated total'' row,
  which we use to rescale the $x$-axis for each curve in Fig.~\refFigPerformancePlot{}b.

For reference, we also measure the true wall-clock time per epoch taken by each method, averaged over the entire RL training run.
Comparing these timings (``Measured total'') against the
  implementation-neutral estimates,
  we see that our implementation overhead, including logging and checkpoint export, is less than 10\% in both~40 and~10 step cases,
  whereas the official Flow-GRPO implementation has overheads exceeding 75\%.

\section{Theoretical analysis of our method}
\label{app:theory}

In this section, we develop a connection between our finite differences and an approximate gradient of a smoothed version of the reward.

\subsection{Reward gradient}

As noted in Section~\refSecRewardmax{}, we assume there is a reward function $\reward(\boldx)$ that maps images to scalar reward values, with higher values representing more desirable images.
Our goal is to fine-tune the pre-trained velocity function $\vel$ to maximize the expected reward over draws from the generative model
\begin{align}
  \argmax_\theta \mathbb{E}_{\boldc\sim\trainingprompts,\,\boldx_0 \sim \mathcal{N}(\boldzero, \boldI)} \Reward\big(\flowtheta(\boldx_0; \condemb)\big)\text{,}
\end{align}
where $\flowtheta$ refers to the entire sampling process that utilizes the learned velocity function $\vel$.

If $\reward$ were differentiable, we could ascend this objective by taking gradient ascent steps with respect to the weights $\theta$. The gradient is the standard backpropagation
\begin{equation}
    \jactf(\boldx_0; \condemb)\transp \nabla \reward\big(\flowtheta(\boldx_0; \condemb)\big)\text{,}
\end{equation}
where $\jactf$ denotes the Jacobian matrix of $\flowtheta$ with respect to $\theta$, and the gradient $\nabla$ is taken with respect to an image at time $t_T$.
As each Euler ODE step contributes to the shared weight gradient additively, this further decomposes to a sum of terms contributed by individual timesteps
\begin{equation}
  \sum_{i=1}^T (t_{i}-t_{i-1})
  \jactv(\boldx_{i-1}; t_{i-1}, \condemb)\transp 
  \jacxfi(\boldx_{i};\condemb)\transp \nabla\reward(\boldx_T)
  \text{.}
\label{eq:backpropstep}
\end{equation}
Here, $\jactv$ is the Jacobian matrix of $\vel$ with respect to $\theta$,
  \mbox{$\flow_i$ denotes running the ODE solver} starting from an intermediate image at step $i$ to completion,
  and $\jacxfi$ is the Jacobian of this process with respect to said intermediate input image.
In other words, the gradient of the reward at the generated image is backpropagated to the step $i$ through the ODE chain,
  yielding a ``gradient image'' of same format as the noisy image, 
  and then further backpropagated through the step velocity into the weights.

\subsection{Ascending the reward by finite differences}
\label{app:ascending}
We cannot generally expect reward functions to be differentiable, and even when they are, their raw derivatives can behave poorly,
  e.g., by pointing to effective reward hacking directions. 
Our method performs an approximate optimization step that implicitly smooths the reward function and does not require direct gradient information, drawing inspiration from finite differences. 

In the following, we analyze a prototypical version of our method
  where some training-technical practicalities and efficiency improvements have been dropped.
In this variant, we generate a pair of similar but slightly perturbed images and train the
  flow to point towards the higher-reward one:

\begin{enumerate}
\item Draw an initial random noise image $\boldx_0$, and generate the full sequence of intermediate noise images $\{\boldx_i\}_{i=1}^T$ by solving the ODE steps.
\item Choose a random timestep $j$ in $[1, T]$ to train at.
\item Make a normal-distributed perturbation $\pertx_j \sim \mathcal{N}(\boldx_j, \sigma_c^2\boldI)$ of chosen scale $\sigma_c$ around the noisy image $\boldx_j$ at step $j$.
\item Solve the remainder of the ODE starting at the perturbed image to obtain a second generated image $\pertx_T$.
\item Form an approximate gradient
\begin{equation}
\tilde{g} = \frac{\reward(\pertx_T) - \reward(\boldx_T)}{\sigma_c^2}(\pertx_T - \boldx_T) \text{.}
\end{equation}
\item Backpropagate through the velocity at step $j-1$ into weights using the standard procedure (Eq.~\ref{eq:backpropstep}), but with the backpropagated reward gradient replaced by $\tilde{g}$.
\end{enumerate}

\subsection{Approximate gradient}
Let us now show that this prototypical algorithm approximates a gradient direction similar to the gradient of the reward.
To reduce notational clutter, we will drop references to $\theta$ and prompt embeddings $\condemb$.
As we focus on step $j$,
  we also denote $\flow_j$, $\boldx_j$ and $\jacxfj$ as simply $f$, $\boldx$, and $\jac$.

Consider the expectation of our reward-weighted difference around a given $\boldx$ perturbed by a Gaussian noise of standard deviation $\sigma$, and taken through the ODE flow
\begin{equation}
    \mathbb{E}_{\noisedraw \sim \mathcal{N}(0,\sigma^2\boldI)} \big[ \reward\big(\flow(\boldx + \noisedraw)\big) - \reward\big(\flow(\boldx)\big) \big] \big[ \flow(\boldx + \noisedraw) - \flow(\boldx) \big]. \label{eq:gradexpect}
\end{equation}
Assuming that the perturbation is small, we can approximate the mapping of the flow by its first order Taylor expansion $\flow(\boldx) + \boldJ\noisedraw$ for a small perturbation $\noisedraw$ at $\boldx$:
\begin{equation}
    \approx~\mathbb{E}_{\noisedraw \sim \mathcal{N}(0,\sigma^2\boldI)} \big[ \reward\big(\flow(\boldx) + \boldJ \noisedraw \big) - \reward\big(\flow(\boldx)\big) \big] \boldJ\noisedraw \text{.}
\end{equation}
The second term in the bracket vanishes, as $\smash{\reward\big(\flow(\boldx)\big)}$ is a constant with respect to the random variable, and the expectation of a linearly transformed normal distribution is zero.\footnote{The role of this term can be seen as variance reduction when estimating the expectation stochastically, as it provides a baseline reward for the center of the perturbations.} We then make a change of variables to absorb the linear transformation and offset into the random variable:
\begin{equation}
    =~\mathbb{E}_{\noisedraw' \sim \mathcal{N}(\flow(\boldx),\,\sigma^2\boldJ\boldJ\transp)}
    \reward(\noisedraw') \big(\noisedraw' - \flow(\boldx)\big)
    \text{.}
\end{equation}
This form is amenable to applying the multivariate Stein's lemma, yielding
\begin{align}
    &=~ \sigma^2\boldJ\boldJ\transp \mathbb{E}_{\noisedraw' \sim \mathcal{N}(\flow(\boldx),\sigma^2\boldJ\boldJ\transp)} \nabla 
        \reward(\noisedraw') \\
    &=~ \sigma^2\boldJ\boldJ\transp \nabla \mathbb{E}_{\noisedraw' \sim \mathcal{N}(\flow(\boldx),\sigma^2\boldJ\boldJ\transp)} 
        \reward(\noisedraw') \\
    &=~ \sigma^2\boldJ\boldJ\transp \nabla \tilde \reward\big(\flow(\boldx)\big) \text{.} \label{eq:gradapprox}
\end{align}
The expectation can thus be read as the gradient of a smoothed reward function $\tilde \reward$, where its value is averaged over a Gaussian kernel of covariance $\sigma^2\boldJ\boldJ\transp$. The smoothing turns the potentially discontinuous reward into a differentiable one. The smoothed gradient is further transformed by $\boldJ\boldJ\transp$, where we can interpret $\boldJ\transp$ as backpropagating the reward gradient through the ODE to the time of the perturbation\,---\,thus approximating the gradient of the combined flow and reward in Eq.~\ref{eq:backpropstep}\,---\,and the extra $\boldJ$ as distorting this gradient estimate.

\subsection{Characterizing the distortion}
We first note that in practice
the Jacobian of a diffusion flow mapping tends to be close to a positive definite symmetric matrix, where $(\boldJ q)\transp q \ge 0$ for any vector $q$. Then, $\boldJ$ will typically map the gradient to the same half-space, and it remains an ascent direction for the reward. The approximate positive definiteness follows from $\boldJ$ being a composition of gently time-varying, genuinely positive definite Jacobians of the infinitesimal ODE steps. The mapping established by diffusion is also typically close to an optimal transport coupling, which would have an exactly positive definite Jacobian everywhere. 

In practice, many editing-flavored diffusion methods implicitly rely on an argument of this nature;
  they make coarse edits to the intermediate noisy images and expect them to retain their approximate
  pixel position and color in the final generated image, but in a more ``fleshed out'', fully generated form.
In a sense, the dot product between the coarse edit and the difference it makes on the final image
  is empirically expected to be positive, hinting at the Jacobian being positive definite for practical purposes.

\subsection{Discussion}

Equations \ref{eq:gradexpect} to \ref{eq:gradapprox}, taken in isolation, imply the claim in Section~\refSecAnalysis{} that
\begin{equation}
\mathbb{E} \big[ \hnabla \Reward(f(\boldx))\transp \boldJ \left[ \DeltaR\Deltax \right] \big] \ge 0,
\end{equation}
where the expectation is taken over the random perturbations. In other words, the intermediate-time update $\DeltaR \Deltax$, pushed through the remaining flow by $\boldJ$, yields a direction of increasing reward.

By Eq.~\ref{eq:gradapprox}, the left hand side becomes
\begin{equation}
\sigma^2 \hnabla \Reward(f(\boldx))\transp \boldJ \boldJ \boldJ^\top  \nabla \tilde \Reward(f(\boldx))
\end{equation}
which is approximately a quadratic form evaluated with a positive definite matrix $\boldJ \boldJ \boldJ^\top$, and as such, has a nonnegative value.

These results rely on assumptions discussed in Section~\ref{app:ascending}. Our practical algorithm modifies this idealized setting in various ways, motivated by efficiency and simplicity. The ablation with stochastic schedules (Section~\ref{app:schedules}) probes a specific difference: our method applies stochasticity throughout the trajectories instead of just a single step, and applies the update on \emph{all} steps along the trajectories. In contrast, the \IntervalS ~schedule models a situation similar to the idealized method, where the paths only deviate on a localized time interval. The empirical results suggest that a variety of schemes work in practice, and the more ``ideal'' scheme does not enjoy a clear benefit despite being more amenable for direct mathematical analysis.

%% file: paper.bbl
\begin{thebibliography}{68}
\providecommand{\natexlab}[1]{#1}
\providecommand{\url}[1]{\texttt{#1}}
\expandafter\ifx\csname urlstyle\endcsname\relax
  \providecommand{\doi}[1]{doi: #1}\else
  \providecommand{\doi}{doi: \begingroup \urlstyle{rm}\Url}\fi

\bibitem[Albergo et~al.(2025)Albergo, Boffi, and
  Vanden-Eijnden]{albergo2025interpolants}
Michael Albergo, Nicholas Boffi, and Eric Vanden-Eijnden.
\newblock Stochastic interpolants: {A} unifying framework for flows and
  diffusions.
\newblock \emph{JMLR}, 26\penalty0 (209), 2025.

\bibitem[Albergo and Vanden-Eijnden(2023)]{Albergo2023NormalizingFlows}
Michael~S. Albergo and Eric Vanden-Eijnden.
\newblock Building normalizing flows with stochastic interpolants.
\newblock In \emph{Proc. ICLR}, 2023.

\bibitem[Amodei et~al.(2016)Amodei, Olah, Steinhardt, Christiano, Schulman, and
  Man\'e]{Amodei2016hacking}
Dario Amodei, Chris Olah, Jacob Steinhardt, Paul Christiano, John Schulman, and
  Dan Man\'e.
\newblock Concrete problems in {AI} safety.
\newblock \emph{CoRR}, abs/1606.06565, 2016.

\bibitem[Ascher and Petzold(1998)]{Ascher1998}
Uri~M. Ascher and Linda~R. Petzold.
\newblock \emph{Computer Methods for Ordinary Differential Equations and
  Differential-Algebraic Equations}.
\newblock Society for Industrial and Applied Mathematics, 1998.

\bibitem[Bengio et~al.(2021)Bengio, Jain, Korablyov, Precup, and
  Bengio]{Bengio2021_GFlowNets}
Emmanuel Bengio, Moksh Jain, Maksym Korablyov, Doina Precup, and Yoshua Bengio.
\newblock Flow network based generative models for non-iterative diverse
  candidate generation.
\newblock In \emph{Proc. NeurIPS}, 2021.

\bibitem[Black et~al.(2024)Black, Janner, Du, Kostrikov, and
  Levine]{Black2023_DDPO}
Kevin Black, Michael Janner, Yilun Du, Ilya Kostrikov, and Sergey Levine.
\newblock Training diffusion models with reinforcement learning.
\newblock In \emph{Proc. ICLR}, 2024.

\bibitem[Chang et~al.(2025)Chang, Fang, Xing, Wu, Cheng, Wang, Zeng, Yu, and
  Chen]{chang2025oneig}
Jingjing Chang, Yixiao Fang, Peng Xing, Shuhan Wu, Wei Cheng, Rui Wang,
  Xianfang Zeng, Gang Yu, and Hai-Bao Chen.
\newblock {OneIG-Bench}: {O}mni-dimensional nuanced evaluation for image
  generation.
\newblock In \emph{Proc. NeurIPS}, 2025.

\bibitem[Chen et~al.(2024)Chen, Vaxman, Ben~Baruch, Asulin, Moreshet, Lien,
  Sra, and Sen]{chen2024tino}
Sherry~X. Chen, Yaron Vaxman, Elad Ben~Baruch, David Asulin, Aviad Moreshet,
  Kuo-Chin Lien, Misha Sra, and Pradeep Sen.
\newblock {TiNO-Edit}: {T}imestep and noise optimization for robust
  diffusion-based image editing.
\newblock In \emph{Proc. CVPR}, 2024.

\bibitem[Choi et~al.(2021)Choi, Kim, Jeong, Gwon, and Yoon]{choi2021ilvr}
Jooyoung Choi, Sungwon Kim, Yonghyun Jeong, Youngjune Gwon, and Sungroh Yoon.
\newblock {ILVR}: {C}onditioning method for denoising diffusion probabilistic
  models.
\newblock In \emph{Proc. ICCV}, 2021.

\bibitem[Clark et~al.(2024)Clark, Vicol, Swersky, and Fleet]{Clark2024_DRaFT}
Kevin Clark, Paul Vicol, Kevin Swersky, and David~J. Fleet.
\newblock Directly fine-tuning diffusion models on differentiable rewards
  ({DRaFT}).
\newblock In \emph{Proc. ICLR}, 2024.

\bibitem[Delbracio and Milanfar(2023)]{Delbracio2023inversion}
Mauricio Delbracio and Peyman Milanfar.
\newblock Inversion by direct iteration: {A}n alternative to denoising
  diffusion for image restoration.
\newblock \emph{TMLR}, 2023.

\bibitem[Domingo-Enrich et~al.(2025)Domingo-Enrich, Drozdzal, Karrer, and
  Chen]{DomingoEnrich2025_AdjointMatching}
Carles Domingo-Enrich, Michal Drozdzal, Brian Karrer, and Ricky T.~Q. Chen.
\newblock Adjoint matching: {F}ine-tuning flow and diffusion generative models
  with memoryless stochastic optimal control.
\newblock In \emph{Proc. ICLR}, 2025.

\bibitem[Dong et~al.(2023)Dong, Xiong, Goyal, Zhang, Chow, Pan, Diao, Zhang,
  Shum, and Zhang]{Dong2023_RAFT}
Hanze Dong, Wei Xiong, Deepanshu Goyal, Yihan Zhang, Winnie Chow, Rui Pan,
  Shizhe Diao, Jipeng Zhang, Kashun Shum, and Tong Zhang.
\newblock {RAFT}: {R}eward ranked finetuning for generative foundation model
  alignment.
\newblock \emph{TMLR}, 2023.

\bibitem[Esser et~al.(2024)Esser, Kulal, Blattmann, Entezari, M\"{u}ller,
  Saini, Levi, Lorenz, Sauer, Boesel, Podell, Dockhorn, English, and
  Rombach]{StableDiffusion35}
Patrick Esser, Sumith Kulal, Andreas Blattmann, Rahim Entezari, Jonas
  M\"{u}ller, Harry Saini, Yam Levi, Dominik Lorenz, Axel Sauer, Frederic
  Boesel, Dustin Podell, Tim Dockhorn, Zion English, and Robin Rombach.
\newblock Scaling rectified flow transformers for high-resolution image
  synthesis.
\newblock In \emph{Proc. ICML}, 2024.
\newblock \url{https://huggingface.co/stabilityai/stable-diffusion-3.5-medium}
  (accessed 6/23/2026).

\bibitem[Fan et~al.(2025)Fan, Shen, Cheng, Chen, Liang, and
  Liu]{Fan2025_Online}
Jiajun Fan, Shuaike Shen, Chaoran Cheng, Yuxin Chen, Chumeng Liang, and Ge Liu.
\newblock Online reward-weighted fine-tuning of flow matching with
  {Wasserstein} regularization.
\newblock In \emph{Proc. ICLR}, 2025.

\bibitem[Fan et~al.(2023)Fan, Watkins, Du, Liu, Ryu, Boutilier, Abbeel,
  Ghavamzadeh, Lee, and Lee]{10.5555/3666122.3669619}
Ying Fan, Olivia Watkins, Yuqing Du, Hao Liu, Moonkyung Ryu, Craig Boutilier,
  Pieter Abbeel, Mohammad Ghavamzadeh, Kangwook Lee, and Kimin Lee.
\newblock {DPOK}: {R}einforcement learning for fine-tuning text-to-image
  diffusion models.
\newblock In \emph{Proc. NeurIPS}, 2023.

\bibitem[Fu et~al.(2023)Fu, Tamir, Sundaram, Chai, Zhang, Dekel, and
  Isola]{fu2023dreamsim}
Stephanie Fu, Netanel Tamir, Shobhita Sundaram, Lucy Chai, Richard Zhang, Tali
  Dekel, and Phillip Isola.
\newblock {DreamSim}: Learning new dimensions of human visual similarity using
  synthetic data.
\newblock In \emph{Proc. NeurIPS}, 2023.

\bibitem[Guo et~al.(2024)Guo, Yuan, Yang, Chen, and Wang]{GGD_2024}
Yingqing Guo, Hui Yuan, Yukang Yang, Minshuo Chen, and Mengdi Wang.
\newblock Gradient guidance for diffusion models: {A}n optimization
  perspective.
\newblock In \emph{Proc. NeurIPS}, 2024.

\bibitem[Heitz et~al.(2023)Heitz, Belcour, and Chambon]{Heitz2023}
Eric Heitz, Laurent Belcour, and Thomas Chambon.
\newblock Iterative $\alpha$-(de)blending: {A} minimalist deterministic
  diffusion model.
\newblock In \emph{Proc. SIGGRAPH}, 2023.

\bibitem[Ho and Salimans(2021)]{Ho2021classifierfree}
Jonathan Ho and Tim Salimans.
\newblock Classifier-free diffusion guidance.
\newblock In \emph{Proc. NeurIPS 2021 Workshop on Deep Generative Models and
  Downstream Applications}, 2021.

\bibitem[Ho et~al.(2020)Ho, Jain, and Abbeel]{Ho2020}
Jonathan Ho, Ajay Jain, and Pieter Abbeel.
\newblock Denoising diffusion probabilistic models.
\newblock In \emph{Proc. NeurIPS}, 2020.

\bibitem[Hu et~al.(2022)Hu, Shen, Wallis, Allen-Zhu, Li, Wang, Wang, and
  Chen]{Hu2022lora}
Edward~J Hu, Yelong Shen, Phillip Wallis, Zeyuan Allen-Zhu, Yuanzhi Li, Shean
  Wang, Lu Wang, and Weizhu Chen.
\newblock Lo{RA}: Low-rank adaptation of large language models.
\newblock In \emph{Proc. ICLR}, 2022.

\bibitem[Jia et~al.(2025)Jia, Nan, Zhao, and Liu]{Jia2025_LaSRO}
Zhiwei Jia, Yuesong Nan, Huixi Zhao, and Gengdai Liu.
\newblock Reward fine-tuning two-step diffusion models via learning
  differentiable latent-space surrogate reward.
\newblock In \emph{Proc. CVPR}, 2025.

\bibitem[Karras et~al.(2022)Karras, Aittala, Aila, and
  Laine]{Karras2022elucidating}
Tero Karras, Miika Aittala, Timo Aila, and Samuli Laine.
\newblock Elucidating the design space of diffusion-based generative models.
\newblock In \emph{Proc. NeurIPS}, 2022.

\bibitem[Khrulkov et~al.(2023)Khrulkov, Ryzhakov, Chertkov, and
  Oseledets]{khrulkov2022understandingddpmlatentcodes}
Valentin Khrulkov, Gleb Ryzhakov, Andrei Chertkov, and Ivan Oseledets.
\newblock Understanding {DDPM} latent codes through optimal transport.
\newblock In \emph{Proc. ICLR}, 2023.

\bibitem[Kim et~al.(2022)Kim, Kwon, and Ye]{kim2022diffusionclip}
Gwanghyun Kim, Taesung Kwon, and Jong~Chul Ye.
\newblock {DiffusionCLIP}: {T}ext-guided diffusion models for robust image
  manipulation.
\newblock In \emph{Proc. CVPR}, 2022.

\bibitem[Kirstain et~al.(2023)Kirstain, Polyak, Singer, Matiana, Penna, and
  Levy]{kirstain2023}
Yuval Kirstain, Adam Polyak, Uriel Singer, Shahbuland Matiana, Joe Penna, and
  Omer Levy.
\newblock {Pick-a-Pic}: {A}n open dataset of user preferences for text-to-image
  generation.
\newblock In \emph{Proc. NeurIPS}, 2023.
\newblock \url{https://huggingface.co/yuvalkirstain/PickScore_v1} (accessed
  6/23/2026).

\bibitem[Lavenant and Santambrogio(2022)]{Lavenant2022flowmap}
Hugo Lavenant and Filippo Santambrogio.
\newblock The flow map of the {F}okker--{P}lanck equation does not provide
  optimal transport.
\newblock \emph{Appl. Math. Lett.}, 133, 2022.

\bibitem[Lee et~al.(2023)Lee, Liu, Ryu, Watkins, Du, Boutilier, Abbeel,
  Ghavamzadeh, and Gu]{Lee2023_AligningTextToImage}
Kimin Lee, Hao Liu, Moonkyung Ryu, Olivia Watkins, Yuqing Du, Craig Boutilier,
  Pieter Abbeel, Mohammad Ghavamzadeh, and Shixiang Gu.
\newblock Aligning text-to-image models using human feedback.
\newblock \emph{CoRR}, abs/2302.12192, 2023.

\bibitem[Levine(2018)]{levine2018reinforcementlearningcontrolprobabilistic}
Sergey Levine.
\newblock Reinforcement learning and control as probabilistic inference:
  {T}utorial and review.
\newblock \emph{CoRR}, abs/1805.00909, 2018.

\bibitem[Lin et~al.(2024)Lin, Pathak, Li, Li, Xia, Neubig, Zhang, and
  Ramanan]{lin2024vqascore}
Zhiqiu Lin, Deepak Pathak, Baiqi Li, Jiayao Li, Xide Xia, Graham Neubig,
  Pengchuan Zhang, and Deva Ramanan.
\newblock Evaluating text-to-visual generation with image-to-text generation.
\newblock In \emph{Proc. ECCV}, 2024.

\bibitem[Lipman et~al.(2023)Lipman, Chen, Ben-Hamu, Nickel, and
  Le]{Lipman2023FlowMatching}
Yaron Lipman, Ricky T.~Q. Chen, Heli Ben-Hamu, Maximilian Nickel, and Matt Le.
\newblock Flow matching for generative modeling.
\newblock In \emph{Proc. ICLR}, 2023.

\bibitem[Liu et~al.(2026)Liu, Shao, Li, Bai, Xu, Xiong, Kwok, Helal, and
  Xie]{Liu2024_AlignmentDiffusionSurvey}
Buhua Liu, Shitong Shao, Bao Li, Lichen Bai, Zhiqiang Xu, Haoyi Xiong, James~T.
  Kwok, Sumi Helal, and Zeke Xie.
\newblock Alignment of diffusion models: {F}undamentals, challenges, and
  future.
\newblock \emph{ACM Comput. Surv.}, 2026.

\bibitem[Liu et~al.(2025{\natexlab{a}})Liu, Liu, Liang, Li, Liu, Wang, Wan,
  Zhang, and Ouyang]{Liu2025_FlowGRPO}
Jie Liu, Gongye Liu, Jiajun Liang, Yangguang Li, Jiaheng Liu, Xintao Wang,
  Pengfei Wan, Di Zhang, and Wanli Ouyang.
\newblock {Flow-GRPO}: {T}raining flow matching models via online {RL}.
\newblock In \emph{Proc. NeurIPS}, 2025{\natexlab{a}}.
\newblock \url{https://github.com/yifan123/flow_grpo} (accessed 6/23/2026).

\bibitem[Liu et~al.(2023)Liu, Gong, and Liu]{Liu2022flow}
Xingchao Liu, Chengyue Gong, and Qiang Liu.
\newblock Flow straight and fast: {L}earning to generate and transfer data with
  rectified flow.
\newblock In \emph{Proc. ICLR}, 2023.

\bibitem[Liu et~al.(2025{\natexlab{b}})Liu, Xiao, Liu, Bengio, and
  Zhang]{Liu2025_NablaGFlowNet}
Zhen Liu, Tim~Z. Xiao, Weiyang Liu, Yoshua Bengio, and Dinghuai Zhang.
\newblock Efficient diversity-preserving diffusion alignment via
  gradient-informed {GFlowNets}.
\newblock In \emph{Proc. ICLR}, 2025{\natexlab{b}}.

\bibitem[Loshchilov and Hutter(2019)]{Loshchilov2019AdamW}
Ilya Loshchilov and Frank Hutter.
\newblock Decoupled weight decay regularization.
\newblock In \emph{Proc. ICLR}, 2019.

\bibitem[Luo et~al.(2025)Luo, Granskog, Holynski, and
  Darrell]{luo2025dualprocess}
Grace Luo, Jonathan Granskog, Aleksander Holynski, and Trevor Darrell.
\newblock Dual-process image generation.
\newblock In \emph{Proc. ICCV}, 2025.

\bibitem[McAllister et~al.(2026)McAllister, Ge, Yi, Kim, Weber, Choi, Feng, and
  Kanazawa]{mcallister2025flowmatchingpolicygradients}
David McAllister, Songwei Ge, Brent Yi, Chung~Min Kim, Ethan Weber, Hongsuk
  Choi, Haiwen Feng, and Angjoo Kanazawa.
\newblock Flow matching policy gradients.
\newblock In \emph{Proc. ICLR}, 2026.

\bibitem[Meng et~al.(2022)Meng, He, Song, Song, Wu, Zhu, and
  Ermon]{meng2023sdedit}
Chenlin Meng, Yutong He, Yang Song, Jiaming Song, Jiajun Wu, Jun-Yan Zhu, and
  Stefano Ermon.
\newblock {SDEdit}: {G}uided image synthesis and editing with stochastic
  differential equations.
\newblock In \emph{Proc. ICLR}, 2022.

\bibitem[Nachum et~al.(2017)Nachum, Norouzi, Xu, and
  Schuurmans]{Nachum2017_PCL}
Ofir Nachum, Mohammad Norouzi, Kelvin Xu, and Dale Schuurmans.
\newblock Bridging the gap between value and policy based reinforcement
  learning.
\newblock In \emph{Proc. NeurIPS}, 2017.

\bibitem[Nakano et~al.(2021)Nakano, Hilton, Balaji, Wu, Ouyang, Kim, Hesse,
  Jain, Kosaraju, Saunders, Jiang, Cobbe, Eloundou, Krueger, Button, Knight,
  Chess, and Schulman]{nakano2021webgptbrowserassistedquestionansweringhuman}
Reiichiro Nakano, Jacob Hilton, Suchir Balaji, Jeff Wu, Long Ouyang, Christina
  Kim, Christopher Hesse, Shantanu Jain, Vineet Kosaraju, William Saunders, Xu
  Jiang, Karl Cobbe, Tyna Eloundou, Gretchen Krueger, Kevin Button, Matthew
  Knight, Benjamin Chess, and John Schulman.
\newblock {WebGPT}: {B}rowser-assisted question-answering with human feedback.
\newblock \emph{CoRR}, abs/2112.09332, 2021.

\bibitem[Ouyang et~al.(2022)Ouyang, Wu, Jiang, Almeida, Wainwright, Mishkin,
  Zhang, Agarwal, Slama, Ray, Schulman, Hilton, Kelton, Miller, Simens, Askell,
  Welinder, Christiano, Leike, and Lowe]{Ouyang2022_InstructGPT}
Long Ouyang, Jeff Wu, Xu Jiang, Diogo Almeida, Carroll~L. Wainwright, Pamela
  Mishkin, Chong Zhang, Sandhini Agarwal, Katarina Slama, Alex Ray, John
  Schulman, Jacob Hilton, Fraser Kelton, Luke Miller, Maddie Simens, Amanda
  Askell, Peter Welinder, Paul Christiano, Jan Leike, and Ryan Lowe.
\newblock Training language models to follow instructions with human feedback.
\newblock In \emph{Proc. NeurIPS}, 2022.

\bibitem[Park et~al.(2025)Park, Li, and Levine]{fql_park2025}
Seohong Park, Qiyang Li, and Sergey Levine.
\newblock Flow {Q}-learning.
\newblock In \emph{Proc. ICML}, 2025.

\bibitem[Peters and Schaal(2007)]{10.1145/1273496.1273590}
Jan Peters and Stefan Schaal.
\newblock Reinforcement learning by reward-weighted regression for operational
  space control.
\newblock In \emph{Proc. ICML}, 2007.

\bibitem[Psenka et~al.(2024)Psenka, Escontrela, Abbeel, and Ma]{psenka2024qsm}
Michael Psenka, Alejandro Escontrela, Pieter Abbeel, and Yi Ma.
\newblock Learning a diffusion model policy from rewards via {Q}-score
  matching.
\newblock In \emph{Proc. ICML}, 2024.

\bibitem[Radford et~al.(2021)Radford, Kim, Hallacy, Ramesh, Goh, Agarwal,
  Sastry, Askell, Mishkin, Clark, Krueger, and Sutskever]{Radford2021}
Alec Radford, Jong~Wook Kim, Chris Hallacy, Aditya Ramesh, Gabriel Goh,
  Sandhini Agarwal, Girish Sastry, Amanda Askell, Pamela Mishkin, Jack Clark,
  Gretchen Krueger, and Ilya Sutskever.
\newblock Learning transferable visual models from natural language
  supervision.
\newblock In \emph{Proc. ICML}, 2021.
\newblock \url{https://huggingface.co/laion/CLIP-ViT-H-14-laion2B-s32B-b79K}
  and \url{https://huggingface.co/openai/clip-vit-large-patch14} (both accessed
  6/23/2026).

\bibitem[Rafailov et~al.(2023)Rafailov, Sharma, Mitchell, Ermon, Manning, and
  Finn]{Rafailov2023_DPO}
Rafael Rafailov, Archit Sharma, Eric Mitchell, Stefano Ermon, Christopher~D.
  Manning, and Chelsea Finn.
\newblock Direct preference optimization: Your language model is secretly a
  reward model.
\newblock In \emph{Proc. NeurIPS}, 2023.

\bibitem[Rubinstein(1999)]{10.1023/A:1010091220143}
Reuven Rubinstein.
\newblock The cross-entropy method for combinatorial and continuous
  optimization.
\newblock \emph{Method. Comput. Appl. Prob.}, 1\penalty0 (2), 1999.

\bibitem[Santambrogio(2015)]{santambrogio2015optimal}
Filippo Santambrogio.
\newblock \emph{Optimal Transport for Applied Mathematicians: {C}alculus of
  Variations, {PDE}s, and Modeling}.
\newblock Birkh\"auser, 2015.

\bibitem[Schulman et~al.(2017)Schulman, Wolski, Dhariwal, Radford, and
  Klimov]{Schulman2017PPO}
John Schulman, Filip Wolski, Prafulla Dhariwal, Alec Radford, and Oleg Klimov.
\newblock Proximal policy optimization algorithms.
\newblock \emph{CoRR}, abs/1707.06347, 2017.

\bibitem[Sohl-Dickstein et~al.(2015)Sohl-Dickstein, Weiss, Maheswaranathan, and
  Ganguli]{SohlDickstein2015}
Jascha Sohl-Dickstein, Eric Weiss, Niru Maheswaranathan, and Surya Ganguli.
\newblock Deep unsupervised learning using nonequilibrium thermodynamics.
\newblock In \emph{Proc. ICML}, 2015.

\bibitem[Song et~al.(2021{\natexlab{a}})Song, Meng, and Ermon]{Song2020ddim}
Jiaming Song, Chenlin Meng, and Stefano Ermon.
\newblock Denoising diffusion implicit models.
\newblock In \emph{Proc. ICLR}, 2021{\natexlab{a}}.

\bibitem[Song et~al.(2023)Song, Zhang, Yin, Mardani, Liu, Kautz, Chen, and
  Vahdat]{Song2023_LossGuidedDiffusion}
Jiaming Song, Qinsheng Zhang, Hongxu Yin, Morteza Mardani, Ming-Yu Liu, Jan
  Kautz, Yongxin Chen, and Arash Vahdat.
\newblock Loss-guided diffusion models for plug-and-play controllable
  generation.
\newblock In \emph{Proc. ICML}, 2023.

\bibitem[Song and Ermon(2019)]{Song2019gradients}
Yang Song and Stefano Ermon.
\newblock Generative modeling by estimating gradients of the data distribution.
\newblock In \emph{Proc. NeurIPS}, 2019.

\bibitem[Song et~al.(2021{\natexlab{b}})Song, Sohl-Dickstein, Kingma, Kumar,
  Ermon, and Poole]{Song2021sde}
Yang Song, Jascha Sohl-Dickstein, Diederik~P. Kingma, Abhishek Kumar, Stefano
  Ermon, and Ben Poole.
\newblock Score-based generative modeling through stochastic differential
  equations.
\newblock In \emph{Proc. ICLR}, 2021{\natexlab{b}}.

\bibitem[Uehara et~al.(2024)Uehara, Zhao, Biancalani, and
  Levine]{Uehara2024_Understanding}
Masatoshi Uehara, Yulai Zhao, Tommaso Biancalani, and Sergey Levine.
\newblock Understanding reinforcement learning-based fine-tuning of diffusion
  models: {A} tutorial and review.
\newblock \emph{CoRR}, abs/2407.13734, 2024.

\bibitem[Wallace et~al.(2024)Wallace, Dang, Rafailov, Zhou, Lou, Purushwalkam,
  Ermon, Xiong, Joty, and Naik]{Wallace2024_DiffusionDPO}
Bram Wallace, Meihua Dang, Rafael Rafailov, Linqi Zhou, Aaron Lou, Senthil
  Purushwalkam, Stefano Ermon, Caiming Xiong, Shafiq Joty, and Nikhil Naik.
\newblock Diffusion model alignment using direct preference optimization.
\newblock In \emph{Proc. CVPR}, 2024.

\bibitem[Wang et~al.(2024)Wang, Bai, Tan, Wang, Fan, Bai, Chen, Liu, Wang, Ge,
  Fan, Dang, Du, Ren, Men, Liu, Zhou, Zhou, and Lin]{QWEN25}
Peng Wang, Shuai Bai, Sinan Tan, Shijie Wang, Zhihao Fan, Jinze Bai, Keqin
  Chen, Xuejing Liu, Jialin Wang, Wenbin Ge, Yang Fan, Kai Dang, Mengfei Du,
  Xuancheng Ren, Rui Men, Dayiheng Liu, Chang Zhou, Jingren Zhou, and Junyang
  Lin.
\newblock {Qwen2-VL}: {E}nhancing vision-language model's perception of the
  world at any resolution.
\newblock \emph{CoRR}, abs/2409.12191, 2024.
\newblock \url{https://huggingface.co/Qwen/Qwen2.5-VL-7B-Instruct} (accessed
  6/23/2026).

\bibitem[Watkins and Dayan(1992)]{Watkins1992}
Christopher J. C.~H. Watkins and Peter Dayan.
\newblock Q-learning.
\newblock \emph{Mach. Learn.}, 8\penalty0 (3), 1992.

\bibitem[Wu et~al.(2023)Wu, Hao, Sun, Chen, Zhu, Zhao, and Li]{wu2023_hpsv2}
Xiaoshi Wu, Yiming Hao, Keqiang Sun, Yixiong Chen, Feng Zhu, Rui Zhao, and
  Hongsheng Li.
\newblock Human preference score v2: {A} solid benchmark for evaluating human
  preferences of text-to-image synthesis.
\newblock \emph{CoRR}, abs/2306.09341, 2023.

\bibitem[Wu et~al.(2024)Wu, Hao, Zhang, Sun, Huang, Song, Liu, and
  Li]{Wu2024_DRS}
Xiaoshi Wu, Yiming Hao, Manyuan Zhang, Keqiang Sun, Zhaoyang Huang, Guanglu
  Song, Yu Liu, and Hongsheng Li.
\newblock Deep reward supervisions for tuning text-to-image diffusion models.
\newblock In \emph{Proc. ECCV}, 2024.

\bibitem[Xie et~al.(2025)Xie, Zhang, Yang, Hutter, and Xu]{Xie2025SPO}
Zhengpeng Xie, Qiang Zhang, Fan Yang, Marco Hutter, and Renjing Xu.
\newblock Simple policy optimization.
\newblock In \emph{Proc. ICML}, 2025.

\bibitem[Xu et~al.(2023)Xu, Liu, Wu, Tong, Li, Ding, Tang, and
  Dong]{Xu2023_ImageReward_ReFL}
Jiazheng Xu, Xiao Liu, Yuchen Wu, Yuxuan Tong, Qinkai Li, Ming Ding, Jie Tang,
  and Yuxiao Dong.
\newblock {ImageReward}: {L}earning and evaluating human preferences for
  text-to-image generation.
\newblock In \emph{Proc. NeurIPS}, 2023.

\bibitem[Xue et~al.(2025{\natexlab{a}})Xue, Ge, Zhang, Li, and
  Ma]{xue2025advantageweightedmatchingaligning}
Shuchen Xue, Chongjian Ge, Shilong Zhang, Yichen Li, and Zhi-Ming Ma.
\newblock Advantage weighted matching: {A}ligning {RL} with pretraining in
  diffusion models.
\newblock \emph{CoRR}, abs/2509.25050, 2025{\natexlab{a}}.

\bibitem[Xue et~al.(2025{\natexlab{b}})Xue, Wu, Gao, Kong, Zhu, Chen, Liu, Liu,
  Guo, Huang, and Luo]{Xue2025_DanceGRPO}
Zeyue Xue, Jie Wu, Yu Gao, Fangyuan Kong, Lingting Zhu, Mengzhao Chen, Zhiheng
  Liu, Wei Liu, Qiushan Guo, Weilin Huang, and Ping Luo.
\newblock {DanceGRPO}: {U}nleashing {GRPO} on visual generation.
\newblock \emph{CoRR}, abs/2505.07818, 2025{\natexlab{b}}.

\bibitem[Zelikman et~al.(2022)Zelikman, Wu, Mu, and
  Goodman]{zelikman2022starbootstrappingreasoningreasoning}
Eric Zelikman, Yuhuai Wu, Jesse Mu, and Noah~D. Goodman.
\newblock {STaR}: {B}ootstrapping reasoning with reasoning.
\newblock In \emph{Proc. NeurIPS}, 2022.

\bibitem[Zhang et~al.(2025)Zhang, Zhang, Gu, Zhang, Susskind, Jaitly, and
  Zhai]{Zhang2024_ImprovingGFlowNets}
Dinghuai Zhang, Yizhe Zhang, Jiatao Gu, Ruixiang Zhang, Joshua~M. Susskind,
  Navdeep Jaitly, and Shuangfei Zhai.
\newblock Improving {GFlowNets} for text-to-image diffusion alignment.
\newblock \emph{TMLR}, 2025.

\end{thebibliography}
